%% file: main.tex
\documentclass[letterpaper, 10 pt, conference]{ieeeconf}  

\IEEEoverridecommandlockouts                              

\usepackage[linesnumbered,ruled,noline]{algorithm2e}
\SetAlgoNoEnd
\SetAlgoNoLine
\DontPrintSemicolon

\usepackage{algorithmicx}
\usepackage[noend]{algpseudocode}
\usepackage{amsmath}
\usepackage{colortbl}
\usepackage{xcolor}
\usepackage{array}
\usepackage{placeins}

\SetAlgoNlRelativeSize{0}
\newcommand{\review}[1] {\textcolor{black}{#1}}

\input{tex_packages.tex}
\input{tex_settings.tex}

\newtheorem*{remark}{Remark}

\usepackage{verbatim}

\title{\LARGE \bf
Incremental Language Understanding for Online Motion Planning of Robot Manipulators
}

\author{Mitchell Abrams$^{*, 1}$, Thies Oelerich$^{*, 2}$, Christian Hartl-Nesic$^{2}$, Andreas Kugi$^{2, 3}$, Matthias Scheutz$^{1}$
\thanks{This work was not supported by any organization.}
\thanks{$^{*}$ Authors contributed equally}%
\thanks{$^{1}$ Authors are with the Human-Robot Interaction Lab, Tufts University,
    Boston, USA,
    {\tt\small \{mitchell.abrams, matthias.scheutz\}@tufts.edu}}%
\thanks{$^{2}$ Authors are with the Automation and Control Institute (ACIN), TU
    Wien, Vienna, Austria,
    {\tt\small \{oelerich, hartl, kugi\}@acin.tuwien.ac.at}}%
\thanks{$^{3}$Andreas Kugi is with the center for Vision, Automation \& Control,
    AIT Austrian Institute of Technology GmbH, Vienna, Austria
{\tt\small andreas.kugi@ait.ac.at}}%
}

\begin{document}

\maketitle
\thispagestyle{empty}
\pagestyle{empty}

\begin{abstract}
	Human-robot interaction requires robots to process language incrementally, adapting their actions in real-time based on evolving speech input. Existing approaches to language-guided robot motion planning typically assume fully specified instructions, resulting in inefficient stop-and-replan behavior when corrections or clarifications occur. In this paper, we introduce a novel reasoning-based incremental parser which integrates an online motion planning algorithm within the cognitive architecture. Our approach enables continuous adaptation to dynamic linguistic input, allowing robots to update motion plans without restarting execution. The incremental parser maintains multiple candidate parses, leveraging reasoning mechanisms to resolve ambiguities and revise interpretations when needed. By combining symbolic reasoning with online motion planning, our system achieves greater flexibility in handling speech corrections and dynamically changing constraints. We evaluate our framework in real-world human-robot interaction scenarios, demonstrating online adaptions of goal poses, constraints, or task objectives. Our results highlight the advantages of integrating incremental language understanding with real-time motion planning for natural and fluid human-robot collaboration. The experiments are demonstrated in the accompanying video at~\url{www.acin.tuwien.ac.at/42d5}.
\end{abstract}

\section{INTRODUCTION}

\begin{figure}[t]
	\centering
	\def\svgwidth{\linewidth}
	\footnotesize
	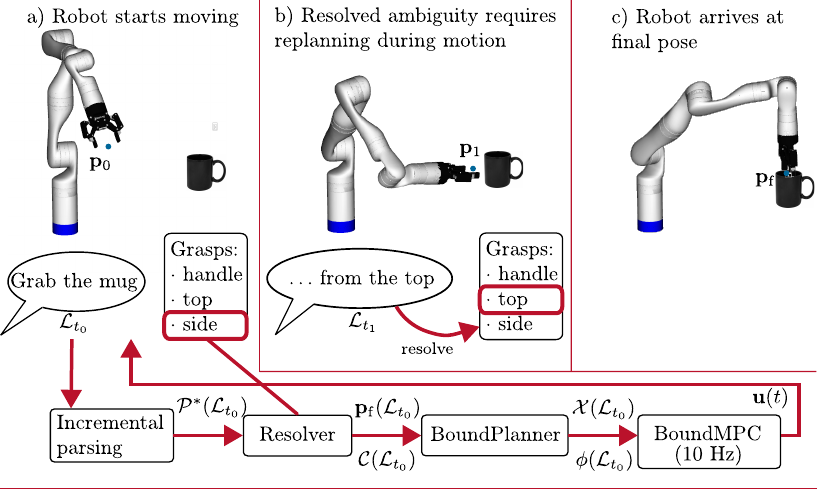
	\caption{Schematic of the incremental motion planning framework. The natural language instruction $\mathcal{L}_{t_{0}}$ is parsed to $\mathcal{P}^{*}(\mathcal{L}_{t_0})$. The resolver translates this to a goal pose $\mathbf{p}_\mathrm{f}(\mathcal{L}_{t_0})$ and a set of constraints $\mathcal{C}(\mathcal{L}_{t_0})$. The ambiguity in of the grasping pose is resolved randomly to a side grasp. Based on this, BoundPlanner~\cite{oelerichBoundPlannerConvexsetbasedApproach2025} computes a set of admissible states $\mathcal{X}(\mathcal{L}_{t_0})$ and cost function parameters $\phi(\mathcal{L}_{t_0})$ and passes it to BoundMPC~\cite{oelerichBoundMPCCartesianPath2025} which computes the joint input $\vec{u}(t)$ at \SI{10}{\hertz}.
	As the grasp assumption is false, the human instructor utters the language instruction $\mathcal{L}_{t_{1}}$ at time \review{$t_{1} > t_{0}$} to correct the robot. This correction requires a replanning at pose $\vec{p}_1$ which uses the same steps as before. The robot then reaches the final pose $\vec{p}_{\mathrm{f}}$.}
	\vspace{-3mm}
	\label{fig:schematic}
\end{figure}

As robots become more involved in everyday life, they are expected to interact seamlessly with humans. The interactions involve commanding the robot to perform a certain task or asking for help. Language instructions play a crucial role in these situations and robots need to understand human intentions. However, natural language is often ambiguous, and humans may need to correct or specify their instructions on the fly. An example is given in~\cref{fig:schematic}. The instruction ``\textit{Grab the mug}'' does not specify a particular way to grasp the mug, whether by the handle or from the top. As the robot starts reaching for the handle, the human can add ``\textit{from the top}'' to the instruction, to which the robot needs to react in real time. Such a behavior requires a continuous interaction between language understanding and motion planning. This is lacking in current works, which often assume natural language instructions to be given in text form~\cite{oelerichLanguageguidedManipulatorMotion2024, linText2MotionNaturalLanguage2023, huangVoxPoserComposable3D2023}. Real speech input is used in~\cite{lynchInteractiveLanguageTalking2022}, but the setting is limited, and the instruction is still parsed in its entirety but not incrementally. When the robot waits for the full instruction in order to begin planning a suitable motion and then stops and replans once the motion needs to be adapted, then this is time-consuming and frustrating for the human instructor.
Existing work in~\cite{cuiNoRightOnline2023} uses language corrections but requires manual control interventions and only wrong actions are corrected, but the task is not changed.
Our work specifically focuses on real-time speech parsing with online motion planning, mandatory for real human-robot interaction.

Incremental language processing offers flexibility, allowing robots to process linguistic input as it arrives rather than waiting for full utterance completion. Humans naturally employ incremental processing, continuously updating their understanding and refining interpretations based on evolving context~\cite{tanenhaus1995sentence, levelt1989speaking}. Similarly, a robot that processes language incrementally can begin formulating an action plan based on an initial command, and dynamically revise its plan as new linguistic constraints emerge.

There are various incremental language processing systems; some focus on syntactic parsing \cite{ peldszus2012incremental} or semantic parsing \cite{constantin2019incremental, devault2003domain}, while others are designed specifically for robotic agents \cite{kennington2020rrsds} or built into incremental dialogue processing software such as InproTK \cite{baumann2012inprotk} and Retico \cite{michael2019retico}. We argue that incremental language processing can benefit online replanning in robotic agents, in contrast to their non-incremental counterparts. Non-incremental parsers—such as Abstract Meaning Representation (AMR) parsers \cite{abrams2020graph}, dependency parsers \cite{nivre2010dependency}, or Large Language Models (LLMs) used for text-based robot instructions \cite{lin2023text2motion}---require utterances to be complete before parsing, delaying action planning and making real-time interaction infeasible.





As mentioned, a motion planner that can react to changes in the instructions online is needed in order to couple it with an incremental parsing system. This includes motion constraints that may be added incrementally. Imagine the robot manipulating a cup filled with a liquid. The robot might not know that it needs to keep it upright to avoid spilling, which the human can communicate with the corrective statement ``\textit{keep the cup upright}''. This new motion constraint needs to be taken into account during the motion of the robot. Considering Cartesian constraints in an online motion planner is challenging.
Resolving ambiguity of language instructions needs to work online and promptly because otherwise, the actions taken by the robot might not be reversible anymore. When the robot transports a filled cup but is not aware of the upright constraint, an online correction is able to limit the damage. When this is not the case, the robot might spill the liquid during its motion, which is not reversible.

Recently,~\cite{oelerichBoundMPCCartesianPath2025} proposed an optimization-based approach using a Cartesian bounded reference path which is used in this work in combination with the bounded reference path planner~\cite{oelerichBoundPlannerConvexsetbasedApproach2025}. This combination is used for online motion planning in this work. It is able to rigorously communicate constraints from the language input to the motion planner using a model-based approach with complete consideration of the environment. The motion instructions are created by the novel incremental language parser that utilizes explicit reasoning to parse utterances incrementally and understand at which point in time the utterance is meaningful to send to the underlying motion planner. This way it is tailored specifically to robot motion planning.
The proposed approach is demonstrated on a 7-DoF robot manipulator in four scenarios. These scenarios demonstrate how the proposed framework handles changes in goal pose, adding motion constraints, i.e., a region avoidance and upright orientation constraint, and changing the speed of the motion while the robot is moving. We further demonstrate that using an offline motion planner degrades performance.

The main contributions of this paper are
\begin{itemize}
	\item a novel incremental language parser specifically tailored to interact with robot motion planning.
	\item the tight integration of this incremental language parser with online motion planning based on BoundMPC~\cite{oelerichBoundMPCCartesianPath2025} and BoundPlanner~\cite{oelerichBoundPlannerConvexsetbasedApproach2025}. This novel combination allows fast acting in human-robot collaborations and quick corrections of undesired motions.
\end{itemize}

\section{RELATED WORK}

\subsection{Incremental Processing}

Humans process language incrementally, interpreting linguistic input as it is received rather than waiting for full utterance completion~\cite{tanenhaus1995sentence, levelt1989speaking}. This ability enables rapid adaptation in interactive settings, allowing speakers and listeners to dynamically adjust their actions based on evolving context~\cite{tanenhaus2008language, steedman1989grammar}. Psycholinguistic research has shown that humans make predictions about meaning in real time, guided by syntactic, semantic, and pragmatic constraints~\cite{berg1990speaking}.

Motivated by these human capabilities, computational systems for incremental natural language understanding have been developed to support real-time processing and early action execution~\cite{skantze2009incremental}. These systems enable dialogue components to process partial utterances, reducing response latency and improving interactive fluency.

Several approaches to incremental language processing have emerged, each focusing on different aspects of the problem. Some systems prioritize incremental syntactic parsing~\cite{schuler2009framework, peldszus2012incremental}, allowing for continuous refinement of the sentence structure as new words arrive. Others emphasize incremental semantic interpretation~\cite{constantin2019incremental, devault2003domain}, where meaning is continuously updated based on available linguistic and contextual information. One important subfield is incremental reference resolution, in which systems dynamically resolve a referent without waiting for a fully disambiguated description~\cite{poesio2011incremental}.

While these systems have demonstrated the advantages of incremental processing, most approaches are designed for interpretation rather than planning and action execution. In situated environments, incremental processing must go beyond language and integrate with an agent\textquotesingle s perception, reasoning, and action planning. Some systems have begun to explore this intersection by incorporating multimodal cues such as gaze and gesture to improve reference resolution~\cite{kennington2017simple}. Others have focused on real-time robotic interaction, such as the RISE system, which allows robots to maintain multiple possible interpretations of an unfolding utterance~\cite{brick2007incremental}. However, a critical gap remains: most incremental systems do not support real-time action planning and execution based on partially complete utterances.

\subsection{Motion Planning}

Motion planning for robot manipulators is a challenging and active research field. It is broadly divided into offline~\cite{marcucciMotionPlanningObstacles2023,jankowskiVPSTOViapointbasedStochastic2023} and online motion planning~\cite{beckModelPredictiveTrajectory2024a,oelerichBoundMPCCartesianPath2025}. Offline motion planning is often employed in industrial settings where minimum-time trajectories can improve the production output or for very precise tasks which are difficult to compute. However, when the robot is acting in an uncertain or dynamic environment, e.g., in an environment shared with humans, the robot must be able to adapt its behavior online~\cite{oelerichModelPredictiveTrajectory2024}. The needed real-time capability can be achieved by planning only over a finite time horizon. Popular methods include sampling- and optimization-based model predictive control (MPC)~\cite{beckModelPredictiveTrajectory2024a,oelerichBoundMPCCartesianPath2025,kalakrishnanSTOMPStochasticTrajectory2011} and reinforcement learning~\cite{schmittPlanningReactiveManipulation2019}.
However, finite horizons for planning often yield local minima due to obstacles and joint limits. Thus, the works~\cite{oelerichLanguageguidedManipulatorMotion2024} and~\cite{oelerichBoundPlannerConvexsetbasedApproach2025} split the planning problem into reference path planning and trajectory planning. The reference path is planned from the initial to the goal pose in Cartesian space and guides the finite-horizon joint-trajectory planning. This separation enables real-time planning. Popular path planning methods include RRT~\cite{karamanSamplingbasedAlgorithmsOptimal2011} and convex sets~\cite{liCollisionFreeTrajectoryOptimization2024, oelerichLanguageguidedManipulatorMotion2024, oelerichBoundPlannerConvexsetbasedApproach2025}. In this work we use BoundPlanner~\cite{oelerichBoundPlannerConvexsetbasedApproach2025} in combination with BoundMPC~\cite{oelerichBoundMPCCartesianPath2025}, which was shown to be computationally efficient and is able to adhere to Cartesian bounds.

\subsection{End-to-End Learning}
Another interesting development is the emergence of end-to-end learning as used in~\cite{cuiNoRightOnline2023, lynchInteractiveLanguageTalking2022}. However, this lacks any guarantees of constraint satisfaction and structured incremental understanding. Therefore, this approach is not further considered in this work.

\section{INCREMENTAL LANGUAGE PROCESSING}
\label{sec:incremental_language}

~\cite{schlangen2011general} present a high-level framework for designing incremental language processing systems and outline key architectural considerations: \textit{modularity}, \textit{granularity}, and \textit{revisability}. In this section, we introduce an incremental parsing approach that enables a robotic agent to process language in real time, submit plans on the fly, and dynamically adjust its actions based on evolving linguistic input. Our system bridges this gap by supporting: 1) Partial parsing (\textit{modularity}): the system generates usable interpretations before an utterance is complete; 2) Integration with motion planning (\textit{granularity}): words and partial parses are used to submit and update robot action plans; and 3) Real-time plan revision (\textit{revisability}): the system dynamically updates actions and utterance interpretations when new constraints are introduced. By combining incremental parsing with the online motion planning framework described in~\cref{sec:motion_planning}, our system efficiently handles dynamic human-robot interactions, adapting to spoken corrections and clarifications utilizing replanning during the robot\textquotesingle s motion.


\begin{algorithm}[ht]
	\footnotesize
	\caption{Incremental Chart Parsing}
	\label{alg:chart_parser}
	\SetAlgoNlRelativeSize{-2}
	\SetNlSty{textbf}{\footnotesize}{}
	\SetNlSkip{0.6em}
	\SetAlgoLined
	\SetAlgoNoEnd
	\SetInd{0.5em}{0.5em}

	\KwIn{Language Instruction $\mathcal{L}_{t} = (w_0, \dots, w_{n-1})$, \\
		Dictionary $D$}
	\KwOut{Parsed utterance $\mathcal{P}(\mathcal{L}_{t})$}

	\If{chart uninitialized}{ InitializeChart$(n)$ }
	\Else{ExpandChart$(n)$ }

	\ForEach{$w_i \in \mathcal{L}_{t}$}{
		\tcp{Add initial word node to chart}
		$\mathcal{E}_i \gets D.\text{lookup}(w_i)$ \;

		\If{$\mathcal{E}_i \neq \emptyset$}{ AddNode$(\mathcal{E}_i, i, i)$ }
	}

	\For{$s = 2$ \textbf{to} $n$}{
		\For{$j = 0$ \textbf{to} $n - s$}{
			$k \gets j + s - 1$ \;
			\For{$m = j$ \textbf{to} $k - 1$}{
				\ForEach{$(L, R) \in \text{chart}[j][m] , \text{chart}[m+1][k]$}{
					\tcp{Try combining constituents}
					$\mathcal{N} \gets \text{Combine}(L, R)$ \;
					\If{$\mathcal{N} \neq \emptyset$}{
						\tcp{Add new node to chart}

						chart$[j][k] \gets$ chart$[j][k] \cup \{\text{Node}(L, R, \mathcal{N})\}$ \;
					}
				}
			}
		}
	}

	\Return best parse $\mathcal{P} \in \text{chart}[0][n-1]$
\end{algorithm}


Our parser constructs a hierarchical structure in several stages. First, during lexical processing, the words are assigned their respective syntactic categories and semantic representations (using CCG-style (combinatory categorial grammar) parsing) combining syntax and semantics \review{based on predefined grammar rules}.
Words are dynamically inserted into the chart and stored in separate cells as they arrive. The parser then begins phrase-level combination. The final parse is selected from the final cell of the chart. If a user stops mid-utterance, such as saying \textit{``grab the mug...''} without completing or adding more to the utterance (\textit{``grab the mug by the top''}), the parser maintains a partial parse and uses it as a basis for future input. This ensures that parsing remains fluid and can accommodate dynamically evolving utterances.

The incremental chart parsing algorithm (\textbf{Algorithm~\ref{alg:chart_parser}}) presents a more formal description. The algorithm takes as input an utterance $\mathcal{L}_{t}= (w_0, w_1, \dots, w_i)$ at time $t$ and a dictionary $D$ containing stored grammar rules. It outputs a parsed representation $\mathcal{P}(\mathcal{L}_{t})$. The chart data structure is initialized (\textbf{lines 1--4}) if not already allocated; otherwise, it is expanded to accommodate newly received words.

Each word $w_i$ in the utterance undergoes lexical processing (\textbf{lines 5--8}). The function $D.\text{lookup}(w_i)$ retrieves a set of grammar rules $\mathcal{E}_i$ associated with the word. If the word is unknown ($\mathcal{E}_i = \emptyset$), it is skipped. Otherwise, a node\footnote{Nodes are created using the constructor \texttt{\small new Node(List<Entry> entries)}, which generates lexical nodes from dictionary entries. As parsing progresses, the method \texttt{\small combine(Node left, Node right)} merges compatible nodes into higher-level phrases.} is created from these rules and inserted into the chart at position $(i, i)$, see~\cref{tab:chart_parsing}. Once all words are inserted, the parser performs phrase combination using a bottom-up dynamic programming approach (\textbf{lines 9--16}). It iterates over spans of length $s$, progressively merging smaller phrases into larger constituents. At each span, split points $m$ are considered, allowing for binary branching combinations of subparses. The function \texttt{Combine}(L, R) checks whether two nodes can be merged using grammatical constraints.

If a valid rule $\mathcal{N}$ permits the combination of a left parse $L$ and a right parse $R$, a new node is created and inserted into the chart (\textbf{line 14}). The process continues until the entire utterance is analyzed. The final parse is retrieved from chart cell $[0,  n-1]$ (\textbf{line 17}), which stores the final interpretations. Multiple parses of the same words can be stored in the chart, allowing a reasoning mechanism to return the best interpretation\footnote{Multiple parses of are stored in the last cell and a single parse can be returned depending on the context (e.g. available referents or objects)  of the utterance that reflects the best interpretation.}. The structured storage of alternative parses also allows the system to defer disambiguation until further context is available. Our parser operates with a worst-case time complexity of $O(n^3)$, similar to CYK parsing \cite{younger1967recognition}.
\review{The proposed parser is designed to work with the English language. Extending this to other languages requires an additional translation step or language-specific adaptions.}


\begin{table}[htbp]
	\centering
	\tiny
	\renewcommand{\arraystretch}{1.2}
	\setlength{\tabcolsep}{2pt}
	\begin{tabular}{|c|c|c|c|c|c|c|}
		\hline
		\textbf{Idx}     & \textbf{0}               & \textbf{1}              & \textbf{2}              & \textbf{3}             & \textbf{4}              & \textbf{5}              \\ \hline
		\tiny \textbf{0} & \cellcolor{gray!20} grab & VP $\to$ grab           & VP $\to$ grab the mug   & {}                     &                         &
		\cellcolor{gray!20}
		\begin{tabular}{@{}c@{}}
			\textbf{S $\to$ grab (the mug) (by the top)} \\
			\textcolor{gray}{S $\to$ grab (the mug by the top)}
		\end{tabular}                                                                                                                           \\ \hline
		\tiny \textbf{1} &                          & \cellcolor{gray!20} the & NP $\to$ the mug        & {}                     & {}                      &                         \\ \hline
		\tiny \textbf{2} &                          &                         & \cellcolor{gray!20} mug & {}                     & {}                      &                         \\ \hline
		\tiny \textbf{3} &                          &                         &                         & \cellcolor{gray!20} by & {}                      &
		\cellcolor{gray!20}
		\begin{tabular}{@{}c@{}}
			\textbf{PP $\to$ by the top} \\
			\textcolor{gray}{PP $\to$ by the top (modifies NP)}
		\end{tabular}                                                                                                                           \\ \hline
		\tiny \textbf{4} &                          &                         &                         &                        & \cellcolor{gray!20} the & NP $\to$ the top        \\ \hline
		\tiny \textbf{5} &                          &                         &                         &                        &                         & \cellcolor{gray!20} top \\ \hline
	\end{tabular}
	\caption{Parsing chart for ``grab the mug by the top''}
	\label{tab:chart_parsing}
\end{table}

\subsection{Parsing Walkthrough}\label{ssec:parsing_walkthrough}

Table~\ref{tab:chart_parsing} illustrates the incremental chart parse for \textit{``grab the mug by the top.''} The parsing process begins at the lexical level, where each word is inserted into its corresponding diagonal cell. The first word, \textit{``grab''}, is identified as a verb and stored at position $(0,0)$. Next, \textit{``the''} and \textit{``mug''} (at positions $(1,1)$ and $(2,2)$, respectively) are recognized as part of a noun phrase (NP $\to$ \textit{``the mug''}). At the phrase combination stage, the parser merges \textit{``the''} and \textit{``mug''} into NP $\to$ \textit{``the mug''}, stored at $(1,2)$. This NP is then combined with the verb at $(0,0)$, forming a VP $\to$ \textit{``grab the mug''}, a candidate parse for a complete verb phrase at $(0,2)$. This is the first completed parse that can be submitted as a goal for the agent ($\mathcal{P}^{*}(\mathcal{L}_{t_0})$ at time $t_0$ rather than \textit{``grab the''} ($\mathcal{P}(\mathcal{L})$).
We use a first-order logic (FOL) style semantic representation\footnote{This representation includes the speaker intent (\texttt{\footnotesize INSTRUCT}) and the core action (\texttt{\footnotesize graspObject}) on an object (\texttt{\footnotesize mug}). The object reference (\texttt{\footnotesize mug}) can resolve to a specific object in the environment known by the agent.}:\\

\vspace{-3mm}
\begingroup
\scriptsize
\begin{verbatim}
  INSTRUCT(speaker,listener,graspObject(listener,mug))
\end{verbatim}
\endgroup

While the agent starts to achieve this goal (grabbing the relevant mug), the parse continues until it gets to another completed parse $\mathcal{P}^{*}(\mathcal{L}_{t_1})$ at time $t_1$. The next word, \textit{``by''}, is inserted at $(3,3)$ as a preposition (PP $\to$ \textit{``by''}). Similarly, \textit{``the top''} is recognized as a noun phrase (NP $\to$ \textit{``the top''}) and stored at $(4,5)$. The parser then combines these elements into PP $\to$ \textit{``by the top''}, stored at $(3,5)$.

The final parse selection occurs at cell $(0,5)$, where the parser considers two interpretations:
(1) \textbf{Best Parse (Bold):} S $\to$ \textit{``grab (the mug) (by the top)''}, where \textit{``by the top''} modifies the verb, indicating the manner of grabbing.
(2) \textbf{Alternative Parse (Gray):} S $\to$ \textit{``grab (the mug by the top)''}, where \textit{``by the top''} modifies the noun, implying a location.

This approach ensures that multiple interpretations are retained within the chart, enabling flexible reasoning about attachment ambiguities in real time. Each parse stored in \texttt{chart[0][n-1]} can be evaluated and resolved using additional context or external constraints. The final parse places additional constraints on the original plan ($\mathcal{P}^{*}(\mathcal{L}_{t_1})$):\\

\vspace{-3mm}
\begingroup
\scriptsize
\begin{verbatim}
INSTRUCT(speaker,listener,graspObject
        (listener,mug), by(mug, top))
\end{verbatim}
\endgroup

\section{ONLINE MOTION PLANNING}
\label{sec:motion_planning}

The incremental language parsing system from~\cref{sec:incremental_language} is used to parse a motion instruction $\mathcal{L}_{t_{k}}$ at time $t_{k}$ for a robot manipulator. Once a meaningful phrase $\mathcal{P}^{*}(\mathcal{L}_{t_{k}})$ is obtained, the motion planner needs to react online to the adaptions. The resolver in~\cref{fig:schematic} resolves $\mathcal{P}^{*}(\mathcal{L}_{t_{k}})$ into a desired final pose $\vec{p}_\mathrm{f}(\mathcal{L}_{t_{k}})$ and a set of constraints $\mathcal{C}(\mathcal{L}_{t_{k}})$ as inputs to the motion planner. For the purpose of this work, the goal poses and obstacles are assumed to be known such that the resolver simply maps the parses, e.g., from~\cref{ssec:parsing_walkthrough}, to poses and constraints. In future work, this will be extended to include vision information. For example, the specification \textit{``keep the cup upright''} will add a constraint on the orientation to $\mathcal{C}(\mathcal{L}_{t_{k}})$.

\subsection{From Parsed Instruction to Safe Robot Motion}

In this work, a combination of global reference path planning in the Cartesian space with subsequent trajectory planning in the joint space over a finite time horizon is used. A reference path $\pi$ is planned by BoundPlanner~\cite{oelerichBoundPlannerConvexsetbasedApproach2025} and bounded by convex sets such that the robot\textquotesingle s end effector is collision free if it remains within the bounds along $\pi$ that ensure the satisfaction of constraints $\mathcal{C}(\mathcal{L}_{t_{k}})$. These bounds define the set of permissible states
\begin{equation}
	\mathcal{X}(\mathcal{L}_{t_{k}}) = \mathcal{X}_{\mathrm{safe}}(\mathcal{L}_{t_{k}}) \cup \mathcal{X}_{\mathrm{task}}(\mathcal{L}_{t_{k}}) \cup \mathcal{X}_{\mathrm{robot}}
\end{equation}
for the robot, which constitutes safe states $\mathcal{X}_{\mathrm{safe}}(\mathcal{L}_{t_{k}})$, task-specific states $\mathcal{X}_{\mathrm{task}}(\mathcal{L}_{t_{k}})$, and kinematic and dynamic states of the robot $\mathcal{X}_{\mathrm{robot}}$. The set of safe states $\mathcal{X}_{\mathrm{safe}}(\mathcal{L}_{t_{k}})$ will always ensure obstacle avoidance.
Convex sets define the task set $\mathcal{X}_{\mathrm{task}}(\mathcal{L}_{t_{k}})$ and the safety set $\mathcal{X}_{\mathrm{safe}}(\mathcal{L}_{t_{k}})$ for the position and orientation of the robot\textquotesingle s end effector and the collision avoidance of the robot\textquotesingle s kinematic chain. The set $\mathcal{X}_{\mathrm{robot}}$ is independent of the instruction $\mathcal{L}_{t_{k}}$ as the robot\textquotesingle s limits do not change. The original formulation of BoundPlanner~\cite{oelerichBoundPlannerConvexsetbasedApproach2025} does not handle orientation constraints. This work lifts this limitation by using box constraints for the orientation formulation introduced in~\cite{oelerichBoundMPCCartesianPath2025}.
The parsed language instruction $\mathcal{P}^{*}(\mathcal{L}_{t_{k}})$ defines $\vec{p}_\mathrm{f}(\mathcal{L}_{t_{k}})$ and $\mathcal{C}(\mathcal{L}_{t_{k}})$, which need to be translated to the set $\mathcal{X}(\mathcal{L}_{t_{k}})$, and the parameters $\phi(\mathcal{L}_{t_{k}})$ using BoundPlanner. The parameters $\phi(\mathcal{L}_{t_{k}})$ parametrize the cost function for the subsequent trajectory optimization influencing the shape and speed of the trajectory.
Specifically, this work presents the following adaptions in several examples:
\begin{itemize}
	\item  $\vec{p}_\mathrm{f}(\mathcal{L}_{t_{k}}) \rightarrow \mathcal{X}_{\mathrm{task}}(\mathcal{L}_{t_{k}})$: Change of the goal pose by replanning the reference path,
	\item $\mathcal{C}(\mathcal{L}_{t_{k}}) \rightarrow \mathcal{X}_{\mathrm{safe}}(\mathcal{L}_{t_{k}})$: Change of the avoidable obstacles or allowed orientations,
	\item $\mathcal{C}(\mathcal{L}_{t_{k}}) \rightarrow \phi(\mathcal{L}_{t_{k}})$: Change of the motion speed as a soft constraint.
\end{itemize}
The adaptions will be discussed in more detail in~\cref{sec:experiments}.

The sets and parameters introduced above are further provided to BoundMPC~\cite{oelerichBoundMPCCartesianPath2025}, which uses online optimization to compute the joint input $\vec{u}(t)$ over a finite time horizon from time $t_{0} \geq t_{k}$ to the end time $t_\mathrm{e}$.
The finite horizon planning problem is formulated as
\begin{equation}
	\begin{aligned}
		\min_{\vec{u}(t)} & \quad c_{\phi(\mathcal{L}_{t_{k}})}(\vec{x}(t_\mathrm{e}), \vec{u}(t_\mathrm{e})) + \int_{t_{0}}^{t_\mathrm{e}} J_{\phi(\mathcal{L}_{t_{k}})}(\vec{x}(t), \vec{u}(t)) \mathrm{d}t \\
		\mathrm{s.t.}     & \quad \vec{x}(t_{0}) = \vec{x}_{0}                                                                                                                                                \\
		                  & \quad \vec{u}(t_{0}) = \vec{u}_{0}                                                                                                                                                \\
		                  & \quad \vec{x}(t) \in \mathcal{X}(\mathcal{L}_{t_{k}}) \quad \forall t : t_{0} \leq t \leq t_\mathrm{e}                                                                            \\
		                  & \quad \vec{u}(t) \in \mathcal{U} \quad \forall t : t_{0} \leq t \leq t_\mathrm{e}  \text{~,}
	\end{aligned}
	\label{eq:motion_planning_problem}
\end{equation}
with the state $\vec{x}(t)$, the input $\vec{u}(t)$, and their initial values $\vec{x}_{0}$ and $\vec{u}_{0}$ at $t_{0}$. The objective function is composed of the terminal cost $c_{\phi(\mathcal{L}_{t_{k}})}(\vec{x}(t_\mathrm{e}), \vec{u}(t_\mathrm{e}))$ and the stage cost $J_{\phi(\mathcal{L}_{t_{k}})}(\vec{x}(t), \vec{u}(t))$, which are parametrized by $\phi(\mathcal{L}_{t_{k}})$. The permissible inputs are given by the set $\mathcal{U}$.
For more information on BoundPlanner and BoundMPC, the reader is referred to~\cite{oelerichBoundMPCCartesianPath2025} and~\cite{oelerichBoundPlannerConvexsetbasedApproach2025}, respectively.
The instruction $\mathcal{L}_{t_{k}}$ is constant in~\cref{eq:motion_planning_problem} because it is not predicted over the time horizon.
\begin{remark}
	This work focuses on adding constraints that need to be enforced immediately at time $t_{k}$, e.g., reacting to a new obstacle, which may render the current state infeasible. To overcome this, the respective constraints are enforced using slack variables. This is a common practice in optimization~\cite{schoelsNMPCApproachUsing2020a}.
\end{remark}

\subsection{Taxonomy of Constraints in Incremental Language Processing}

Incremental language processing enables dynamic adaptation of robot motion plans based on evolving linguistic constraints $\mathcal{C}(\mathcal{L}_{t})$. These constraints may be explicitly stated in speech or inferred from context and can emerge at different stages of execution. We categorize them into six types: manner, target, object, action, safety, and \review{sequential} constraints, as summarized in Table~\ref{tab:taxonomy}.

\begin{table*}[ht]
	\begin{minipage}[b]{0.62\linewidth}
		\centering
		\caption{Taxonomy of incremental constraints. Each constraint corresponds to a specific adaption in the motion planning~\cref{eq:motion_planning_problem}.} \label{tab:taxonomy}
		\scriptsize
		\begin{tabular}{lcl}  
			\textbf{Type of Constraint $\mathcal{C}(\mathcal{L}_{t})$} & \textbf{Adaptions}                             & \textbf{Example}                                                   \\
			\hline
			Manner Constraints                                         & $\phi(\mathcal{L}_{t})$                        & ``Pass the screwdriver... \textit{but go faster}.''                \\
			Target Constraints                                         & $\mathcal{X}_{\mathrm{task}}(\mathcal{L}_{t})$ & ``Put the apple in the box... \textit{no, the one on the right}.'' \\
			Object Constraints                                         & $\mathcal{X}_{\mathrm{task}}(\mathcal{L}_{t})$ & ``Grab the mug... \textit{no, the blue one}.''                     \\
			Action Constraints                                         & $\mathcal{X}_{\mathrm{task}}(\mathcal{L}_{t})$ & ``Grab the apple... \textit{no, push it}.''                        \\
			Safety Constraints                                         & $\mathcal{X}_{\mathrm{safe}}(\mathcal{L}_{t})$ & ``Move the cup... \textit{and keep it upright}.''                  \\
			\review{Sequential} Constraints                            & $\mathcal{X}_{\mathrm{task}}(\mathcal{L}_{t})$ & ``Pick up the apple... \textit{after you put down the mug}.''      \\
			\hline
		\end{tabular}
	\end{minipage}%
	\begin{minipage}[b]{0.35\linewidth}
		\centering
		\caption{Experiment 1: Comparison of our method with VP-STO~\cite{jankowskiVPSTOViapointbasedStochastic2023} for the two planning instances during the grabbing of the mug.}\label{tab:mug_comp}
		\scriptsize
		\begin{tabular}{ccc}
			                                                       & \multicolumn{1}{c}{\textbf{Ours}} & \multicolumn{1}{c}{\textbf{VP-STO}} \\
			\hline
			Planning time $t_{\mathrm{plan}}$ / \si{\second}       & 0.02                              & 5.60                                \\
			                                                       & 0.02                              & 5.62                                \\
			\hline
			Trajectory duration $t_{\mathrm{traj}}$ / \si{\second} & 3.10                              & 2.89                                \\
			                                                       & 4.70                              & 2.97                                \\
			\hline
			Total task duration $t_{\mathrm{task}}$ / \si{\second} & 7.84                              & 17.08                               \\
			\hline
		\end{tabular}
	\end{minipage}
\end{table*}

Manner constraints adjust how an action is performed without altering its goal. These constraints are often introduced via adverbs or prepositional phrases, such as ``Pass the screwdriver \textit{but go faster},'' which modifies the execution speed. Target constraints refine the intended destination. For instance, ``Put the apple in the box... \textit{no, the one on the right}'' forces the robot to update its spatial goal in real time. Object constraints resolve referential ambiguity when multiple items fit a description. If a robot initially reaches for the wrong object, additional clarification—such as ``Grab the mug... \textit{no, the blue one}''—helps disambiguate the target. Action constraints redefine or modify the task itself. A command like ``Grab the apple'' might initially be interpreted as a grasping action, but a correction such as ``\textit{no, push it}'' redirects the execution to a different motion. Safety constraints introduce requirements to prevent damage or failure. Commands like ``Move the cup \textit{upright}'' ensure that the robot adjusts its trajectory to maintain a stable orientation. \review{Sequential} constraints establish dependencies between actions. A directive like ``Pick up the apple \textit{after you put down the mug}'' enforces a sequence that the robot must integrate dynamically into its plan.

This taxonomy highlights the need for an incremental parsing system capable of maintaining multiple candidate interpretations and refining them in real time as new constraints emerge.

\section{EXPERIMENTS}
\label{sec:experiments}

In this section, the proposed framework is evaluated in three experiments on the real robot:

\begin{enumerate}
	\item Grabbing a mug and changing the desired grasp direction during the motion,
	\item transferring a mug filled with liquid where additional constraints (keeping the mug upright and avoiding going over a laptop) are added during the motion, and
	\item giving a screwdriver to a human where the human controls the speed of the robot\textquotesingle s motion through speech.
\end{enumerate}

These and more experiments are demonstrated in the accompanying video at~\url{www.acin.tuwien.ac.at/42d5}.

Natural language from the human operator is the only input for the motion generation in all experiments. The position of the objects, as well as the grasp poses, are assumed to be known to focus on the language processing and motion generation of the framework.
The proposed method is compared to the offline trajectory planner VP-STO~\cite{jankowskiVPSTOViapointbasedStochastic2023}, a state-of-the-art global trajectory planner. The maximum iterations are set to be 1000 to balance optimality and planning time.
The reader is referred to~\cite{oelerichBoundMPCCartesianPath2025} and~\cite{oelerichBoundPlannerConvexsetbasedApproach2025} for more comparisons of the used motion planner with state-of-the-art planners.
\review{The computational effort for the language parsing is below \SI{10}{\milli\second} for each parse which makes it real-time capable and more performant than querying an LLM.}

\subsection{Experiment 1: Grab a mug with replanning}

\begin{figure}
	\centering
	\def\axisdefaultwidth{\linewidth}
	\def\axisdefaultheight{0.7\linewidth}
	\addtolength{\abovecaptionskip}{-15pt}
	\input{tikz/scen_mug_graps.tex}
	\caption{Experiment 1: The robot is visualized during the grabbing of the mug at the initial pose $\vec{p}_{0}$, the replanning pose $\vec{p}_{1}$, and the final pose $\vec{p}_{\mathrm{f}}$. The black mug rests on a green box in front of the robot.}
	\label{fig:scen_mug_grasp}
\end{figure}
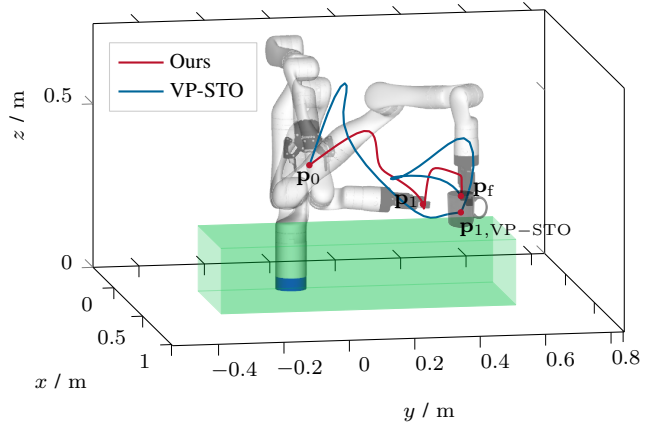

\begin{figure}
	\centering
	\def\axisdefaultwidth{0.95\linewidth}
	\def\axisdefaultheight{0.35\linewidth}
	\addtolength{\abovecaptionskip}{-5pt}
	\input{tikz/mug_speed.tex}
	\caption{Experiment 1: Norm of joint velocity $\dot{\vec{q}}$ while grabbing the mug.}
	\label{fig:scen_mug_speed}
\end{figure}
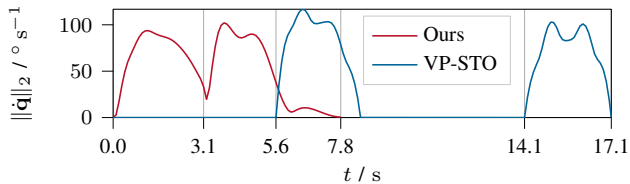

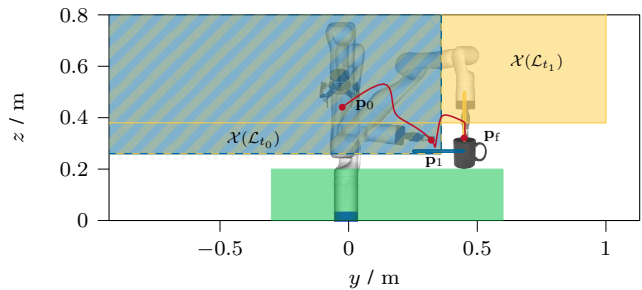
\begin{figure}
	\centering
	\def\axisdefaultwidth{\linewidth}
	\def\axisdefaultheight{0.5\linewidth}
	\addtolength{\abovecaptionskip}{-5pt}
	\input{tikz/mug_sets_fix.tex}
	\caption{Experiment 1: Visualization of the sets of permissible states $\mathcal{X}(\mathcal{L}_{t_{0}})$ in blue and $\mathcal{X}(\mathcal{L}_{t_{1}})$ in yellow for the mug grabbing in the $y$-$z$-plane. The blue sets overlaps with the yellow set in the striped area.}
	\label{fig:scen_mug_sets}
	\vspace{-3mm}
\end{figure}

In the first experiment, the robot is instructed to grab a mug using $\mathcal{L}_{0} = $ \textit{``Grab the mug''} at time $t_{0}$. This instruction does not contain information about how to grab the mug. There are different grasp possibilities, i.e., grab the handle or the body from the side or from the top. The robot starts to grab the mug from the side, but then the human operator intervenes and adds the instruction $\mathcal{L}_{1} = $ \textit{``from the top''} at time $t_{1}$. This requires the robot to change its plan during its motion. Specifically, it changes the set $\mathcal{X}_{\mathrm{task}}$ at $t_{1}$.
The trajectory of the robot is visualized in~\cref{fig:scen_mug_grasp}. The robot\textquotesingle s initial end effector pose is $\vec{p}_{0}$. When hearing the instruction $\mathcal{L}_{t_{0}}$, it starts moving to grab the mug from the side until the instruction $\mathcal{L}_{t_{1}}$ triggers a replanning at time $t_{1}$, where the end effector is at $\vec{p}_{1}$. The robot then proceeds to the final pose $\vec{p}_{\mathrm{f}}$. The end-effector trajectory of planning with VP-STO is also visualized. As VP-STO is unable to plan online, see~\cref{tab:mug_comp}, the robot proceeds to the first grasping point at $\vec{p}_{\mathrm{1, VP-STO}}$ and stops. Then, the trajectory from $\vec{p}_{\mathrm{1, VP-STO}}$ to $\vec{p}_{\mathrm{f}}$ is planned, and the robot moves to $\vec{p}_{\mathrm{f}}$.
With our proposed framework, the robot reacts promptly to the replanning and repositions its end effector according to the new grasping position. The norm of the joint velocity $\dot{\vec{q}}$ in~\cref{fig:scen_mug_speed} is only zero at the start and end of the motion, but there is no need to stop for the replanning at $t_1 = \SI{3.1}{\second}$. In comparison, VP-STO needs to plan for $t_{\mathrm{plan}} = \SI{5.6}{\second}$ after receiving the first instruction $\mathcal{L}_{t_{0}}$ before starting to move and has to remain at $\vec{p}_{\mathrm{1, VP-STO}}$ to plan a motion according to $\mathcal{L}_{t_{1}}$. Due to the longer planning times $t_{\mathrm{plan}}$, VP-STO takes longer to finish the task. The planning times $t_{\mathrm{plan}}$ and the trajectory times $t_{\mathrm{traj}}$ are summarized in~\cref{tab:mug_comp}. Our method has faster planning times $t_{\mathrm{plan}}$ due to the fast planning times of BoundPlanner~\cite{oelerichBoundPlannerConvexsetbasedApproach2025}.
Resolving ambiguity in instructions needs to work online and promptly. The trajectory times $t_{\mathrm{traj}}$ are faster for VP-STO as it plans the full trajectory instead of only a finite-horizon approach. However, the overall task time $t_\mathrm{task}$ is considerably lower with the proposed framework, which justifies using an online trajectory planner.
The sets of permissible states $\mathcal{X}(\mathcal{L}_{t_{0}})$ and $\mathcal{X}(\mathcal{L}_{t_{1}})$ for the robot\textquotesingle s end effector before and after the replanning are shown in~\cref{fig:scen_mug_sets}. These sets are defined by BoundPlanner and initially constrain the robot to stay to the left of the mug and then grab it. After the replanning, the space above the mug is added to the permissible states $\mathcal{X}(\mathcal{L}_{t_{1}})$ and the final convex set guides the robot in performing a safe grasping motion from the top.

\subsection{Experiment 2: Transfer a mug with task constraints}

In the second experiment, visualized in~\cref{fig:scen_avoid}, the robot is instructed to $\mathcal{L}_{t_0, \mathrm{upright}} = $ \textit{``Pass the mug''} at time $t_{0}$. For this, the robot has to pick up a mug and transfer it to a predefined position where the human can receive it. The mug is initially placed on a table next to a laptop. As the mug contains a liquid, spilling the liquid onto the laptop must be avoided, but the robot is not aware of this constraint. As the robot starts moving, the human realizes that the robot will move over the laptop and specifies the instruction at time $t_1 = \SI{6.4}{\second}$ by saying $\mathcal{L}_{t_1, \mathrm{upright}} = $ \textit{``but don't spill it''}. In the next run the human specifies $\mathcal{L}_{t_0, \mathrm{avoid}} = $ \textit{``Pass the mug but keep it upright''} at time $t_{0}$, the human further specifies $\mathcal{L}_{t_1, \mathrm{avoid}} = $ \textit{``and avoid going over the laptop''} at $t_{1}$ to increase safety. Both applications create a task constraint online. This does not change the desired final pose $\vec{p}_{\mathrm{f}}$ but the motion to reach it. Spillage avoidance creates a constraint on the orientation of the robot\textquotesingle s end effector, whereas avoiding the laptop adds an obstacle to the path planning.

\begin{figure}
	\centering
	\def\axisdefaultwidth{0.95\linewidth}
	\def\axisdefaultheight{0.28\linewidth}
	\pgfplotsset{group/vertical sep=14}
	\addtolength{\abovecaptionskip}{-10pt}
	\input{tikz/upright_angles.tex}
	\pgfplotsset{group/vertical sep=1cm}
	\caption{Experiment 2: Deviations from the upright orientation during the mug transfer after picking up the mug at $t = \SI{4.5}{\second}$. The two error angles $e_1$ and $e_2$ correspond to the two RPY-angles that constitute the deviation from the upright, see~\cite{oelerichBoundMPCCartesianPath2025} for the orientation error formulation. The gray areas are the constraints that change during the motion and define the set of permissible states $\mathcal{X}(\mathcal{L}_{t_{0}})$ that change at time $t_1$ to $\mathcal{X}(\mathcal{L}_{t_{1}, \mathrm{upright}})$ for the upright scenario. For the avoidance, the robot is aware of the upright constraint at all times.}
	\vspace{-2mm}
	\label{fig:scen_upright sets}
\end{figure}
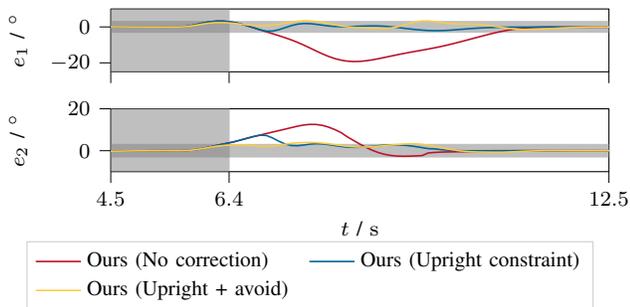

The utterance $\mathcal{L}_{t_{1}, \mathrm{upright}}$ avoids spillage on the laptop by keeping the mug upright. The resulting deviations from the upright orientation are described by the two error angles in~\cref{fig:scen_upright sets} for the time after the pickup, i.e., $t \geq \SI{4.5}{\second}$. The gray areas indicate the set of permissible deviations from the upright. Initially, the trajectory deviates largely from the upright as no constraint is specified. When the language specification $\mathcal{L}_{t_1, \mathrm{upright}}$ is taken into account at $t_1 = \SI{6.4}{\second}$, the upright constraint is added, changing $\mathcal{X}_\mathrm{task}$. At this point, the end effector has already deviated from the upright, which makes strictly enforcing $\vec{x}(t) \in \mathcal{X}(\mathcal{L}_{t_{1}, \mathrm{upright}})$ in~\cref{eq:motion_planning_problem} impossible. Therefore, the upright constraint is enforced with slack variables, effectively bringing the orientation into the desired bounds. Furthermore, the maximum deviation from the upright is at the time of replanning, meaning no further spilling happens afterward. This behavior is safe as the constraint is added before reaching the space over the laptop. Without upright constraints, the deviation from the upright increases further, leading to spillage onto the laptop.

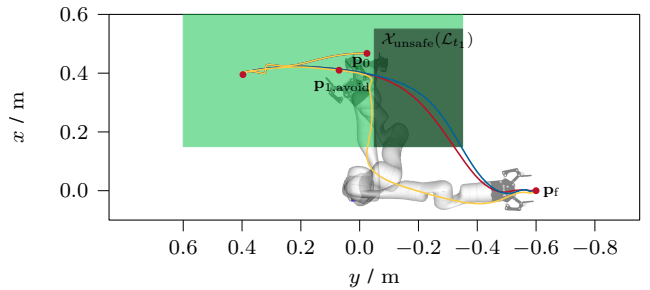
\begin{figure}
	\centering
	\def\axisdefaultwidth{\linewidth}
	\def\axisdefaultheight{0.5\linewidth}
	\addtolength{\abovecaptionskip}{-25pt}
	\input{tikz/avoid_sets_fix.tex}
	\caption{Experiment 2: Trajectories for the laptop avoidance during the mug transfer in the $x$-$y$-plane. The set of disallowed states $\mathcal{X}_\mathrm{unsafe}(\mathcal{L}_{t_{1}})$ is shaded and constitutes the space above the laptop.}
	\vspace{-1mm}
	\label{fig:scen_avoid_sets}
\end{figure}

As the upright constraint without consideration of the dynamics is only an approximation for avoiding spillage, the human decides in the second trial to specify $\mathcal{L}_{t_1, \mathrm{avoid}}$ for additional safety. In this scenario, the robot is aware that the mug should be upright during the entire transfer according to $\mathcal{L}_{t_{0}, \mathrm{avoid}}$, but the laptop avoidance is added online. The trajectories in the $x$-$y$-plane in~\cref{fig:scen_avoid_sets} show that the motion planner is able to adapt its trajectory online to account for the added task constraint. The replanning happens at pose $\vec{p}_{\mathrm{1, avoid}}$ shortly before the robot\textquotesingle s end effector reaches the space above the laptop. This requires a fast, safe, and decisive reaction from the robot. The added obstacle is only relevant to the robot\textquotesingle s end effector pose but does not affect the position of other parts of the kinematic chain.

\begin{figure}
	\centering
	\vspace{-10mm}
	\def\axisdefaultwidth{\linewidth}
	\def\axisdefaultheight{0.7\linewidth}
	\addtolength{\abovecaptionskip}{-15pt}
	\input{tikz/scen_avoid.tex}
	\caption{Experiment 2: The robot is visualized during the transfer of the mug at the initial pose $\vec{p}_{0}$ and the final pose $\vec{p}_{\mathrm{f}}$. The replanning for the upright constraint happens at pose $\vec{p}_{1, \mathrm{upright}}$, which is approximately at the same position as for the avoidance $\vec{p}_{1, \mathrm{avoid}}$. The black mug and the laptop rest on a green box.}
	\label{fig:scen_avoid}
\end{figure}
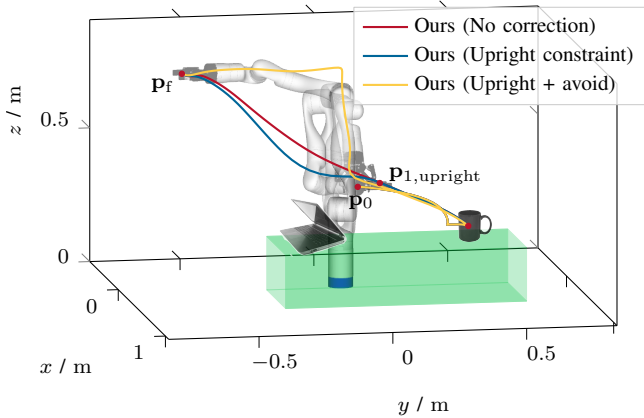

\subsection{Experiment 3: Handover of a screwdriver}

In a human-robot interaction scenario, the robot needs to be able to adapt to human preferences. This scenario is visualized in~\cref{fig:scen_handover} and assumes that a human works on a task for which a screwdriver is needed, but the screwdriver is not accessible to the human due to obstacles. The human instructs the robot with the utterance $\mathcal{L}_{t_0} = $ \textit{``Hand me the screwdriver''} at time $t_{0}$ to help with the task. The robot plans to pick up the screwdriver and provide it to the human at a predetermined handover location $\vec{p}_\mathrm{f}$. The human expects this task to take a certain time, but the robot moves too slowly to adhere to these expectations. Thus, at time $t_1$, the human specifies the utterance at time $t_1$ to $\mathcal{L}_{t_1} = $ \textit{``but move faster''}. This prompts the robot to change the cost function parameters $\phi(\mathcal{L}_{t_{i}})$ in~\cref{eq:motion_planning_problem} to value speed more strongly.

\begin{remark}
	Human-robot handovers are complicated interactions requiring many factors to be considered. This experiment simplifies this scenario by disregarding, e.g., human motion prediction and psychological factors, for exemplary reasons. A more rigorous approach is presented in~\cite{oelerichModelPredictiveTrajectory2024}.
\end{remark}

\begin{figure}
	\centering
	\def\axisdefaultwidth{0.95\linewidth}
	\def\axisdefaultheight{0.35\linewidth}
	\addtolength{\abovecaptionskip}{-5pt}
	\input{tikz/handover_speed_fix.tex}
	\caption{Experiment 3: Norm of the end-effector velocity $\vec{v}$ during the screwdriver handover with and without the speedup. The speedup is instructed at $t_1 = \SI{6.1}{\second}$ using $\mathcal{L}_{t_{1}}$.}
	\vspace{-3mm}
	\label{fig:scen_handover_speed}
\end{figure}
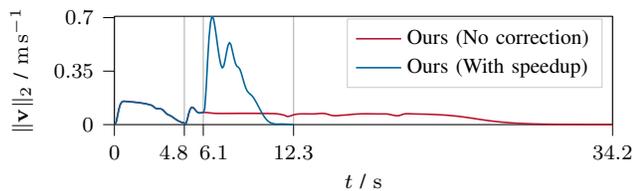

With and without the speedup instruction $\mathcal{L}_{t_1}$, the robot is able to finish the task and reach $\vec{p}_\mathrm{f}$ as seen in~\cref{fig:scen_handover}. However,
the robot takes considerably longer without the speedup, as shown by the norm of the Cartesian end effector velocity $\vec{v}$ in~\cref{fig:scen_handover_speed}. In both cases, the robot avoids obstacle collisions and safely reaches $\vec{p}_\mathrm{f}$. Thus, our framework enables online modifications of the objective to account for human preferences.

\begin{figure}
	\centering
	\def\axisdefaultwidth{\linewidth}
	\def\axisdefaultheight{0.7\linewidth}
	\addtolength{\abovecaptionskip}{-5pt}
	\input{tikz/scen_handover.tex}
	\caption{Experiment 3: The robot is visualized during the handover of the screwdriver at the initial pose $\vec{p}_{0}$, the pickup pose, and the handover pose $\vec{p}_{\mathrm{f}}$. The screwdriver rests on the green table and the robot has to avoid the black walls while transitioning to the handover location $\vec{p}_\mathrm{f}$.}
	\label{fig:scen_handover}
\end{figure}
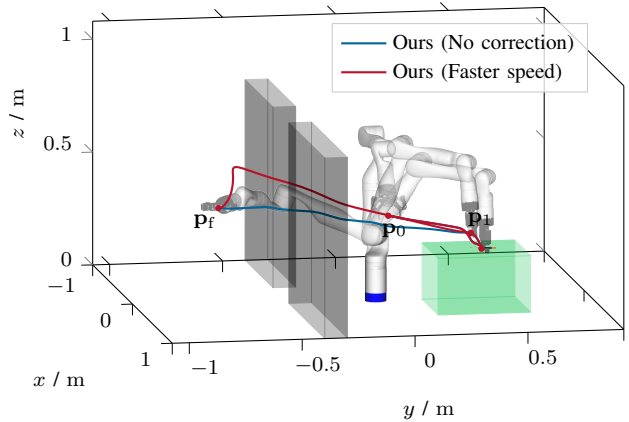

\section{CONCLUSION}

This paper presents an incremental language processing system integrated with an online motion planner for real-time human-robot interaction. Our approach enables a robotic agent to dynamically interpret linguistic instructions, continuously update motion plans, and refine its actions based on evolving constraints, unlike previous methods that assume fully specified language input. The system adapts to speech corrections and new task constraints in real time, such as modifying grasp strategies or avoiding obstacles, reflecting more naturalistic communication and interactions. The effectiveness of our approach was demonstrated in the real world using a 7-DoF robot manipulator. Future work will focus on extending the system to incorporate visual information, refining adaptation mechanisms to support more complex task hierarchies, and conducting a thorough human subject evaluation study.





\addtolength{\textheight}{-2cm}   


\bibliographystyle{IEEEtran}
\bibliography{IEEEabrv,motion_planning}  
\end{document}

%% file: tex_packages.tex
\usepackage{moreverb,url}
\usepackage{pgfplots}
\usepackage{pgfplotstable}
\usepgfplotslibrary{groupplots}
\usepackage{amsmath}
\usepackage{amsfonts}
\usepackage{amsthm}
\usepackage{import}
\usepackage{siunitx}
\usepackage{cleveref}
\usepackage{array}

%% file: tex_settings.tex
\crefformat{figure}{Fig.~#2#1#3}
\Crefformat{figure}{Figure~#2#1#3}
\crefformat{equation}{(#2#1#3)}
\crefformat{section}{Section~#2#1#3}
\crefformat{table}{Table~#2#1#3}
\crefformat{algorithm}{Algorithm~#2#1#3}
\crefformat{appendix}{Appendix~#2#1#3}

\DeclareRobustCommand{\vec}[1]{
	\ifthenelse{\equal{#1}{\omega} \OR \equal{#1}{\varphi} \OR \equal{#1}{\alpha} \OR \equal{#1}{\beta} \OR \equal{#1}{\chi} \OR \equal{#1}{\delta} \OR \equal{#1}{\varepsilon} \OR \equal{#1}{\phi} \OR \equal{#1}{\epsilon} \OR \equal{#1}{\gamma} \OR \equal{#1}{\eta} \OR \equal{#1}{\iota} \OR \equal{#1}{\kappa} \OR \equal{#1}{\lambda} \OR \equal{#1}{\mu} \OR \equal{#1}{\nu} \OR \equal{#1}{\pi} \OR \equal{#1}{\theta} \OR \equal{#1}{\vartheta} \OR \equal{#1}{\rho} \OR \equal{#1}{\sigma} \OR \equal{#1}{\varsigma} \OR \equal{#1}{\tau} \OR \equal{#1}{\upsilon} \OR \equal{#1}{\xi} \OR \equal{#1}{\psi} \OR \equal{#1}{\zeta}}{
		\boldsymbol{#1}
	}{
		\mathbf{#1}
	}
}

\usetikzlibrary {3d}
\usetikzlibrary{decorations.markings}
\usepgfplotslibrary{fillbetween}
\usepgfplotslibrary{patchplots}
\usetikzlibrary{external}
\usetikzlibrary {arrows.meta}
\usetikzlibrary{patterns,patterns.meta}
\pgfplotsset{every axis/.append style={
			font=\footnotesize,
			line join=round,
			legend style={/tikz/every even column/.append style={column sep=0.2cm}}
		}}

\DeclareRobustCommand{\qs}[1]{\textsinglequote s}

\renewcommand{\dddot}[1]{%
	{\mathop{\kern\z@#1}\limits^{\makebox[0pt][c]{\vbox to-1.4\ex@{\kern-\tw@\ex@
						\hbox{\normalfont...}\vss}}}}}
\newcommand{\dddotsuper}[1]{%
	{\mathop{\kern\z@#1}\limits^{\makebox[0pt][c]{\vbox to-1.6\ex@{\kern-\tw@\ex@
						\hbox{\scriptsize...}\vss}}}}}
\newcommand{\dddotalt}[1]{%
	{\mathop{\kern\z@#1}\limits^{\makebox[0pt][c]{\vbox to-2.2\ex@{\kern-\tw@\ex@
						\hbox{\normalfont\scaleddot\kern-0.5pt\scaleddot\kern-0.5pt\scaleddot}\vss}}}}}
\makeatother

%% file: inkscape/schematic.pdf_tex
\begingroup%
  \makeatletter%
  \providecommand\color[2][]{%
    \errmessage{(Inkscape) Color is used for the text in Inkscape, but the package 'color.sty' is not loaded}%
    \renewcommand\color[2][]{}%
  }%
  \providecommand\transparent[1]{%
    \errmessage{(Inkscape) Transparency is used (non-zero) for the text in Inkscape, but the package 'transparent.sty' is not loaded}%
    \renewcommand\transparent[1]{}%
  }%
  \providecommand\rotatebox[2]{#2}%
  \newcommand*\fsize{\dimexpr\f@size pt\relax}%
  \newcommand*\lineheight[1]{\fontsize{\fsize}{#1\fsize}\selectfont}%
  \ifx\svgwidth\undefined%
    \setlength{\unitlength}{392.26485605bp}%
    \ifx\svgscale\undefined%
      \relax%
    \else%
      \setlength{\unitlength}{\unitlength * \real{\svgscale}}%
    \fi%
  \else%
    \setlength{\unitlength}{\svgwidth}%
  \fi%
  \global\let\svgwidth\undefined%
  \global\let\svgscale\undefined%
  \makeatother%
  \begin{picture}(1,0.59852113)%
    \lineheight{1}%
    \setlength\tabcolsep{0pt}%
    \put(0,0){\includegraphics[width=\unitlength,page=1]{schematic.pdf}}%
  \end{picture}%
\endgroup%

%% file: tikz/scen_mug_graps.tex
\begin{tikzpicture}
	\definecolor{darkgray176}{RGB}{176,176,176}
	\definecolor{acinblue}{RGB}{0,102,153}
	\definecolor{acinyellow}{RGB}{252, 204, 71}
	\definecolor{acingreen}{RGB}{0,190,65}
	\definecolor{acinred}{RGB}{186,18,43}
	\definecolor{lightgray204}{RGB}{204,204,204}

	\node[inner sep=0pt, opacity=0.5] (qf) at (3.73,2.75)
	{\includegraphics[width=.305\textwidth]{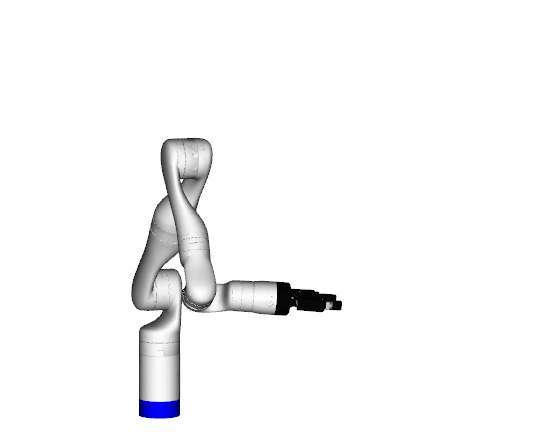}};
	\node[inner sep=0pt, opacity=0.5] (qf) at (3.73,2.75)
	{\includegraphics[width=.305\textwidth]{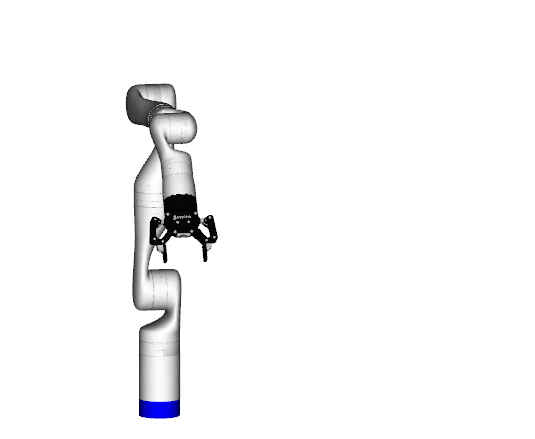}};
	\node[inner sep=0pt, opacity=0.5] (qf) at (3.73,2.75)
	{\includegraphics[width=.305\textwidth]{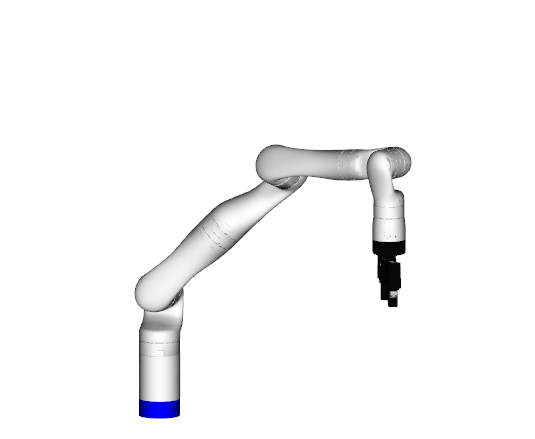}};
	\begin{axis}[
			view={80}{10},
			xlabel={$x$ / \si{\meter}},
			ylabel={$y$ / \si{\meter}},
			zlabel={$z$ / \si{\meter}},
			zmin=0,
			axis equal,
			tick align=outside,
			tick pos=left,
			x grid style={darkgray176},
			y grid style={darkgray176},
			xtick style={color=black},
			ytick style={color=black},
			legend cell align={left},
			legend columns=1,
			legend style={
					fill opacity=0.8,
					draw opacity=1,
					text opacity=1,
					draw=lightgray204,
					at={(0.03,0.8)},
					anchor=west,
				},
			legend entries ={Ours, VP-STO}
		]
		\pgfplotstableread{data/p_traj_mug_grasp.txt}\ptraj;
		\pgfplotstableread{data/r_traj_mug_grasp.txt}\rtraj;

		\pgfplotstableread{data/p_vpsto_mug_grasp.txt}\pvpsto;

		\addplot3 [acinred, thick]
		table [
				x expr=\thisrowno{0},
				y expr=\thisrowno{1},
				z expr=\thisrowno{2}
			] {\ptraj};
		\addplot3 [acinblue, thick]
		table [
				x expr=\thisrowno{0},
				y expr=\thisrowno{1},
				z expr=\thisrowno{2}
			] {\pvpsto};
		%
		%
		%
		%

		\pgfplotstablegetrowsof{\ptraj} 

		\pgfmathsetmacro{\totalrows}{\pgfplotsretval-1}

		\addplot3 [patch,patch type=rectangle, patch refines=0, shader=flat, opacity=0.25, color=acingreen] coordinates
			{
				(0.1,-0.3,0.0) (0.5,-0.3,0.0) (0.5,0.6,0.0) (0.1,0.6,0.0)
				(0.1,-0.3,0.2) (0.5,-0.3,0.2) (0.5,0.6,0.2) (0.1,0.6,0.2)
				(0.1,-0.3,0.0) (0.1,-0.3,0.2)  (0.1,0.6,0.2) (0.1,0.6,0.0)
				(0.5,-0.3,0.0) (0.5,-0.3,0.2)  (0.5,0.6,0.2) (0.5,0.6,0.0)
			};
		\node[inner sep=0pt, opacity=0.5] (q0) at (axis cs: 0.4,0.47,0.28)
		{\includegraphics[width=.03\textwidth]{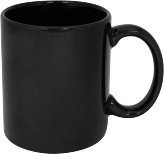}};
		\addplot3[
			acinred,
			only marks,
			mark=*,
			mark size=1pt,
		] coordinates {
				(4.673250006555440539e-01, -2.485086265104662073e-02, 4.406266176809824353e-01)
				(4.001223247977102603e-01, 4.501343020825520069e-01, 3.2e-01)
				(3.980007300263146974e-01, 3.343721071331716699e-01, 2.983243975555132653e-01)
				(0.40002527, 0.45001845, 0.27003069)
			};
		\draw (axis cs:0.5,0.44,0.35) node[
		scale=1.0,
		anchor=base west,
		text=black,
		rotate=0.0
		]{\bfseries $\vec{p}_{\mathrm{f}}$};
		\draw (axis cs:0.5,-0.1,0.4) node[
		scale=1.0,
		anchor=base west,
		text=black,
		rotate=0.0
		]{\bfseries $\vec{p}_{\mathrm{0}}$};
		\draw (axis cs:0.5,0.20,0.32) node[
		scale=1.0,
		anchor=base west,
		text=black,
		rotate=0.0
		]{\bfseries $\vec{p}_{\mathrm{1}}$};
		\draw (axis cs:0.5,0.40,0.23) node[
		scale=1.0,
		anchor=base west,
		text=black,
		rotate=0.0
		]{\bfseries $\vec{p}_{\mathrm{1, VP-STO}}$};
	\end{axis}
\end{tikzpicture}

%% file: tikz/mug_speed.tex
\begin{tikzpicture}

	\definecolor{darkgrey176}{RGB}{176,176,176}
	\definecolor{firebrick1861843}{RGB}{186,18,43}
	\definecolor{lightgrey204}{RGB}{204,204,204}
	\definecolor{teal0101153}{RGB}{0,101,153}

	\begin{axis}[
			legend cell align={left},
			legend style={
					fill opacity=0.8,
					draw opacity=1,
					text opacity=1,
					at={(0.65,0.3)},
					anchor=south,
					draw=lightgrey204
				},
			tick align=outside,
			tick pos=left,
			x grid style={darkgrey176},
			xlabel={\(\displaystyle t\) / \si{\second}},
			xmajorgrids,
			xmin=0, xmax=17.072,
			xtick style={color=black},
			xtick={0,3.1,7.800000117,5.58,14.099,17.072},
			xticklabels={0.0,3.1,7.8,5.6,14.1,17.1},
			y grid style={darkgrey176},
			ylabel={\(\displaystyle \lVert \dot{\vec{q}} \rVert_2\) / \si{\degree\per\second}},
			ymin=0, ymax=116.769216622035,
			ytick style={color=black},
			ytick={0,50,100},
			yticklabels={\(\displaystyle 0\),\(\displaystyle 50\),\(\displaystyle 100\)}
		]
		\addplot [semithick, firebrick1861843]
		table {%
				0 1.63827532752781e-05
				0.100000002 2.82854895196794
				0.200000003 17.1940377952261
				0.300000005 34.7363549305031
				0.400000006 49.344431025931
				0.500000008 59.9152429240015
				0.600000009 67.4558949456208
				0.700000011 74.3936893509226
				0.800000012 81.2683644196065
				0.900000014 87.1817816805301
				1.000000015 91.3240026923607
				1.100000017 93.4303166208687
				1.200000018 93.6847951448315
				1.30000002 92.6248749726907
				1.400000021 91.051679843393
				1.500000023 89.5725145338831
				1.600000024 88.3824016748467
				1.700000026 87.455858490429
				1.800000027 86.561781038774
				1.900000029 85.4095668495613
				2.00000003 83.7953764666737
				2.100000032 81.6293526985567
				2.200000033 78.8956308386883
				2.300000035 75.6387713829543
				2.400000036 71.9501970024223
				2.500000038 67.9007603638074
				2.600000039 63.39128678504
				2.700000041 58.3511539530311
				2.800000042 52.8870610823573
				2.900000044 47.1930032439267
				3.000000045 41.4225165087518
				3.100000047 33.158635629407
				3.200000048 19.7289088918471
				3.30000005 31.8754330474131
				3.400000051 56.1839536705913
				3.500000053 76.8982138203162
				3.600000054 91.2937686550253
				3.700000056 99.370931601883
				3.800000057 102.043811264505
				3.900000059 100.746253237778
				4.00000006 97.1278634114421
				4.100000062 92.8340555155279
				4.200000063 89.0265339910168
				4.300000065 86.542806241975
				4.400000066 85.8579576535926
				4.500000067 86.8585929133237
				4.600000069 88.5537853329859
				4.70000007 89.8299303645129
				4.800000072 89.8677775548719
				4.900000073 88.2153592991571
				5.000000075 84.7367195531332
				5.100000076 79.4885367981036
				5.200000078 72.6340008914807
				5.300000079 64.4331725064641
				5.400000081 55.2547224600751
				5.500000082 45.5707914016489
				5.600000084 35.9299848622291
				5.700000085 26.9437563235444
				5.800000087 19.0727439320042
				5.900000088 12.5557289881129
				6.00000009 7.81697180040749
				6.100000091 5.85777231890881
				6.200000093 6.75163159428024
				6.300000094 8.42404888150841
				6.400000096 9.73556852988985
				6.500000097 10.4392729375847
				6.600000099 10.5508516998162
				6.7000001 10.1694271877093
				6.800000102 9.41634603060887
				6.900000103 8.41031209572915
				7.000000105 7.25715337660528
				7.100000106 6.0469830591282
				7.200000108 4.85101481843774
				7.300000109 3.7239059529776
				7.400000111 2.70661375233199
				7.500000112 1.83037841084709
				7.600000114 1.12721327937502
				7.700000115 0.664377567086481
				7.800000117 0.569500651176727
			};
		\addlegendentry{Ours}
		\addplot [semithick, teal0101153]
		table {%
				0 0
				5.58 1.01825959552757e-13
				5.68 34.9621931104309
				5.78 60.7353199958969
				5.88 78.0985441219015
				5.98 88.4441660853469
				6.08 94.3082331911727
				6.18 100.04946140949
				6.28 107.687761665426
				6.38 113.899830102349
				6.48 116.769216622035
				6.58 115.69464053842
				6.68 111.058936518208
				6.78 104.785488423647
				6.88 101.669702710033
				6.98 101.163073221502
				7.08 101.7917705334
				7.18 102.58034275686
				7.28 103.121376503472
				7.38 103.351331211014
				7.48 102.0243747198
				7.58 97.977853588574
				7.68 90.780818700827
				7.78 80.7542802888777
				7.88 69.4832994427262
				7.98 60.729441232029
				8.08 54.4163262030175
				8.18 47.1286785748021
				8.28 36.4572698847232
				8.38 20.9659100368359
				8.479 0.0756599560206541
				14.099 1.08599472056785e-13
				14.199 21.7171842362527
				14.299 37.5906771799672
				14.399 48.6015582221768
				14.499 56.4570378568113
				14.599 63.9396641711475
				14.699 74.7918384871189
				14.799 88.02804341025
				14.899 97.9971975258169
				14.999 102.895727766501
				15.099 102.392177650574
				15.199 97.1036893964007
				15.299 88.9782309000436
				15.399 84.0471716498023
				15.499 82.8562954782476
				15.599 83.129071583209
				15.699 84.0585112038609
				15.799 86.4516533816255
				15.899 92.3919236199284
				15.999 98.6187133604581
				16.099 100.932918489186
				16.199 98.4239094237796
				16.299 91.0975861406042
				16.399 80.0234110363734
				16.499 68.575187864603
				16.599 62.6306164532937
				16.699 58.5904470182848
				16.799 51.6334008669337
				16.899 39.034777249519
				16.999 19.4030653721291
				17.072 0.201689153828768
			};
		\addlegendentry{VP-STO}
	\end{axis}

\end{tikzpicture}

%% file: tikz/mug_sets_fix.tex
\begin{tikzpicture}

	\definecolor{darkgrey176}{RGB}{176,176,176}
	\definecolor{firebrick1861843}{RGB}{186,18,43}
	\definecolor{limegreen019166}{RGB}{0,191,66}
	\definecolor{sandybrown25220370}{RGB}{252,203,70}
	\definecolor{teal0101153}{RGB}{0,101,153}

	\node[inner sep=0pt, opacity=0.5] (qf) at (4.35,1.4)
	{\includegraphics[width=.23\textwidth]{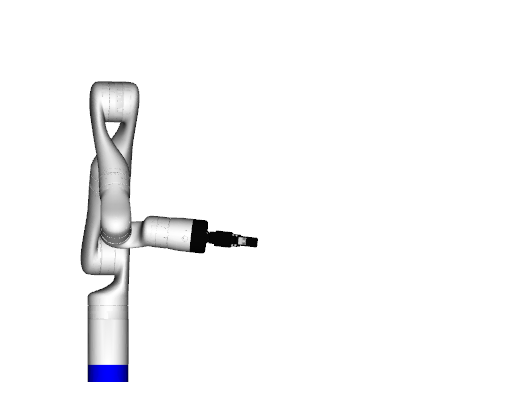}};
	\node[inner sep=0pt, opacity=0.5] (q0) at (4.35,1.4)
	{\includegraphics[width=.23\textwidth]{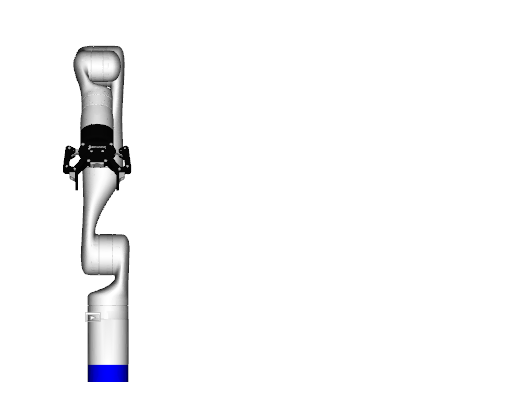}};
	\node[inner sep=0pt, opacity=0.5] (q0) at (4.35,1.4)
	{\includegraphics[width=.23\textwidth]{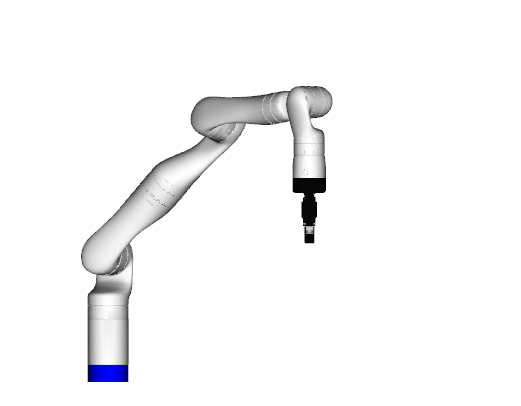}};
	\node[inner sep=0pt, opacity=0.8] (q0) at (4.8,0.9)
	{\includegraphics[width=.025\textwidth]{data/mug.png}};

	\begin{axis}[
			tick align=outside,
			tick pos=left,
			xlabel={$y$ / \si{\meter}},
			ylabel={$z$ / \si{\meter}},
			x grid style={darkgrey176},
			xmin=-0.4, xmax=0.6,
			axis equal,
			xtick style={color=black},
			y grid style={darkgrey176},
			ymin=0, ymax=0.8,
			ytick style={color=black},
		]
		\path [draw=limegreen019166, fill=limegreen019166, opacity=0.5]
		(axis cs:-0.3,0.2)
		--(axis cs:-0.3,0)
		--(axis cs:0.6,0)
		--(axis cs:0.6,0.2)
		--(axis cs:0.6,0.2)
		--(axis cs:-0.3,0.2)
		--cycle;

		\path [draw=teal0101153, fill=teal0101153, opacity=0.5]
		(axis cs:-1,1)
		--(axis cs:-1,0.26)
		--(axis cs:0.36,0.26)
		--(axis cs:0.36,1)
		--(axis cs:0.36,1)
		--(axis cs:-1,1)
		--cycle;

		\path [draw=teal0101153, fill=teal0101153, opacity=0.5]
		(axis cs:0.25,0.274)
		--(axis cs:0.25,0.266)
		--(axis cs:0.45,0.266)
		--(axis cs:0.45,0.274)
		--(axis cs:0.45,0.274)
		--(axis cs:0.25,0.274)
		--cycle;

		\path [draw=sandybrown25220370, opacity=0.5, pattern={Lines[angle=45,distance=6pt,line width=3pt]}, pattern color=sandybrown25220370]
		(axis cs:-1,1)
		--(axis cs:-1,0.26)
		--(axis cs:0.36,0.26)
		--(axis cs:0.36,1)
		--(axis cs:0.36,1)
		--(axis cs:-1,1)
		--cycle;

		\path [draw=sandybrown25220370, fill=sandybrown25220370, opacity=0.5]
		(axis cs:0.36,1)
		--(axis cs:0.36,0.38)
		--(axis cs:1,0.38)
		--(axis cs:1,1)
		--(axis cs:1,1)
		--(axis cs:0.36,1)
		--cycle;

		\path [draw=sandybrown25220370, fill=sandybrown25220370, opacity=0.5]
		(axis cs:0.446,0.5)
		--(axis cs:0.446,0.31)
		--(axis cs:0.454,0.31)
		--(axis cs:0.454,0.5)
		--(axis cs:0.454,0.5)
		--(axis cs:0.446,0.5)
		--cycle;

		\addplot [teal0101153]
		table {%
				0.36 0.26
				-1 0.260000000000007
			};
		\addplot [teal0101153]
		table {%
				0.36 0.26
				0.36 1.2
			};
		\addplot [teal0101153]
		table {%
				-1 1.2
				-1 0.260000000000007
			};
		\addplot [teal0101153]
		table {%
				-1 1.2
				0.36 1.2
			};
		\addplot [teal0101153]
		table {%
				0.36 0.26
				-1 0.26
			};
		\addplot [teal0101153]
		table {%
				0.36 1.2
				0.36 0.26
			};
		\addplot [teal0101153]
		table {%
				-1 1.2
				-1 0.26
			};
		\addplot [teal0101153]
		table {%
				-1 1.2
				0.36 1.2
			};
		\addplot [sandybrown25220370, dashed]
		table {%
				0.36 0.26
				-1 0.26
			};
		\addplot [sandybrown25220370, dashed]
		table {%
				0.36 1.2
				0.36 0.26
			};
		\addplot [sandybrown25220370, dashed]
		table {%
				-1 1.2
				-1 0.26
			};
		\addplot [sandybrown25220370, dashed]
		table {%
				-1 1.2
				0.36 1.2
			};
		\addplot [sandybrown25220370]
		table {%
				1 0.38
				-1 0.38
			};
		\addplot [sandybrown25220370]
		table {%
				1 1.2
				1 0.38
			};
		\addplot [sandybrown25220370]
		table {%
				-1 1.2
				-1 0.38
			};
		\addplot [sandybrown25220370]
		table {%
				-1 1.2
				1 1.2
			};
		\addplot [semithick, sandybrown25220370]
		table {%
				0.446 0.31
				0.454 0.31
			};
		\addplot [semithick, sandybrown25220370]
		table {%
				0.454 0.31
				0.454 0.5
			};
		\addplot [semithick, sandybrown25220370]
		table {%
				0.454 0.5
				0.446 0.5
			};
		\addplot [semithick, sandybrown25220370]
		table {%
				0.446 0.5
				0.446 0.31
			};
		\addplot [semithick, teal0101153]
		table {%
				0.25 0.266
				0.45 0.266
			};
		\addplot [semithick, teal0101153]
		table {%
				0.45 0.266
				0.45 0.274
			};
		\addplot [semithick, teal0101153]
		table {%
				0.45 0.274
				0.25 0.274
			};
		\addplot [semithick, teal0101153]
		table {%
				0.25 0.274
				0.25 0.266
			};
		\addplot [semithick, firebrick1861843]
		table {%
				-0.0248508636635696 0.44062661754965
				-0.0248508770242424 0.440626639581166
				-0.0243395233286825 0.441019632210177
				-0.0177017773873823 0.446078982812125
				0.000322391040373 0.459328116976795
				0.0275231847773597 0.478304110001475
				0.0588271393035783 0.498577407563411
				0.0895402565762451 0.516052331406141
				0.115765421427154 0.527355093828053
				0.135365092525097 0.53076266278826
				0.148750828319626 0.526849282883206
				0.15741207389866 0.517293703759177
				0.162909937527035 0.503978824059183
				0.166480424846642 0.488544281614852
				0.16900835299879 0.472260547861022
				0.171174737199119 0.456092987777417
				0.173642972982439 0.440840927981245
				0.177156577295009 0.427221434761162
				0.18243147869934 0.415731271028876
				0.189855872786135 0.406325871795839
				0.199328985159517 0.398443539304808
				0.210401915757607 0.391378465271992
				0.222504737385671 0.384531677660818
				0.23511418248871 0.377505384276244
				0.247831604458976 0.370108168907364
				0.260392508118025 0.362309832764552
				0.272628140923089 0.354182044048529
				0.284330434754529 0.345830645176107
				0.295231041176503 0.337377420293333
				0.305110859751858 0.328945864277412
				0.313830032572041 0.320624388473408
				0.321338775352772 0.312502525469235
				0.327585555314135 0.304643206049774
				0.331871151635724 0.296565452841037
				0.334242212666474 0.289202341716312
				0.335688584947472 0.285431848381746
				0.336887383919519 0.28728767218697
				0.338332452802003 0.295094996400194
				0.3402623600226 0.307633556370307
				0.342638539612531 0.322915489539112
				0.345246134744919 0.339032349938172
				0.347863162005644 0.354657038111296
				0.350412893895542 0.369088254606688
				0.353049654839844 0.381989428924308
				0.356115747676905 0.393034825166641
				0.35999674245367 0.401710283036819
				0.365059511651598 0.407424757148781
				0.371666502044729 0.410081278159271
				0.37995960378322 0.410172908564474
				0.389648546416345 0.408318194702375
				0.400138525524601 0.405106559051497
				0.410751170128475 0.401110775573284
				0.420852090433351 0.39687875193065
				0.429914559719505 0.392852129253842
				0.437587108146214 0.389288775556594
				0.443728485205591 0.386204873776853
				0.448323261001394 0.383310424744778
				0.451286135423182 0.379999991876898
				0.452536483164012 0.375636793479428
				0.45253741508002 0.370163799698132
				0.452155877446195 0.364132835296232
				0.451946244035411 0.358159113046107
				0.452009796828119 0.352650421481423
				0.452211942319664 0.347781431916381
				0.452384550175075 0.343559811047602
				0.452414301862019 0.339915420998018
				0.452261693282307 0.336765141429546
				0.451944687715475 0.334041532592315
				0.451510704318334 0.331696086821959
				0.451014909562979 0.329691884346872
				0.450507311190351 0.327995730462224
				0.450027388556556 0.326573856293144
				0.449603047298418 0.325391368902758
				0.44925192152792 0.324413709198614
				0.448983355739803 0.323608513074542
				0.448799781857114 0.322947079735716
				0.448698001679517 0.322404959975177
				0.448670425471213 0.321961806873306
				0.448706261998517 0.321600829033584
				0.448792782689464 0.321308109070055
			};
		\addplot [semithick, firebrick1861843, mark=*, mark size=1, mark options={solid}, only marks]
		table {%
				0.448792782689464 0.321308109070055
			};
		\addplot [semithick, firebrick1861843, mark=*, mark size=1, mark options={solid}, only marks]
		table {%
				-0.0248508636635696 0.44062661754965
			};
		\addplot [semithick, firebrick1861843, mark=*, mark size=1, mark options={solid}, only marks]
		table {%
				0.321338775352772 0.312502525469235
			};
		\draw (axis cs:0.0,0.44062661754965) node[
			scale=0.75,
			anchor=base west,
			text=black,
			rotate=0.0
		]{\bfseries $\vec{p}_{0}$};
		\draw (axis cs:0.498792782689464,0.321308109070055) node[
			scale=0.75,
			anchor=base west,
			text=black,
			rotate=0.0
		]{\bfseries $\vec{p}_\mathrm{f}$};
		\draw (axis cs:0.27,0.22) node[
			scale=0.75,
			anchor=base west,
			text=black,
			rotate=0.0
		]{\bfseries $\vec{p}_\mathrm{1}$};
		\draw (axis cs:-0.5,0.3) node[
			scale=0.75,
			anchor=base west,
			text=black,
			rotate=0.0
		]{\bfseries $\mathcal{X}(\mathcal{L}_{t_0})$};
		\draw (axis cs:0.6,0.6) node[
			scale=0.75,
			anchor=base west,
			text=black,
			rotate=0.0
		]{\bfseries $\mathcal{X}(\mathcal{L}_{t_1})$};
	\end{axis}

\end{tikzpicture}

%% file: tikz/scen_avoid.tex
\begin{tikzpicture}
	\definecolor{darkgray176}{RGB}{176,176,176}
	\definecolor{acinblue}{RGB}{0,102,153}
	\definecolor{acinyellow}{RGB}{252, 204, 71}
	\definecolor{acingreen}{RGB}{0,190,65}
	\definecolor{acinred}{RGB}{186,18,43}
	\definecolor{lightgray204}{RGB}{204,204,204}

	\node[inner sep=0pt, opacity=0.5] (qf) at (4.6,3.15)
	{\includegraphics[width=.435\textwidth]{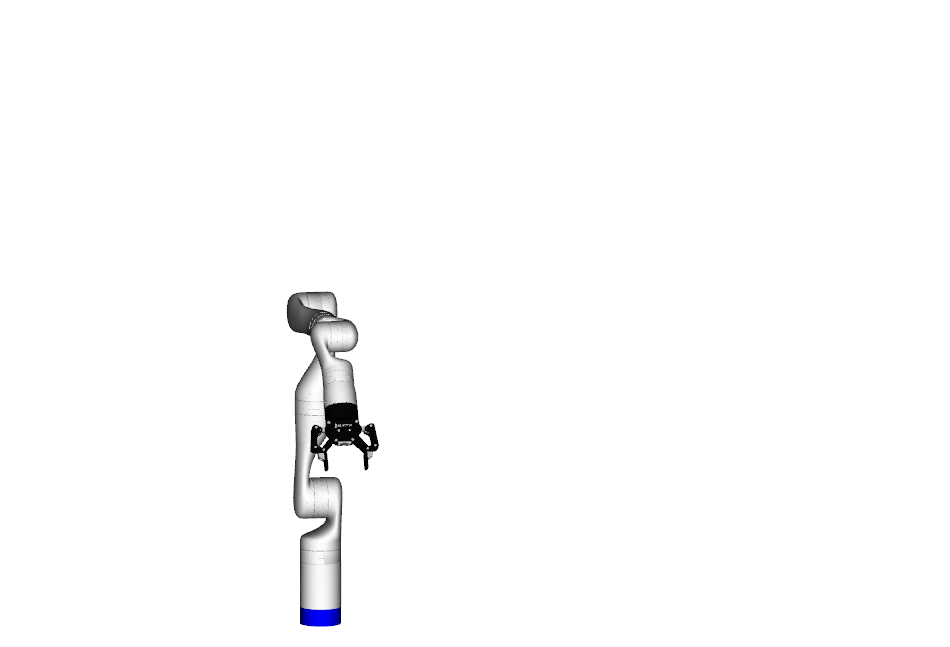}};
	\node[inner sep=0pt, opacity=0.5] (qf) at (4.6,3.15)
	{\includegraphics[width=.435\textwidth]{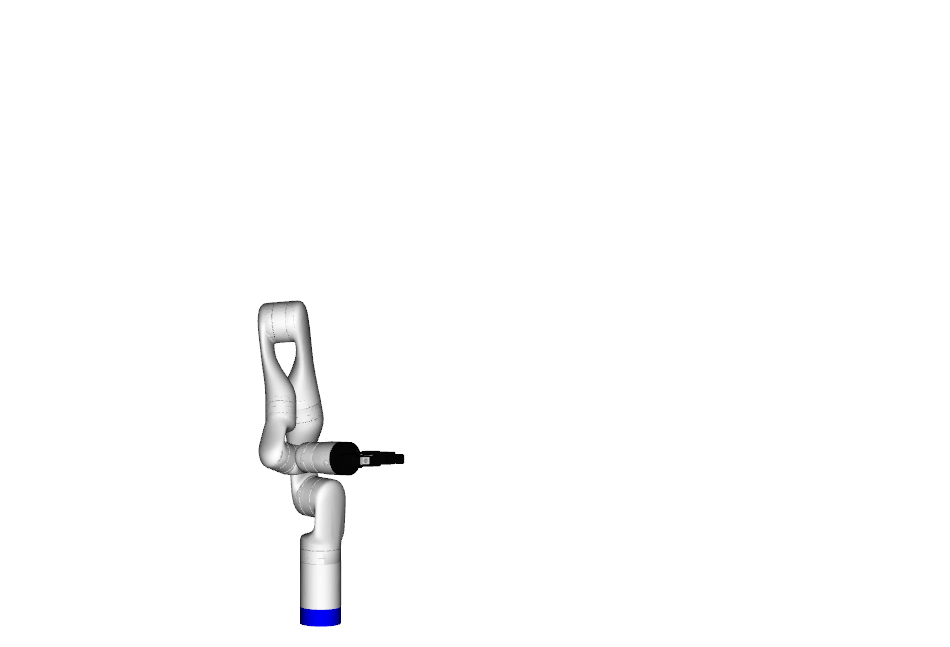}};
	\node[inner sep=0pt, opacity=0.5] (qf) at (4.6,3.15)
	{\includegraphics[width=.435\textwidth]{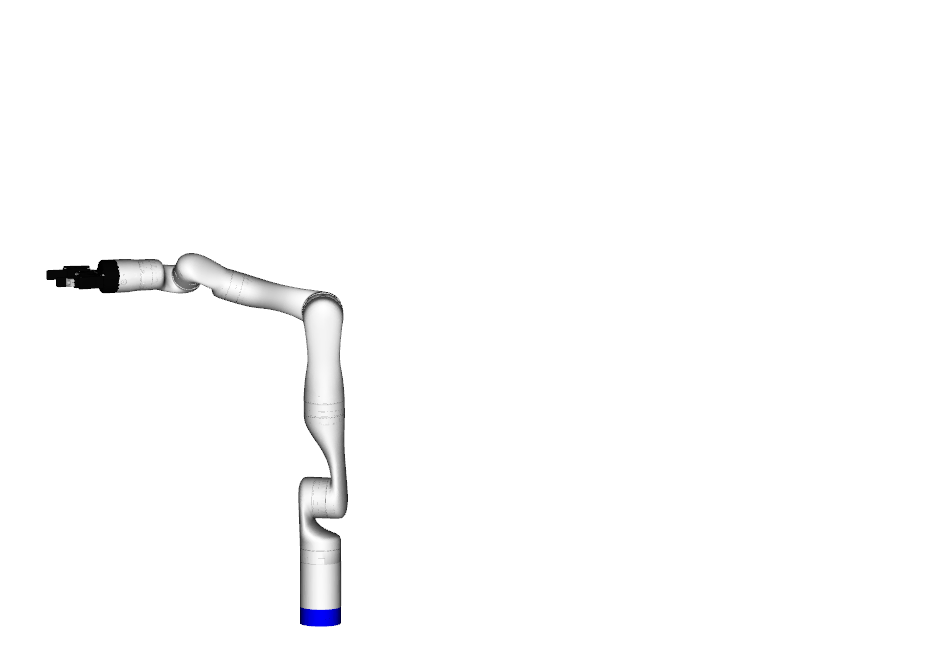}};
	\begin{axis}[
			view={80}{10},
			xlabel={$x$ / \si{\meter}},
			ylabel={$y$ / \si{\meter}},
			zlabel={$z$ / \si{\meter}},
			zmin=0,
			axis equal,
			tick align=outside,
			tick pos=left,
			x grid style={darkgray176},
			y grid style={darkgray176},
			xtick style={color=black},
			ytick style={color=black},
			legend cell align={left},
			legend columns=1,
			legend style={
					fill opacity=0.8,
					draw opacity=1,
					text opacity=1,
					draw=lightgray204,
					at={(0.5,0.85)},
					anchor=west,
				},
			legend entries ={Ours (No correction), Ours (Upright constraint), Ours (Upright + avoid)}
		]
		\addplot3 [patch,patch type=rectangle, patch refines=0, shader=flat, opacity=0.25, color=acingreen, forget plot] coordinates
			{
				(0.1,-0.3,0.0) (0.5,-0.3,0.0) (0.5,0.6,0.0) (0.1,0.6,0.0)
				(0.1,-0.3,0.2) (0.5,-0.3,0.2) (0.5,0.6,0.2) (0.1,0.6,0.2)
				(0.1,-0.3,0.0) (0.1,-0.3,0.2)  (0.1,0.6,0.2) (0.1,0.6,0.0)
				(0.5,-0.3,0.0) (0.5,-0.3,0.2)  (0.5,0.6,0.2) (0.5,0.6,0.0)
			};
		\node[inner sep=0pt, opacity=0.8] (q0) at (axis cs:0.4,0.425,0.27)
		{\includegraphics[width=.025\textwidth]{data/mug.png}};

		\pgfplotstableread{data/p_traj_avoid.txt}\ptraj;
		\pgfplotstableread{data/r_traj_avoid.txt}\rtraj;

		\pgfplotstableread{data/p_no_avoid_avoid.txt}\pnoavoid;
		\pgfplotstableread{data/p_upright_avoid.txt}\pupright;

		\addplot3 [acinred, thick]
		table [
				x expr=\thisrowno{0},
				y expr=\thisrowno{1},
				z expr=\thisrowno{2}
			] {\pnoavoid};
		\addplot3 [acinblue, thick]
		table [
				x expr=\thisrowno{0},
				y expr=\thisrowno{1},
				z expr=\thisrowno{2}
			] {\pupright};
		\addplot3 [acinyellow, thick]
		table [
				x expr=\thisrowno{0},
				y expr=\thisrowno{1},
				z expr=\thisrowno{2}
			] {\ptraj};
		%
		%
		%
		%

		\pgfplotstablegetrowsof{\ptraj} 

		\pgfmathsetmacro{\totalrows}{\pgfplotsretval-1}

		\addplot3[
			acinred,
			only marks,
			mark=*,
			mark size=1pt,
		] coordinates {
				(4.673250006555440539e-01, -2.485086265104662073e-02, 4.406266176809824353e-01)
				(0.51, 0.05, 0.46)
				(0.40002527, 0.40000, 0.27003069)
				(-9.148345809389030780e-07, -5.998314186791845470e-01, 0.8)
			};
		\draw (axis cs:0.0,-0.75,0.75) node[
		scale=1.0,
		anchor=base west,
		text=black,
		rotate=0.0
		]{\bfseries $\vec{p}_{\mathrm{f}}$};
		\draw (axis cs:0.5,-0.1,0.38) node[
		scale=1.0,
		anchor=base west,
		text=black,
		rotate=0.0
		]{\bfseries $\vec{p}_{\mathrm{0}}$};
		\draw (axis cs:0.5,0.05,0.48) node[
		scale=1.0,
		anchor=base west,
		text=black,
		rotate=0.0
		]{\bfseries $\vec{p}_{\mathrm{1, upright}}$};
	\end{axis}

	\node[inner sep=0pt, opacity=0.8] (q0) at (3.1,1.55)
	{\includegraphics[width=.08\textwidth]{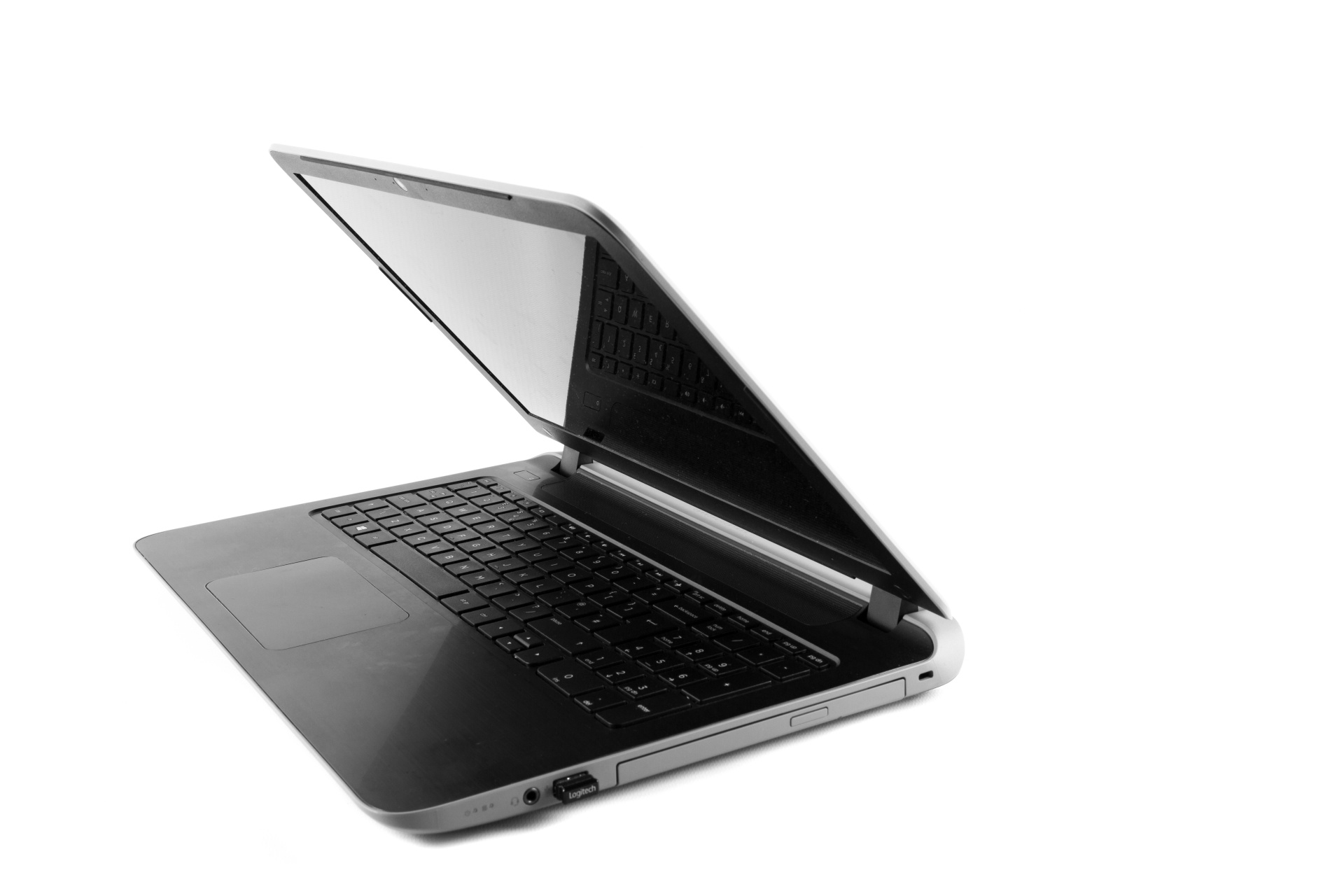}};
\end{tikzpicture}

%% file: tikz/upright_angles.tex
\begin{tikzpicture}

	\definecolor{darkgrey176}{RGB}{176,176,176}
	\definecolor{firebrick1861843}{RGB}{186,18,43}
	\definecolor{grey}{RGB}{128,128,128}
	\definecolor{lightgrey204}{RGB}{204,204,204}
	\definecolor{sandybrown25220370}{RGB}{252,203,70}
	\definecolor{teal0101153}{RGB}{0,101,153}

	\begin{groupplot}[group style={group size=1 by 2}]
		\nextgroupplot[
			legend cell align={left},
			legend columns=2,
			legend style={
					fill opacity=0.8,
					draw opacity=1,
					text opacity=1,
					at={(0.4,-3.8)},
					anchor=south,
					draw=lightgrey204
				},
			tick align=outside,
			tick pos=left,
			x grid style={darkgrey176},
			xmin=4.5, xmax=12.5,
			xtick style={color=black},
			xtick={4.5,6.400000095,12.5},
			xticklabels={\empty,\empty,\empty},
			y grid style={darkgrey176},
			ylabel={\(\displaystyle e_1\) / \si{\degree}},
			ymin=-25, ymax=10,
			ytick style={color=black},
			ytick={-60,-40,-20,0,20},
			yticklabels={
					\(\displaystyle {\ensuremath{-}60}\),
					\(\displaystyle {\ensuremath{-}40}\),
					\(\displaystyle {\ensuremath{-}20}\),
					\(\displaystyle {0}\),
					\(\displaystyle {20}\)
				}
		]
		\path [draw=grey, fill=grey, opacity=0.5]
		(axis cs:0,90)
		--(axis cs:0,-90)
		--(axis cs:6.400000095,-90)
		--(axis cs:6.400000095,90)
		--(axis cs:6.400000095,90)
		--(axis cs:0,90)
		--cycle;

		\path [draw=grey, fill=grey, opacity=0.5]
		(axis cs:0,3)
		--(axis cs:0,-3)
		--(axis cs:12.5,-3)
		--(axis cs:12.5,3)
		--(axis cs:12.5,3)
		--(axis cs:0,3)
		--cycle;

		\addplot [semithick, firebrick1861843]
		table {%
				0.100000001 0
				0.600000008 -1.34793162074261e-06
				0.70000001 -3.47957711655768e-07
				0.800000011 -0.0238722887760150
				0.900000013 -0.365213543508051
				1.000000014 -1.65326302151029
				1.100000016 -4.57925811007626
				1.200000017 -9.55386729410262
				1.300000019 -16.3765284990955
				1.40000002 -24.0247562002509
				1.500000022 -34.7172004560657
				1.600000023 -42.2510432571841
				1.700000025 -47.7693268523526
				1.800000026 -51.3147760759652
				1.900000028 -53.1263808212254
				2.000000029 -53.4282057655858
				2.100000031 -52.4413785559082
				2.200000032 -50.4047721250734
				2.300000034 -47.5685227622748
				2.400000035 -44.174475211355
				2.500000037 -39.3253040515559
				2.600000038 -35.4699859947088
				2.70000004 -31.4609909019191
				2.800000041 -27.3554512527988
				2.900000043 -23.2701794302037
				3.000000044 -19.357266670148
				3.100000046 -15.7668675951124
				3.200000047 -12.6155509587681
				3.300000049 -9.95724365733582
				3.40000005 -7.74602369341351
				3.500000052 -5.89129582774124
				3.600000053 -4.33676259694913
				3.700000055 -3.05850272576577
				3.800000056 -2.04075208281541
				3.900000058 -1.26686461227112
				4.000000059 -0.712316707062122
				4.100000061 -0.342000949152777
				4.200000062 -0.114856498160631
				4.300000064 0.00998559968150841
				4.400000065 0.0678807131395399
				4.500000067 0.0858930741645321
				4.600000068 0.0825547006857522
				4.70000007 0.0692583258949787
				4.800000071 0.0522969781049769
				4.900000073 0.0348227804426217
				5.000000074 0.0183269385214758
				5.100000075 0.00356243561569839
				5.200000077 -0.00899843110774076
				5.300000078 -0.0190127196887832
				5.40000008 -0.0262684011578188
				5.500000081 -0.000711775007042639
				5.600000083 0.0150291606711045
				5.700000084 0.230656091259941
				5.800000086 0.785753239572132
				5.900000087 1.54957172811139
				6.000000089 2.31668597740543
				6.10000009 2.90910092704849
				6.200000092 3.21748288377885
				6.300000093 3.21278944278378
				6.400000095 2.92246093239601
				6.500000096 2.36923687520919
				6.600000098 1.55304922850789
				6.700000099 0.546391427121814
				6.800000101 -0.548674622189107
				6.900000102 -1.65766200345911
				7.000000104 -2.73940730775937
				7.100000105 -3.80938659658588
				7.200000107 -4.89749716393539
				7.300000108 -6.02986930460589
				7.40000011 -7.25349159804289
				7.500000111 -8.61800974286104
				7.600000113 -10.1349430098935
				7.700000114 -11.7667376921532
				7.800000116 -13.43494572732
				7.900000117 -15.0376886355991
				8.000000119 -16.4723422693652
				8.10000012 -17.6539332166545
				8.200000122 -18.5242874864881
				8.300000123 -19.0755107559171
				8.400000125 -19.2720072533849
				8.500000126 -19.1883273852328
				8.600000128 -18.8610277313095
				8.700000129 -18.3419990262483
				8.800000131 -17.6925017777883
				8.900000132 -16.9732658135737
				9.000000134 -16.2357939452566
				9.100000135 -15.5143846264058
				9.200000137 -14.8252124948723
				9.300000138 -14.1725474720847
				9.40000014 -13.5515425932332
				9.500000141 -12.9484890606631
				9.600000143 -12.299002941801
				9.700000144 -11.6474027905092
				9.800000146 -10.9321799236268
				9.900000147 -10.1454371649839
				10.000000149 -9.30177345129329
				10.10000015 -8.42065866327486
				10.200000151 -7.51582060030387
				10.300000153 -6.5953550359988
				10.400000154 -5.66824387407645
				10.500000156 -4.75176043744544
				10.600000157 -3.87227846167654
				10.700000159 -3.05657625507425
				10.80000016 -2.32391901288341
				10.900000162 -1.68733852219596
				11.000000163 -1.15868604915421
				11.100000165 -0.745835392043126
				11.200000166 -0.447271416401552
				11.300000168 -0.251097573144531
				11.400000169 -0.137705598360008
				11.500000171 -0.0844040878431391
				11.600000172 -0.0694959375480504
				11.700000174 -0.0750797976706604
				11.800000175 -0.0883140759120389
				11.900000177 -0.101371033203583
				12.000000178 -0.110563513997253
				12.10000018 -0.114990829262199
				12.200000181 -0.115327232990481
				12.300000183 -0.112949425817169
				12.400000184 -0.10924369670326
				12.500000186 -0.105251192416023
				12.600000187 -0.101572012760035
				12.700000189 -0.0984283638549262
				12.80000019 -0.0957902646875654
				12.900000192 -0.0933894578537838
				13.000000193 -0.0909264242880281
				13.100000195 -0.0882583386504513
				13.200000196 -0.0853715442678462
				13.300000198 -0.0824033971831088
				13.400000199 -0.0795310650781394
				13.500000201 -0.0768252984670733
				13.600000202 -0.0742862845793497
				13.700000204 -0.0719262357534025
				13.800000205 -0.0697519904322536
				13.900000207 -0.0677716171790392
				14.000000208 -0.0659488250232031
				14.10000021 -0.0642148589043083
				14.200000211 -0.0625096794511691
				14.300000213 -0.0607789416366331
				14.400000214 -0.0590020884914173
				14.500000216 -0.0571742959154925
				14.600000217 -0.0552885785642504
				14.700000219 -0.0533432890871132
				14.80000022 -0.0513464849790701
				14.900000222 -0.0493101180190669
				15.000000223 -0.0472464960335422
				15.100000225 -0.0451668718423555
				15.200000226 -0.0430812031949463
				15.300000227 -0.0409984986371857
				15.400000229 -0.0389272783027571
				15.50000023 -0.0368758778602205
				15.600000232 -0.034852512059055
				15.700000233 -0.0328651409884684
				15.800000235 -0.0309212435066633
			};
		\addlegendentry{Ours (No correction)}
		\addplot [semithick, teal0101153]
		table {%
				0.100000001 0
				0.600000008 -1.34897270983799e-06
				0.70000001 -3.48254670367955e-07
				0.800000011 -0.0238722894001689
				0.900000013 -0.365213544347868
				1.000000014 -1.65326302352925
				1.100000016 -4.57925811781092
				1.200000017 -9.55386731723748
				1.300000019 -16.3765285544511
				1.40000002 -24.0247563079311
				1.500000022 -34.7172006493192
				1.600000023 -42.2510435063867
				1.700000025 -47.7693271473629
				1.800000026 -51.3147764777616
				1.900000028 -53.1263814942197
				2.000000029 -53.4282069273898
				2.100000031 -52.4413803390011
				2.200000032 -50.4047744553335
				2.300000034 -47.5685254383898
				2.400000035 -44.174478010163
				2.500000037 -39.325307156811
				2.600000038 -35.4699915232364
				2.70000004 -31.4610008365647
				2.800000041 -27.3554647434209
				2.900000043 -23.2701948216089
				3.000000044 -19.357282845562
				3.100000046 -15.7668834038809
				3.200000047 -12.6155649661364
				3.300000049 -9.95725519464565
				3.40000005 -7.74603252813944
				3.500000052 -5.89130180728674
				3.600000053 -4.33676600512059
				3.700000055 -3.05850439697475
				3.800000056 -2.04075286341849
				3.900000058 -1.26686434768945
				4.000000059 -0.712315125105261
				4.100000061 -0.341998512455471
				4.200000062 -0.114853841329064
				4.300000064 0.00998796460827445
				4.400000065 0.0678825001027722
				4.500000067 0.085894214505674
				4.600000068 0.0825552840715784
				4.70000007 0.0692585175663268
				4.800000071 0.0522969431086361
				4.900000073 0.0348226492704685
				5.000000074 0.0183267919761589
				5.100000075 0.00356230948686248
				5.200000077 -0.0089985366353628
				5.300000078 -0.0190128201038275
				5.40000008 -0.0262685127211521
				5.500000081 -0.00071190990035084
				5.600000083 0.015028954432121
				5.700000084 0.230655601620737
				5.800000086 0.7857521112659
				5.900000087 1.54956968046436
				6.000000089 2.31668281702752
				6.10000009 2.90909687476321
				6.200000092 3.21747874624668
				6.300000093 3.21278607359436
				6.400000095 2.92245898196072
				6.500000096 2.36923662575662
				6.600000098 1.55305047247937
				6.700000099 0.546393888472681
				6.800000101 -0.548671196174728
				6.900000102 -1.63915141922689
				7.000000104 -2.37360438171966
				7.100000105 -2.31585219719019
				7.200000107 -1.54755794500753
				7.300000108 -0.40647607082559
				7.40000011 0.724128308641018
				7.500000111 1.53954759433451
				7.600000113 1.89307079219439
				7.700000114 1.80938753615737
				7.800000116 1.43191408470231
				7.900000117 0.952828602104716
				8.000000119 0.527477029496484
				8.10000012 0.240388505325941
				8.200000122 0.117231841241069
				8.300000123 0.138509586499845
				8.400000125 0.252547023327872
				8.500000126 0.392626786521314
				8.600000128 0.508056730514767
				8.700000129 0.560829317803628
				8.800000131 0.515551438918578
				8.900000132 0.351397429821729
				9.000000134 0.0713184787035217
				9.100000135 -0.296497917525909
				9.200000137 -0.708356347692945
				9.300000138 -1.11939524494269
				9.40000014 -1.49192934617556
				9.500000141 -1.7970031700861
				9.600000143 -2.01248698862813
				9.700000144 -2.0944434214171
				9.800000146 -2.07564183706618
				9.900000147 -1.93564143587935
				10.000000149 -1.70210315600031
				10.10000015 -1.42313908918221
				10.200000151 -1.14217199002619
				10.300000153 -0.886450074110148
				10.400000154 -0.668206305208697
				10.500000156 -0.490693556483494
				10.600000157 -0.353159536578504
				10.700000159 -0.2526941414767
				10.80000016 -0.184553049016461
				10.900000162 -0.142597911886884
				11.000000163 -0.119940005004995
				11.100000165 -0.110070067842625
				11.200000166 -0.107739766135952
				11.300000168 -0.10903777017559
				11.400000169 -0.11132332046197
				11.500000171 -0.113098315541749
				11.600000172 -0.113644310567013
				11.700000174 -0.11268440773973
				11.800000175 -0.110283740962894
				11.900000177 -0.10675075018024
				12.000000178 -0.102402882204477
				12.10000018 -0.0973701411594683
				12.200000181 -0.0917763186282339
				12.300000183 -0.0858703034776615
				12.400000184 -0.079933596705211
				12.500000186 -0.0742157336764759
				12.600000187 -0.0688936501757465
				12.700000189 -0.0640555814660847
				12.80000019 -0.0597117126905506
				12.900000192 -0.0558184446122396
				13.000000193 -0.052304789213716
				13.100000195 -0.0490941295923039
				13.200000196 -0.0461184777077179
				13.300000198 -0.0433252740084299
				13.400000199 -0.0406784404853677
				13.500000201 -0.03815595476086
				13.600000202 -0.0357510676394312
				13.700000204 -0.0335121897962909
				13.800000205 -0.0314666399003075
				13.900000207 -0.0295501009561604
				14.000000208 -0.027680944714318
				14.10000021 -0.0258038818675544
				14.200000211 -0.0238992550699374
				14.300000213 -0.0219517475190865
				14.400000214 -0.0199893291799264
				14.500000216 -0.0181082573010847
				14.600000217 -0.0164036050709057
				14.700000219 -0.0149322702631516
				14.80000022 -0.013704107199026
				14.900000222 -0.0126907397944053
				15.000000223 -0.0118425302468193
				15.100000225 -0.0111055385191014
				15.200000226 -0.0104337635990758
				15.300000227 -0.00979524111726199
				15.400000229 -0.00917280197013116
				15.50000023 -0.00856134302321688
				15.600000232 -0.00796359844168795
				15.700000233 -0.00738596494966488
				15.800000235 -0.00683528908661377
				15.900000236 -0.00631692531909443
				16.000000238 -0.00583402611269689
				16.100000239 -0.00538821986881118
				16.200000241 -0.00497895741471821
				16.300000242 -0.0046031578575048
				16.400000244 -0.00425701624791414
				16.500000245 -0.00393712935301522
				16.600000247 -0.00364092400574693
				16.700000248 -0.00336662411271093
				16.80000025 -0.00311298721373393
				16.900000251 -0.00287899974677356
				17.000000253 -0.00266364611924932
				17.100000254 -0.00246578927034673
				17.200000256 -0.00228414999871824
				17.300000257 -0.00211735443502903
				17.400000259 -0.00196404058948679
				17.50000026 -0.00182323993393872
				17.600000262 -0.00169459724658622
				17.700000263 -0.00157744941978072
				17.800000265 -0.0014710885063866
				17.900000266 -0.00138856978709129
				18.000000268 -0.00133364874970716
				18.100000269 -0.00128118601374848
				18.200000271 -0.00120345487761652
				18.300000272 -0.00107749232801979
				18.400000274 -0.0009024432150974
				18.500000275 -0.00070660556527338
				18.600000277 -0.000524357650170235
				18.700000278 -0.000381396165614144
				18.80000028 -0.000289186472102878
				18.900000281 -0.000245912647448044
			};
		\addlegendentry{Ours (Upright constraint)}
		\addplot [semithick, sandybrown25220370]
		table {%
				0.100000001 0
				0.600000008 -1.34824061453569e-06
				0.70000001 -3.48041213370406e-07
				0.800000011 -0.023872289020053
				0.900000013 -0.36521354408125
				1.000000014 -1.65326302179093
				1.100000016 -4.57925810593346
				1.200000017 -9.55386727390503
				1.300000019 -16.3765284419555
				1.40000002 -24.0247560865865
				1.500000022 -34.7172002559302
				1.600000023 -42.2510429968554
				1.700000025 -47.7693265249783
				1.800000026 -51.3147756192579
				1.900000028 -53.1263800802889
				2.000000029 -53.4282045295858
				2.100000031 -52.4413766907086
				2.200000032 -50.4047696898598
				2.300000034 -47.5685199393886
				2.400000035 -44.1744722060417
				2.500000037 -39.3253006460656
				2.600000038 -35.4699797717342
				2.70000004 -31.460979726299
				2.800000041 -27.3554364621338
				2.900000043 -23.270163229926
				3.000000044 -19.3572503863049
				3.100000046 -15.7668523747037
				3.200000047 -12.6155383167278
				3.300000049 -9.95723432286451
				3.40000005 -7.74601790145985
				3.500000052 -5.89129358919087
				3.600000053 -4.33676318206783
				3.700000055 -3.05850484200744
				3.800000056 -2.04075462203978
				3.900000058 -1.26686803144595
				4.000000059 -0.712321644764221
				4.100000061 -0.342006763016075
				4.200000062 -0.114862226965587
				4.300000064 0.00998069909688591
				4.400000065 0.0678770415495422
				4.500000067 0.0858907237922932
				4.600000068 0.0825535549704638
				4.70000007 0.069258150856402
				4.800000071 0.0522974863522569
				4.900000073 0.0348236865560251
				5.000000074 0.018327994580235
				5.100000075 0.00356345155501408
				5.200000077 -0.00899758112456996
				5.300000078 -0.0190120957809434
				5.40000008 -0.0262680111900042
				5.500000081 -0.000711786327306932
				5.600000083 0.00958122216966241
				5.700000084 0.157440116842536
				5.800000086 0.557875920270947
				5.900000087 1.13252885443593
				6.000000089 1.71756598513323
				6.10000009 2.16516984245212
				6.200000092 2.39156860322522
				6.300000093 2.39161581764854
				6.400000095 2.21291158784089
				6.500000096 1.89706589774681
				6.600000098 1.48637890975987
				6.700000099 1.09477993287155
				6.800000101 0.821537594124029
				6.900000102 0.729195562808374
				7.000000104 0.749397377347173
				7.100000105 0.983497687558706
				7.200000107 1.45823840000143
				7.300000108 2.06324142273117
				7.40000011 2.67876622617261
				7.500000111 3.18414987087309
				7.600000113 3.45198722704377
				7.700000114 3.39298837477174
				7.800000116 3.03428431525078
				7.900000117 2.4687201150143
				8.000000119 1.79861661544861
				8.10000012 1.12697660337053
				8.200000122 0.54091073617664
				8.300000123 0.0885393876474953
				8.400000125 -0.234877127195271
				8.500000126 -0.467842850409016
				8.600000128 -0.653108693773475
				8.700000129 -0.809507755589299
				8.800000131 -0.911793325270672
				8.900000132 -0.882453104772974
				9.000000134 -0.60451282409304
				9.100000135 0.0258068528773974
				9.200000137 0.961977690375994
				9.300000138 1.97871751208058
				9.40000014 2.82344208954542
				9.500000141 3.31191599039132
				9.600000143 3.34939008076998
				9.700000144 3.01607385819823
				9.800000146 2.51799512816205
				9.900000147 2.05675722022707
				10.000000149 1.76993110419309
				10.10000015 1.64257357179448
				10.200000151 1.61090795535896
				10.300000153 1.56037521235984
				10.400000154 1.40041779357405
				10.500000156 1.0940598000894
				10.600000157 0.65853307879088
				10.700000159 0.144764932614479
				10.80000016 -0.379626127269436
				10.900000162 -0.832452045404371
				11.000000163 -1.16917626375529
				11.100000165 -1.36142089802918
				11.200000166 -1.41667036247799
				11.300000168 -1.36222300675627
				11.400000169 -1.23170928724786
				11.500000171 -1.05645267898568
				11.600000172 -0.861560822173659
				11.700000174 -0.66613843695238
				11.800000175 -0.486038892553635
				11.900000177 -0.333367774372765
				12.000000178 -0.214302899429904
				12.10000018 -0.129395068297961
				12.200000181 -0.0747420270696254
				12.300000183 -0.0438750556789295
				12.400000184 -0.0296366668260848
				12.500000186 -0.0255617535901693
				12.600000187 -0.0266179876424292
				12.700000189 -0.0294046664114755
				12.80000019 -0.0319835076721637
				12.900000192 -0.0335080710380659
				13.000000193 -0.0338191805578759
				13.100000195 -0.0331037334056502
				13.200000196 -0.0316555609166491
				13.300000198 -0.0297318053555938
				13.400000199 -0.027520769313334
				13.500000201 -0.025156811615995
				13.600000202 -0.0227238177987976
				13.700000204 -0.0202736263471485
				13.800000205 -0.0178461979358921
				13.900000207 -0.0154869948042108
				14.000000208 -0.0132442324850562
				14.10000021 -0.0111433845659179
				14.200000211 -0.00920946629957835
				14.300000213 -0.00748938472713788
				14.400000214 -0.00601425609646339
				14.500000216 -0.00478899449790135
				14.600000217 -0.00380620440398765
				14.700000219 -0.00304886478604175
				14.80000022 -0.00249331208640557
				14.900000222 -0.00211295084218384
				15.000000223 -0.00187535741078775
				15.100000225 -0.00174307970458547
				15.200000226 -0.0016853471584019
				15.300000227 -0.0017033827034239
				15.400000229 -0.00174454673715738
				15.50000023 -0.00173250347270432
				15.600000232 -0.00167413844408673
				15.700000233 -0.00160550160283581
				15.800000235 -0.00155574461370836
				15.900000236 -0.00153460260791341
				16.000000238 -0.00153420192183742
				16.100000239 -0.0015375005687946
				16.200000241 -0.00152713345126612
				16.300000242 -0.00149129673139401
				16.400000244 -0.00142587490070102
				16.500000245 -0.00133359919452248
				16.600000247 -0.00122162131369502
				16.700000248 -0.00109878522141876
				16.80000025 -0.000973436835832697
				16.900000251 -0.000852116661118614
			};
		\addlegendentry{Ours (Upright + avoid)}

		\nextgroupplot[
			tick align=outside,
			tick pos=left,
			x grid style={darkgrey176},
			xlabel={\(\displaystyle t\) / \si{\second}},
			xmin=4.5, xmax=12.5,
			xtick style={color=black},
			xtick={4.5,6.400000095,12.5},
			xticklabels={\(\displaystyle 4.5\),\(\displaystyle 6.4\),\(\displaystyle 12.5\)},
			y grid style={darkgrey176},
			ylabel={\(\displaystyle e_2\) / \si{\degree}},
			ymin=-10, ymax=20,
			ytick style={color=black},
			ytick={-20,0,20,40,60},
			yticklabels={
					\(\displaystyle {\ensuremath{-}20}\),
					\(\displaystyle {0}\),
					\(\displaystyle {20}\),
					\(\displaystyle {40}\),
					\(\displaystyle {60}\)
				}
		]
		\path [draw=grey, fill=grey, opacity=0.5]
		(axis cs:0,90)
		--(axis cs:0,-90)
		--(axis cs:6.400000095,-90)
		--(axis cs:6.400000095,90)
		--(axis cs:6.400000095,90)
		--(axis cs:0,90)
		--cycle;

		\path [draw=grey, fill=grey, opacity=0.5]
		(axis cs:0,3)
		--(axis cs:0,-3)
		--(axis cs:12.5,-3)
		--(axis cs:12.5,3)
		--(axis cs:12.5,3)
		--(axis cs:0,3)
		--cycle;

		\addplot [semithick, firebrick1861843]
		table {%
				0.100000001 -0
				0.600000008 -8.4461456897099e-07
				0.70000001 -6.9931460649804e-07
				0.800000011 0.0345720280136915
				0.900000013 0.493699287797859
				1.000000014 1.81377847440933
				1.100000016 3.55187891760016
				1.200000017 4.57267694223289
				1.300000019 4.00032580439311
				1.40000002 1.95105531120966
				1.500000022 -7.81362529730431
				1.600000023 -14.2455381909561
				1.700000025 -19.6486409917472
				1.800000026 -23.710053162025
				1.900000028 -26.4117317295219
				2.000000029 -27.8254430944414
				2.100000031 -28.0855419206823
				2.200000032 -27.3933914580268
				2.300000034 -25.9956854927722
				2.400000035 -24.1400024995435
				2.500000037 -16.1919212897233
				2.600000038 -14.0891946928102
				2.70000004 -11.9956523010194
				2.800000041 -9.79705845449348
				2.900000043 -7.48124300732369
				3.000000044 -5.15905331986625
				3.100000046 -3.01863510720732
				3.200000047 -1.24833323074875
				3.300000049 0.0249266925597142
				3.40000005 0.806836070182474
				3.500000052 1.19164454760213
				3.600000053 1.27397588931111
				3.700000055 1.14905051394211
				3.800000056 0.900109083364545
				3.900000058 0.600079777183495
				4.000000059 0.3098731106413
				4.100000061 0.07004029419943
				4.200000062 -0.0999538104923162
				4.300000064 -0.198949081160171
				4.400000065 -0.237987837634844
				4.500000067 -0.233749076053393
				4.600000068 -0.20332699225105
				4.70000007 -0.160956038551952
				4.800000071 -0.11665611129214
				4.900000073 -0.0763289057652555
				5.000000074 -0.0426201311148548
				5.100000075 -0.0160456291396881
				5.200000077 0.00399190309334424
				5.300000078 0.0184788437211365
				5.40000008 0.0283811251011258
				5.500000081 -0.00520659494162844
				5.600000083 -0.00189415929578371
				5.700000084 0.0628438203067099
				5.800000086 0.272563208582131
				5.900000087 0.660205939392149
				6.000000089 1.20458488390347
				6.10000009 1.85040522390105
				6.200000092 2.52422175367735
				6.300000093 3.16265909651879
				6.400000095 3.747810201664
				6.500000096 4.42014738576323
				6.600000098 5.22322930179097
				6.700000099 6.03459495102927
				6.800000101 6.8092944392947
				6.900000102 7.48650523486655
				7.000000104 8.19305772457611
				7.100000105 8.90817988117468
				7.200000107 9.60530496177118
				7.300000108 10.3005808193325
				7.40000011 11.0220328504836
				7.500000111 11.7137452924811
				7.600000113 12.2673870482457
				7.700000114 12.5665885667628
				7.800000116 12.5187191758464
				7.900000117 12.0748573405958
				8.000000119 11.2389026844021
				8.10000012 10.0603775923955
				8.200000122 8.61664440538716
				8.300000123 6.58014797369348
				8.400000125 4.86594706460535
				8.500000126 3.16983165943968
				8.600000128 1.58409735079642
				8.700000129 0.196765847913992
				8.800000131 -0.924692178088041
				8.900000132 -1.74793916624596
				9.000000134 -2.28019425457805
				9.100000135 -2.55961312889153
				9.200000137 -2.64272969616317
				9.300000138 -2.59372564262218
				9.40000014 -2.47316890184104
				9.500000141 -2.32922721938895
				9.600000143 -1.04326418078642
				9.700000144 -0.825232082581112
				9.800000146 -0.677830860359397
				9.900000147 -0.58591572763134
				10.000000149 -0.5362670891604
				10.10000015 -0.511351575076085
				10.200000151 -0.492432084361361
				10.300000153 -0.464164506613693
				10.400000154 -0.417718692339829
				10.500000156 -0.352542983634055
				10.600000157 -0.275617120963919
				10.700000159 -0.196540546736782
				10.80000016 -0.122652389947301
				10.900000162 -0.0588384290911946
				11.000000163 -0.00930614818283866
				11.100000165 0.0236154421180588
				11.200000166 0.0409082006224472
				11.300000168 0.0462139518800699
				11.400000169 0.0441825173204078
				11.500000171 0.0388963521577171
				11.600000172 0.0330679282522451
				11.700000174 0.0279736047396734
				11.800000175 0.0238467828173557
				11.900000177 0.0204061455785935
				12.000000178 0.0172844985034084
				12.10000018 0.0142819369762514
				12.200000181 0.0114205066635096
				12.300000183 0.00886691100273995
				12.400000184 0.00682272368530563
				12.500000186 0.00543018213393783
				12.600000187 0.0047222689693329
				12.700000189 0.0046209045663429
				12.80000019 0.00496787617337399
				12.900000192 0.00554190550038727
				13.000000193 0.0061319253059374
				13.100000195 0.00659824362588051
				13.200000196 0.00687269228054356
				13.300000198 0.00696762859347748
				13.400000199 0.00693276214266441
				13.500000201 0.00679741548326339
				13.600000202 0.00658404756071447
				13.700000204 0.00633285168365898
				13.800000205 0.00608440882384437
				13.900000207 0.00587666604383737
				14.000000208 0.00571825013472904
				14.10000021 0.00559694200709346
				14.200000211 0.00549861432682537
				14.300000213 0.00540648678668413
				14.400000214 0.00532004466563341
				14.500000216 0.00524198209920734
				14.600000217 0.00516628907807096
				14.700000219 0.00508752084317328
				14.80000022 0.00500632483447439
				14.900000222 0.0049250204202091
				15.000000223 0.00484498175953829
				15.100000225 0.00476585574783536
				15.200000226 0.00468582301122211
				15.300000227 0.00460234531313577
				15.400000229 0.00451294954862227
				15.50000023 0.00441578732838735
				15.600000232 0.00430988811971848
				15.700000233 0.00419514651108367
				15.800000235 0.00407214530898018
			};
		\addplot [semithick, teal0101153]
		table {%
				0.100000001 -0
				0.600000008 -8.45104700622785e-07
				0.70000001 -7.0119990404444e-07
				0.800000011 0.0345720261714542
				0.900000013 0.493699288675808
				1.000000014 1.81377848090383
				1.100000016 3.55187893013498
				1.200000017 4.57267695442133
				1.300000019 4.00032580098588
				1.40000002 1.95105527653865
				1.500000022 -7.81362541302555
				1.600000023 -14.2455383527841
				1.700000025 -19.6486411838198
				1.800000026 -23.7100533455367
				1.900000028 -26.411731814131
				2.000000029 -27.825443001694
				2.100000031 -28.0855416643606
				2.200000032 -27.3933911578333
				2.300000034 -25.9956853473784
				2.400000035 -24.1400027157298
				2.500000037 -16.1919218293005
				2.600000038 -14.0891960026999
				2.70000004 -11.9956541253152
				2.800000041 -9.79706149472784
				2.900000043 -7.48124804074164
				3.000000044 -5.15906027084861
				3.100000046 -3.01864231166215
				3.200000047 -1.2483387706991
				3.300000049 0.0249240426673296
				3.40000005 0.806836641591259
				3.500000052 1.19164747448814
				3.600000053 1.27397983444283
				3.700000055 1.14905467614065
				3.800000056 0.900113002159796
				3.900000058 0.600081374092727
				4.000000059 0.309870637461659
				4.100000061 0.0700345472362865
				4.200000062 -0.099960971315991
				4.300000064 -0.198955840257629
				4.400000065 -0.237993012066674
				4.500000067 -0.233752243477837
				4.600000068 -0.203328364098347
				4.70000007 -0.160956163549016
				4.800000071 -0.116655566055022
				4.900000073 -0.0763281400183969
				5.000000074 -0.0426195223470196
				5.100000075 -0.0160454098901982
				5.200000077 0.00399175014731864
				5.300000078 0.0184784623945937
				5.40000008 0.0283806889876813
				5.500000081 -0.00520676424787206
				5.600000083 -0.00189446284744487
				5.700000084 0.0628434789172265
				5.800000086 0.272562861848282
				5.900000087 0.660205591303738
				6.000000089 1.20458446477421
				6.10000009 1.8504039825995
				6.200000092 2.52421879614023
				6.300000093 3.16265415514498
				6.400000095 3.7478033122358
				6.500000096 4.42013920938228
				6.600000098 5.22322080477771
				6.700000099 6.03458665002375
				6.800000101 6.80928657087975
				6.900000102 7.41437898288848
				7.000000104 7.25765535203151
				7.100000105 6.062526023018
				7.200000107 4.50648239411191
				7.300000108 3.18340829976454
				7.40000011 2.4646429001315
				7.500000111 2.38463249259362
				7.600000113 2.72243668040847
				7.700000114 3.1387915144431
				7.800000116 3.34226761002602
				7.900000117 3.24071395719924
				8.000000119 2.90305695318145
				8.10000012 2.45238471252782
				8.200000122 2.02338125433159
				8.300000123 1.71976494641429
				8.400000125 1.60033226318418
				8.500000126 1.6467697100918
				8.600000128 1.82870559639698
				8.700000129 2.09869096175591
				8.800000131 2.39723382228948
				8.900000132 2.65425382776096
				9.000000134 2.80351174345586
				9.100000135 2.80597227723755
				9.200000137 2.6519533757556
				9.300000138 2.35573745600234
				9.40000014 1.95005194500131
				9.500000141 1.47997814933998
				9.600000143 0.996400653281868
				9.700000144 0.875622565940982
				9.800000146 0.54210837526921
				9.900000147 0.289891377019651
				10.000000149 0.122944939946903
				10.10000015 0.0268296090315885
				10.200000151 -0.0184390875563715
				10.300000153 -0.0311375336487554
				10.400000154 -0.0254595944394333
				10.500000156 -0.0117151901085212
				10.600000157 0.00298270217568804
				10.700000159 0.01438851468017
				10.80000016 0.0207962231508127
				10.900000162 0.0224003429012941
				11.000000163 0.0205302351921508
				11.100000165 0.0168182855299689
				11.200000166 0.0126210293546031
				11.300000168 0.00885230024087816
				11.400000169 0.00596740986824994
				11.500000171 0.00403399781234879
				11.600000172 0.00290258270296477
				11.700000174 0.00235463165366893
				11.800000175 0.00215493247935427
				11.900000177 0.00208846509919247
				12.000000178 0.00202677649465895
				12.10000018 0.00196722505131996
				12.200000181 0.00193299861229148
				12.300000183 0.00190561579675288
				12.400000184 0.00185398477237754
				12.500000186 0.00175298687809278
				12.600000187 0.00159281323409725
				12.700000189 0.00138176379989926
				12.80000019 0.00114108796131519
				12.900000192 0.000896772878007544
				13.000000193 0.000672228051327874
				13.100000195 0.000483727655784969
				13.200000196 0.000338935024336651
				13.300000198 0.000237840620026407
				13.400000199 0.000175041170714478
				13.500000201 0.000142340230392428
				13.600000202 0.000127988152483167
				13.700000204 9.28201962716417e-05
				13.800000205 1.33423295243323e-05
				13.900000207 -8.00071247025041e-05
				14.000000208 -0.000146665743873394
				14.10000021 -0.000160238162328616
				14.200000211 -0.000112671048423327
				14.300000213 7.10639822117982e-06
				14.400000214 0.000186694521555985
				14.500000216 0.000371955025442539
				14.600000217 0.000509940114743312
				14.700000219 0.000570589496671453
				14.80000022 0.000551270753160319
				14.900000222 0.000470631651693365
				15.000000223 0.000357663233003667
				15.100000225 0.000241190436718007
				15.200000226 0.000142767764786763
				15.300000227 7.37775827306552e-05
				15.400000229 3.60339315866327e-05
				15.50000023 2.45061497702141e-05
				15.600000232 3.07246135494476e-05
				15.700000233 4.5787105815123e-05
				15.800000235 6.23872819752935e-05
				15.900000236 7.57451234913937e-05
				16.000000238 8.3570370106342e-05
				16.100000239 8.49862935759663e-05
				16.200000241 8.06334456933098e-05
				16.300000242 7.27608833719403e-05
				16.400000244 6.37994175267473e-05
				16.500000245 5.55391340123914e-05
				16.600000247 4.8872708069228e-05
				16.700000248 4.39091312257424e-05
				16.80000025 4.02675542421442e-05
				16.900000251 3.7391140240906e-05
				17.000000253 3.47817909774182e-05
				17.100000254 3.21230710905659e-05
				17.200000256 2.93042912913726e-05
				17.300000257 2.63766840779121e-05
				17.400000259 2.34544504605534e-05
				17.50000026 2.03733337514725e-05
				17.600000262 1.64489390148722e-05
				17.700000263 1.11844172442999e-05
				17.800000265 4.02985232092924e-06
				17.900000266 -1.77488919440147e-05
				18.000000268 -6.07948257421805e-05
				18.100000269 -0.000108257909450717
				18.200000271 -0.000139302013182203
				18.300000272 -0.000133773085506722
				18.400000274 -8.6633922549174e-05
				18.500000275 -1.56819176277375e-05
				18.600000277 5.52783756883602e-05
				18.700000278 0.000107531604046654
				18.80000028 0.000132090077098278
				18.900000281 0.000129501531849029
			};
		\addplot [semithick, sandybrown25220370]
		table {%
				0.100000001 -0
				0.600000008 -8.44680633075786e-07
				0.70000001 -7.00155616707153e-07
				0.800000011 0.034572027044234
				0.900000013 0.493699286578002
				1.000000014 1.81377847132517
				1.100000016 3.55187891094422
				1.200000017 4.57267693709635
				1.300000019 4.0003258179016
				1.40000002 1.95105536049724
				1.500000022 -7.81362515445697
				1.600000023 -14.2455379898916
				1.700000025 -19.6486407294899
				1.800000026 -23.7100528528337
				1.900000028 -26.4117314281628
				2.000000029 -27.825442849118
				2.100000031 -28.0855417077449
				2.200000032 -27.3933911722925
				2.300000034 -25.9956849695302
				2.400000035 -24.1400015894321
				2.500000037 -16.1919203307746
				2.600000038 -14.0891939371946
				2.70000004 -11.9956524433755
				2.800000041 -9.79705851547327
				2.900000043 -7.48124147735939
				3.000000044 -5.15904958975601
				3.100000046 -3.01863055880772
				3.200000047 -1.24832978852517
				3.300000049 0.0249276412345093
				3.40000005 0.806834115722191
				3.500000052 1.19164070268939
				3.600000053 1.27397159602991
				3.700000055 1.14904674753912
				3.800000056 0.9001064829323
				3.900000058 0.600081335162499
				4.000000059 0.30988125411609
				4.100000061 0.0700532566476816
				4.200000062 -0.0999393074359567
				4.300000064 -0.198935948538696
				4.400000065 -0.237977882270603
				4.500000067 -0.233742883670238
				4.600000068 -0.203324236579722
				4.70000007 -0.160955880606142
				4.800000071 -0.116657531028879
				4.900000073 -0.0763309677259959
				5.000000074 -0.0426222164801432
				5.100000075 -0.0160474261003411
				5.200000077 0.00399055117177327
				5.300000078 0.0184779704272824
				5.40000008 0.0283806751876102
				5.500000081 -0.00520691936737915
				5.600000083 -0.00142101003848407
				5.700000084 0.0684206620814562
				5.800000086 0.282685508797426
				5.900000087 0.65498004976815
				6.000000089 1.14714551732539
				6.10000009 1.680942643331
				6.200000092 2.15090915353015
				6.300000093 2.47083503081948
				6.400000095 2.62124983707629
				6.500000096 2.69986804967031
				6.600000098 2.71585817174463
				6.700000099 2.61261115882238
				6.800000101 2.45021813565912
				6.900000102 2.08876882773213
				7.000000104 2.03628756495276
				7.100000105 2.18632782961654
				7.200000107 2.58183053633405
				7.300000108 3.09359320201495
				7.40000011 3.53160722844455
				7.500000111 3.78619038906191
				7.600000113 3.89501604229198
				7.700000114 3.92335475183415
				7.800000116 3.84524405664154
				7.900000117 3.62553472255007
				8.000000119 3.2797257021084
				8.10000012 2.87297094384783
				8.200000122 2.46385473488569
				8.300000123 2.10276932966195
				8.400000125 1.84655989796854
				8.500000126 1.73304348443
				8.600000128 1.76784979809189
				8.700000129 1.92752460938236
				8.800000131 2.1726310743586
				8.900000132 2.45818894108916
				9.000000134 2.73860143873651
				9.100000135 2.96860011812237
				9.200000137 3.18769541606714
				9.300000138 3.3128545157192
				9.40000014 3.27743901753723
				9.500000141 3.04258489239114
				9.600000143 2.61885980213793
				9.700000144 2.06297669834074
				9.800000146 1.44473145464665
				9.900000147 0.827183260034318
				10.000000149 0.416041101121629
				10.10000015 -0.0383621678000754
				10.200000151 -0.344790018667117
				10.300000153 -0.521315306316049
				10.400000154 -0.612357781609893
				10.500000156 -0.661894171275089
				10.600000157 -0.6950648847657
				10.700000159 -0.717170247250153
				10.80000016 -0.721230994002021
				10.900000162 -0.602676203611501
				11.000000163 -0.467428089624749
				11.100000165 -0.316042063353049
				11.200000166 -0.176108184072181
				11.300000168 -0.0668960403985419
				11.400000169 0.00362771320870565
				11.500000171 0.0375607418076848
				11.600000172 0.0436552963148487
				11.700000174 0.0332032255945535
				11.800000175 0.0162841071617608
				11.900000177 -5.0221318645708e-06
				12.000000178 -0.0116607336592775
				12.10000018 -0.0174289235993003
				12.200000181 -0.0179625559515912
				12.300000183 -0.0149207039181024
				12.400000184 -0.010208587380931
				12.500000186 -0.00544490030139999
				12.600000187 -0.00168082770532393
				12.700000189 0.000648193006367187
				12.80000019 0.00160562436996553
				12.900000192 0.0015628228707509
				13.000000193 0.0010002638864302
				13.100000195 0.000350683922776866
				13.200000196 -9.14496420808851e-05
				13.300000198 -0.000191193243244035
				13.400000199 4.10659488533588e-05
				13.500000201 0.000492975935790824
				13.600000202 0.00101562448686193
				13.700000204 0.00147421422827651
				13.800000205 0.00177729637324685
				13.900000207 0.00188358287862997
				14.000000208 0.00179956912861761
				14.10000021 0.00157755794945233
				14.200000211 0.00128372562960311
				14.300000213 0.00096680231172931
				14.400000214 0.000666575057228211
				14.500000216 0.000414369529914951
				14.600000217 0.000224496629416381
				14.700000219 9.66879671393005e-05
				14.80000022 2.04712116330554e-05
				14.900000222 -2.04883963064074e-05
				15.000000223 -4.09068329522279e-05
				15.100000225 -4.98862179357415e-05
				15.200000226 -5.5190407201095e-05
				15.300000227 -7.77860658514298e-05
				15.400000229 -0.000102300794309172
				15.50000023 -9.3085623636015e-05
				15.600000232 -5.71074183632519e-05
				15.700000233 -1.67218400051816e-05
				15.800000235 8.99620393761711e-06
				15.900000236 1.15225520724878e-05
				16.000000238 -7.13858091930237e-06
				16.100000239 -3.83983764026952e-05
				16.200000241 -7.18466646628328e-05
				16.300000242 -9.89059697441187e-05
				16.400000244 -0.0001146659204045
				16.500000245 -0.000118051068083474
				16.600000247 -0.000110904379314458
				16.700000248 -9.66421069087612e-05
				16.80000025 -7.8991678088162e-05
				16.900000251 -6.10992979078821e-05
			};
	\end{groupplot}

\end{tikzpicture}

%% file: tikz/avoid_sets_fix.tex
\begin{tikzpicture}

	\definecolor{darkgrey176}{RGB}{176,176,176}
	\definecolor{sandybrown25220370}{RGB}{252,203,70}
	\definecolor{firebrick1861843}{RGB}{186,18,43}
	\definecolor{limegreen019166}{RGB}{0,191,66}
	\definecolor{teal0101153}{RGB}{0,101,153}

	\node[inner sep=0pt, opacity=0.5] (qf) at (2.95,0.5)
	{\includegraphics[width=.40\textwidth]{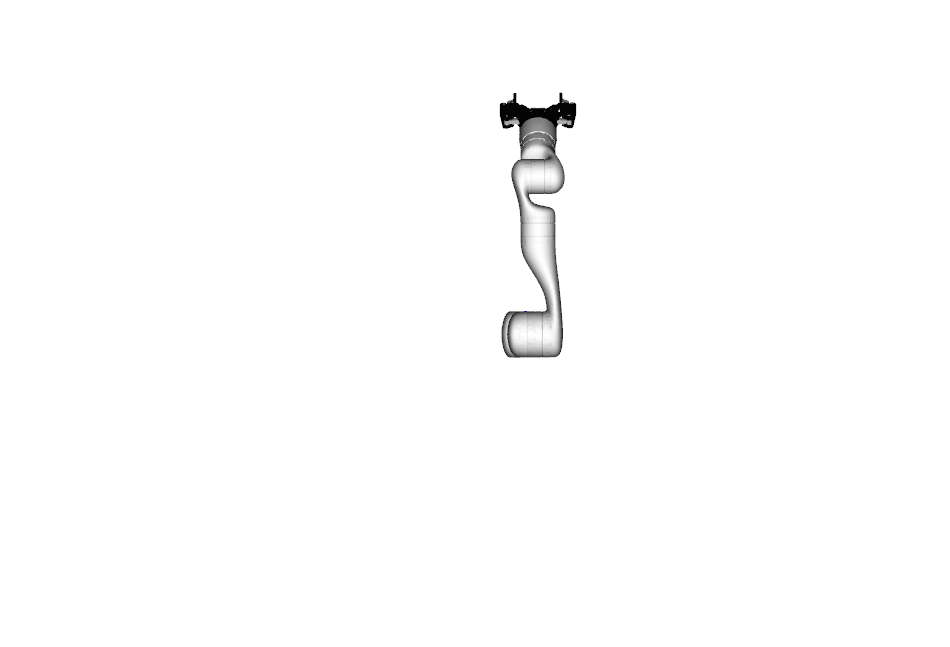}};
	\node[inner sep=0pt, opacity=0.5] (qf) at (2.95,0.5)
	{\includegraphics[width=.40\textwidth]{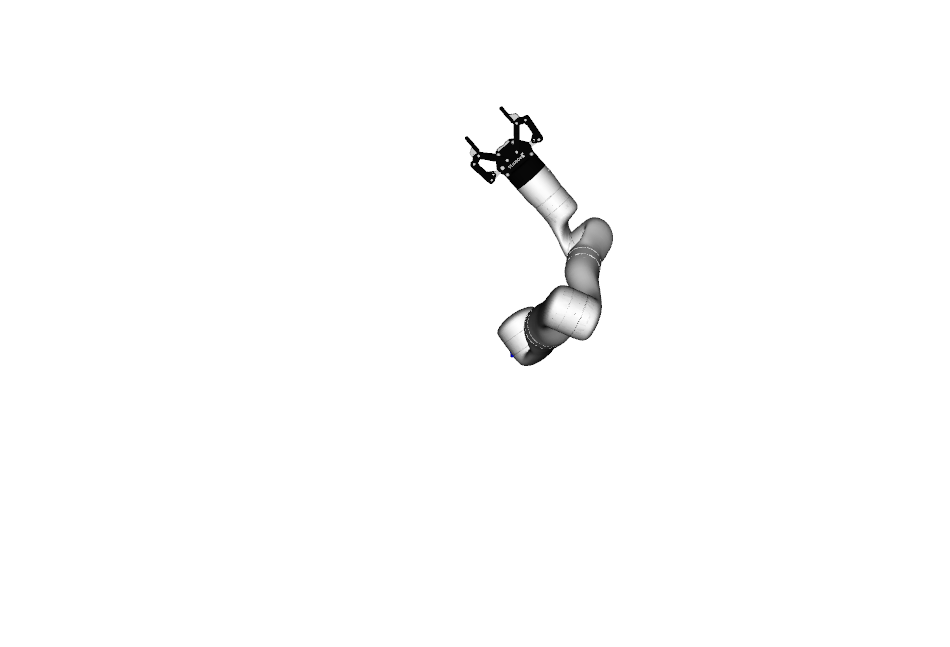}};
	\node[inner sep=0pt, opacity=0.5] (qf) at (2.95,0.5)
	{\includegraphics[width=.40\textwidth]{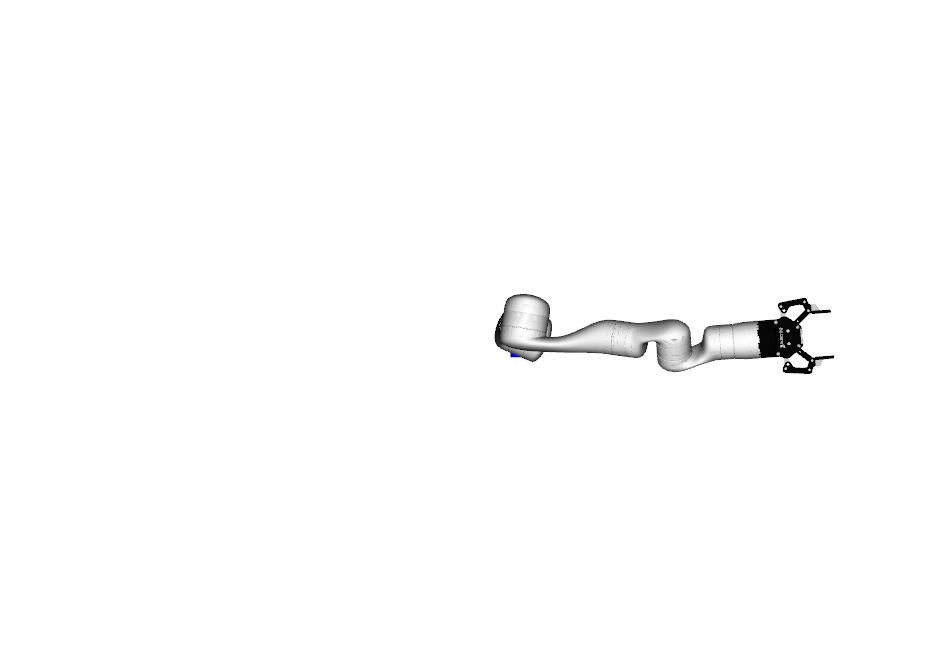}};
	\begin{axis}[
			tick align=outside,
			tick pos=left,
			x dir=reverse,
			x grid style={darkgrey176},
			xmin=-0.7, xmax=0.6,
			ylabel={$x$ / \si{\meter}},
			xlabel={$y$ / \si{\meter}},
			xtick style={color=black},
			axis equal,
			xtick={-0.8,-0.6,-0.4,-0.2,0,0.2,0.4,0.6},
			xticklabels={
					\(\displaystyle {\ensuremath{-}0.8}\),
					\(\displaystyle {\ensuremath{-}0.6}\),
					\(\displaystyle {\ensuremath{-}0.4}\),
					\(\displaystyle {\ensuremath{-}0.2}\),
					\(\displaystyle {0.0}\),
					\(\displaystyle {0.2}\),
					\(\displaystyle {0.4}\),
					\(\displaystyle {0.6}\)
				},
			y grid style={darkgrey176},
			ymin=-0.1, ymax=0.6,
			ytick style={color=black},
			ytick={0.0,0.2,0.4,0.6},
			yticklabels={
					\(\displaystyle {0.0}\),
					\(\displaystyle {0.2}\),
					\(\displaystyle {0.4}\),
					\(\displaystyle {0.6}\),
				}
		]
		\path [draw=limegreen019166, fill=limegreen019166, opacity=0.5]
		(axis cs:-0.35,1)
		--(axis cs:-0.35,0.15)
		--(axis cs:0.6,0.15)
		--(axis cs:0.6,1)
		--(axis cs:0.6,1)
		--(axis cs:-0.35,1)
		--cycle;

		\path [draw=black, fill=black, opacity=0.5]
		(axis cs:-0.35,0.55)
		--(axis cs:-0.35,0.15)
		--(axis cs:-0.05,0.15)
		--(axis cs:-0.05,0.55)
		--(axis cs:-0.05,0.55)
		--(axis cs:-0.35,0.55)
		--cycle;

		\addplot [semithick, firebrick1861843]
		table {%
				-0.0248508639923239 0.467325000633354
				-0.0248509288907806 0.467324999675867
				-0.0241243167648951 0.467346389545293
				-0.0146591272827481 0.467520110994555
				0.0114081504427863 0.467009609506318
				0.0511002856454595 0.463739998812222
				0.0968089545940944 0.456571775583715
				0.141609383916668 0.446414679222866
				0.181312986237193 0.436071532519546
				0.214580074627691 0.428449270857232
				0.241850305679005 0.424757302501843
				0.26410333261957 0.424534543721221
				0.282125795141336 0.426271155628308
				0.296281608971712 0.428286952469713
				0.306709364485161 0.429312347904342
				0.313644796941416 0.428636602514127
				0.317600781674785 0.426087159683893
				0.319348112314963 0.421939516172306
				0.319771233577314 0.416776945560317
				0.319696086283113 0.411365293757279
				0.31977868428876 0.406544592038721
				0.32051998583859 0.402909525195911
				0.322240342717361 0.400646671110175
				0.325136345747929 0.39964417335249
				0.329333041950214 0.399626009270782
				0.334785938226507 0.400325402731335
				0.341260390127911 0.401519919900673
				0.348430427415329 0.402981157405483
				0.355869145284673 0.40446203069972
				0.363191829040546 0.405648850116077
				0.370116108911936 0.40628533698134
				0.376344579312956 0.406344302741684
				0.381646459064114 0.405983097738347
				0.385944581662528 0.405330107163052
				0.389316267627815 0.404408204284048
				0.391914973005728 0.403231314765475
				0.393909433713647 0.401853672986824
				0.395442616066571 0.400390867073971
				0.39662017401778 0.398995277505097
				0.39751605164363 0.397815183192338
				0.398183742595663 0.396960152282106
				0.398666420092419 0.396482559684092
				0.399002859703833 0.396376573741688
				0.399229155319309 0.396590071735445
				0.399377977224092 0.397042394168523
				0.399476893819855 0.397642511018939
				0.399547057602827 0.398303842870385
				0.399602892602358 0.398954075032854
				0.399652772233171 0.399539966648816
				0.399700389762016 0.400028070288815
				0.399533800638521 0.400451059809721
				0.396708496406518 0.401371320949911
				0.387743837988793 0.403637627138558
				0.371273060324659 0.407328300519556
				0.347800102859288 0.41170644683536
				0.318795218183989 0.415928481264015
				0.285942335391315 0.419418944355631
				0.250738305532103 0.421855494813252
				0.214411039987939 0.423061521589251
				0.178996382079406 0.422480213143943
				0.145280624143891 0.420377207751304
				0.111896159806147 0.417656510786841
				0.0779897081212808 0.414404139950025
				0.0434093640528539 0.410188712835001
				0.00846453278391317 0.404492011031947
				-0.0263767591819681 0.396955284685468
				-0.0607402767379965 0.387364367756358
				-0.0945978848924591 0.374997472292603
				-0.127548910310566 0.35892712723856
				-0.158677818724361 0.339086976964308
				-0.187363442841244 0.316078682553853
				-0.213523036995636 0.290747025789143
				-0.237470540802299 0.263889404860253
				-0.259683527821227 0.236134631965829
				-0.280618064620697 0.207925185534197
				-0.300597753236526 0.179620586779152
				-0.319814016762152 0.151649576565397
				-0.338399355174449 0.124538653801069
				-0.356472203538935 0.0988475484802932
				-0.374128754952314 0.0751296266760756
				-0.391434616527808 0.0540206347864366
				-0.408410599944618 0.0361080917890277
				-0.424977334319975 0.0216709593712401
				-0.440947214387041 0.0106875625932154
				-0.456059703610315 0.00292819657307323
				-0.470048426212116 -0.00196473833981451
				-0.482698199503668 -0.00442389299059433
				-0.493872055719775 -0.00493469424053507
				-0.503527140410298 -0.0040542725321707
				-0.511722024247724 -0.0024234945652204
				-0.518616088176499 -0.000720333052966818
				-0.524457268452704 0.000510072633545431
				-0.529540945781726 0.00112837539952825
				-0.534116360302619 0.00138468926945591
				-0.538342841325757 0.00156452008545143
				-0.542327351642818 0.00180329147133807
				-0.546153746845987 0.00211529373245899
				-0.549890244264027 0.00244299421610364
				-0.55358259203417 0.002690545351632
				-0.557243147091647 0.00276757864330888
				-0.560851943913056 0.0026397986023039
				-0.564376076431892 0.00232668782548547
				-0.567785285685888 0.0018677786258483
				-0.571048184914258 0.00132734191140956
				-0.574123572686314 0.000811155427725135
				-0.576968562075394 0.000426479143240384
				-0.579551035777818 0.000230504540619758
				-0.581854522058818 0.000218299798422464
				-0.583877750653433 0.000339402700517414
				-0.585631092721079 0.000525630568049479
				-0.587132954471182 0.00071511429556739
				-0.588406412554013 0.000866254299844826
				-0.589476491940434 0.000961391767881568
				-0.590368248950458 0.0010031261971153
				-0.591105455106406 0.00100724311638973
				-0.591710470427628 0.000994533885990452
				-0.592204064522057 0.000984347695522574
				-0.59260480386374 0.000991169822538087
				-0.592929074217989 0.00102322787672666
				-0.593191149124978 0.00108269100250475
				-0.593403271658686 0.00116688433065474
				-0.593575757540422 0.00126990978126587
				-0.593717150703356 0.00138433801540338
				-0.593834870047592 0.00150425278513679
				-0.59393523463448 0.00162511925585365
				-0.594023244657797 0.00174277627828181
				-0.594102812592993 0.00185402130514555
				-0.594176690227939 0.00195561308893567
				-0.59424680060794 0.00204524673683086
				-0.594314856851192 0.00212348275381425
				-0.594382379722424 0.00219230963631007
				-0.59445044861357 0.00225306514318094
				-0.59451978578116 0.00230657206078619
				-0.594590759060835 0.00235289073864372
				-0.594663658317671 0.00239263255207252
				-0.594738769844672 0.00242685486762062
				-0.594816300182543 0.00245642779264116
				-0.594896448646925 0.00248229146361675
				-0.594979265460937 0.00250447672286201
				-0.595064728013217 0.00252271565738154
				-0.595152868297927 0.00253721280259574
				-0.595243733669023 0.00254822698266115
				-0.595337326736487 0.00255568347116835
				-0.595433617840003 0.0025593432936409
				-0.595532557016505 0.0025589357136837
				-0.595634078697068 0.0025542225967286
				-0.595738102091021 0.00254502152952493
				-0.595844529313684 0.0025312064176264
				-0.595953242825873 0.00251269991902651
				-0.596064103192233 0.00248946616189996
				-0.596176947654538 0.00246150702839467
				-0.596291589643417 0.00242886191059578
				-0.59640781909819 0.00239160914878518
				-0.596525403597207 0.00234986801339017
			};
		\addplot [semithick, teal0101153]
		table {%
				-0.0248508639923239 0.467325000633355
				-0.0248509289685088 0.46732499966877
				-0.0241243168884044 0.467346389535622
				-0.0146591274103699 0.467520110983365
				0.0114081503218481 0.467009609478635
				0.0511002855076945 0.46373999873525
				0.0968089544026565 0.456571775427431
				0.141609383633813 0.446414678972591
				0.181312985837729 0.436071532173852
				0.214580074115734 0.428449270372709
				0.241850305069765 0.424757301791435
				0.264103331906276 0.424534542694607
				0.282125794070758 0.426271153831109
				0.296281606846753 0.428286948932081
				0.306709360379866 0.429312341479619
				0.313644790494669 0.428636592796581
				0.317600773653915 0.426087147570998
				0.319348104279748 0.42193950335357
				0.319771226953996 0.416776934259471
				0.319696080243432 0.411365286221043
				0.319778669449537 0.406544582416448
				0.320519962130525 0.402909502332574
				0.322240322400317 0.400646634328786
				0.325136333054202 0.399644131150776
				0.329333029347845 0.399625973205891
				0.334785917673901 0.400325378200656
				0.341260361909876 0.401519904086171
				0.348430394536506 0.402981144957287
				0.355869111714088 0.404462016641616
				0.363191798816934 0.405648832828796
				0.370116082794497 0.406285319718307
				0.37634455809614 0.406344284134643
				0.38164644262975 0.405983076174558
				0.3859445621593 0.405330106873602
				0.389316239716539 0.404408248454316
				0.391914941617359 0.403231395900958
				0.393909405828975 0.401853770794452
				0.395442596217657 0.400390961780249
				0.396620163203576 0.398995356406388
				0.397516048204223 0.397815241501609
				0.398183743574823 0.396960191419342
				0.39866642261869 0.396482584248897
				0.399002861919002 0.396376588677635
				0.399229156473714 0.39659008102561
				0.399377976686541 0.397042400850623
				0.39947689120503 0.397642516851001
				0.399547053417764 0.398303848107933
				0.399602887646633 0.398954079314266
				0.399652767266825 0.399539969572635
				0.399700385324301 0.400028071689412
				0.399533796886741 0.400451059909531
				0.396708492137465 0.401371321079808
				0.387743829208286 0.403637629868677
				0.371273043205653 0.407328307460928
				0.347800076500839 0.411706456936255
				0.318795183007896 0.415928492544246
				0.285942292145993 0.419418955773093
				0.250738254794683 0.421855505883585
				0.214410980833103 0.423061532843049
				0.178996315819182 0.422480224754114
				0.145280554828483 0.420377219639203
				0.111896089497907 0.417656523840333
				0.0779896371534107 0.414404155265061
				0.0433647922738226 0.410179507375659
				0.00761325259550984 0.404485907027047
				-0.0300367735558701 0.39733543074843
				-0.0693399846435884 0.38865086613032
				-0.109203236679013 0.377538786299606
				-0.147802347673828 0.362987520767323
				-0.183264248166398 0.344857891986179
				-0.214527288335575 0.323626352366942
				-0.241418649083304 0.299992021378827
				-0.264519358349933 0.274601687072288
				-0.284855086010954 0.247867056393908
				-0.303374886290407 0.220120488944202
				-0.320743032776254 0.191803171342897
				-0.337391018420924 0.163469427763822
				-0.353582744503939 0.135763670059037
				-0.369485324588578 0.109338233995063
				-0.385208677366591 0.0847678854095881
				-0.400880078116379 0.0625481759618801
				-0.416592709444305 0.043089900858372
				-0.432274281994423 0.0266573540864962
				-0.447731966503783 0.0133793941350445
				-0.462746337866763 0.00326163641760448
				-0.477127083754254 -0.00380141616107087
				-0.490730070825759 -0.00797905337472181
				-0.503451211639028 -0.00951371020949759
				-0.515212372738837 -0.00877798556008813
				-0.52595586839269 -0.00633651592186413
				-0.535653476974099 -0.00299067774125907
				-0.544322214752417 0.000269952945521297
				-0.552028740904065 0.00251662125938633
				-0.55883852804768 0.00331929251665384
				-0.564758011949128 0.00286150314362733
				-0.569774954606241 0.00161952901847764
				-0.573928379930585 6.86743376567657e-05
				-0.577319431335982 -0.00144685983390498
				-0.580083763530649 -0.0027330707197247
				-0.582356892262552 -0.003709929155209
				-0.584253140415384 -0.00436727985622937
				-0.585860762887496 -0.00473553562811938
				-0.587245299510988 -0.004867730033861
				-0.588454483838456 -0.0048257293893054
				-0.58952330578265 -0.0046707684221175
				-0.590476405056296 -0.00445409030641159
				-0.59132973000473 -0.00421176790629764
				-0.592093854915863 -0.00396699544052597
				-0.5927763949336 -0.00373307921886617
				-0.593383288735702 -0.00351582003528028
				-0.593920243940829 -0.00331698542101494
				-0.594393681526736 -0.00313689815384693
				-0.594810344752478 -0.00297453645096456
				-0.59517675934267 -0.00282754639771923
				-0.595499576031913 -0.00269377290684424
				-0.595786237662957 -0.0025729103907789
				-0.59604385202436 -0.00246461174096753
				-0.596278109018428 -0.0023669901168181
				-0.596493438334822 -0.0022774831184111
				-0.596693213261957 -0.00219350516879238
				-0.596879968502943 -0.00211284593066358
				-0.597055615420616 -0.00203388071583772
				-0.597221595133299 -0.00195556755975315
				-0.597378985402143 -0.00187733618751578
				-0.597528582607711 -0.00179895502454962
				-0.597670969798543 -0.00172041828037684
				-0.597806575277311 -0.00164186595501412
				-0.597935722325175 -0.00156353291014361
				-0.598058669160662 -0.00148571678836421
				-0.598175638387463 -0.00140875493507781
				-0.598286777724608 -0.00133293281690152
				-0.598391630933712 -0.00125783065377034
				-0.598489648325001 -0.00118308937197023
				-0.598581026490479 -0.00110937662441647
				-0.598666366001711 -0.00103770041427194
				-0.598746383379709 -0.000968952690521781
				-0.598821816079229 -0.000903800351462586
				-0.598893756692049 -0.000843080362155986
				-0.598963186275424 -0.000787291808041478
				-0.599030343468355 -0.000736040983607501
				-0.599095061361927 -0.000688608552762471
				-0.599157036897111 -0.000644311567118663
				-0.599215983992849 -0.000602644745311876
				-0.599271702302387 -0.000563289908340853
				-0.599324090735744 -0.000526064654873609
				-0.599373134601058 -0.000490862015736538
				-0.59941888545509 -0.000457606243634681
				-0.599461442857215 -0.000426229825701111
				-0.599500940560859 -0.000396666429008851
				-0.59953753652287 -0.000368851885075222
				-0.599571405075027 -0.00034272674461137
				-0.599602729950714 -0.00031823694447471
				-0.599631697654398 -0.000295331867331977
				-0.599658491325211 -0.000273960732551097
				-0.599683284286666 -0.000254067707996313
				-0.599706224521707 -0.000235579351248544
				-0.599727443357176 -0.000218414668693859
				-0.599747073143709 -0.000202501168452433
				-0.599765240934407 -0.000187767836462137
				-0.599782063761434 -0.00017414101351279
				-0.599797647158831 -0.000161544364312864
				-0.599812085737632 -0.00014990094550974
				-0.59982546472775 -0.000139135752835567
				-0.599837861703387 -0.000129177799207039
				-0.599849348023023 -0.000119961338340597
				-0.599859989859228 -0.000111426300741711
				-0.599869848894464 -0.000103518202983683
				-0.599878982750242 -9.61877386130502e-05
				-0.599887444723525 -8.93898098466129e-05
				-0.599895278072216 -8.30782720462382e-05
				-0.599902508670331 -7.72002309439495e-05
				-0.599909153710035 -7.17042730762953e-05
				-0.599915216777261 -6.65362100366412e-05
				-0.599920429160498 -6.14303913165306e-05
				-0.59992448761785 -5.61371747673e-05
				-0.599927456799712 -5.07464358811113e-05
				-0.599929612158138 -4.55007851403897e-05
				-0.599931439747826 -4.07646707747496e-05
				-0.599933404357936 -3.68207670082485e-05
				-0.599935684003344 -3.36766736778239e-05
				-0.599938275817873 -3.12030302634883e-05
				-0.599941096246932 -2.92410752374716e-05
				-0.599944041400967 -2.76520353277485e-05
				-0.599947016652136 -2.63292730401305e-05
				-0.599949947291125 -2.51938522405661e-05
			};
		\addplot [semithick, sandybrown25220370]
		table {%
				-0.0248508639923239 0.467325000633355
				-0.0248509289070195 0.467324999667654
				-0.0241243167829239 0.46734638953122
				-0.0146591272954259 0.467520110976392
				0.0114081504481431 0.467009609497562
				0.0511002856863132 0.463739998839003
				0.0968089546936518 0.456571775671264
				0.141609384104662 0.446414679382162
				0.181312986540067 0.436071532745999
				0.214580075062868 0.428449271175861
				0.241850306253802 0.424757302957646
				0.264103333353425 0.424534544358167
				0.282125796283721 0.4262711569183
				0.296281611151301 0.428286955425399
				0.306709368514821 0.429312353645991
				0.313644803119783 0.428636611417593
				0.317600789283698 0.426087170883059
				0.319348119934053 0.421939528050095
				0.319771239916678 0.416776956076125
				0.31969609231874 0.411365301515035
				0.31977870144225 0.406544604916299
				0.320520013975197 0.40290955615139
				0.322240365345442 0.400646719667278
				0.325136356174843 0.399644227581243
				0.329333049698615 0.39962605317222
				0.334785954250195 0.40032542851122
				0.341260414275241 0.401519931783999
				0.348430455762122 0.402981163799217
				0.355869172892041 0.404462039537799
				0.36319185143722 0.405648864723193
				0.370116126288662 0.406285354581223
				0.37634459299395 0.406344322113083
				0.381646471086604 0.40598311627942
				0.385944603949163 0.405330085616879
				0.389316307099387 0.404408107288394
				0.391915019527284 0.403231158524998
				0.39390947481099 0.401853491467242
				0.395442644361558 0.400390690393357
				0.39662018830882 0.398995124490667
				0.397516055076266 0.397815060701838
				0.398183739887419 0.396960058762768
				0.398666415504766 0.396482489277981
				0.399002856171267 0.396376519353065
				0.399229154294115 0.396590027177019
				0.399377978458621 0.397042355466963
				0.399476896159137 0.397642476189659
				0.399547060177819 0.398303811042432
				0.399602894902043 0.398954046184265
				0.399652774019988 0.399539941223606
				0.39970039098492 0.400028048809462
				0.399535750892789 0.400437044704606
				0.396731283739294 0.40118600975893
				0.387777041251983 0.403084431938246
				0.371274025767912 0.406357984939143
				0.347751480155509 0.410360971682618
				0.318708799422423 0.414289436212295
				0.285829822620113 0.417572584521915
				0.25057739563871 0.419867548030005
				0.214227220296469 0.420896104792388
				0.179010997493431 0.419903816516631
				0.145601181810899 0.417223435184413
				0.112233198836895 0.414024821085664
				0.0778719348463748 0.410535056776612
				0.0423803762057789 0.406352744366651
				0.00673385995090689 0.4008189464924
				-0.0223601519807616 0.393151716745436
				-0.0380547412834413 0.384586654988988
				-0.0418470893605073 0.376710636692961
				-0.0392360164612617 0.369419569805097
				-0.0354890296599352 0.361636172935008
				-0.0336804622219442 0.352207101968756
				-0.034457467254916 0.340313716757526
				-0.0368326331032656 0.325679294954995
				-0.0393805939051173 0.308549326620376
				-0.0411258617210024 0.289473445353088
				-0.0418103099014964 0.269070990312491
				-0.0414353102043036 0.248000873501495
				-0.0399703125704125 0.226940656918189
				-0.0374674612501489 0.20637113306568
				-0.0341342000886192 0.186449888516822
				-0.030341042310968 0.167042815062601
				-0.0267297875101526 0.147844926205724
				-0.0243936422830315 0.128544981130627
				-0.0249842755263782 0.108999071927473
				-0.0306405629102846 0.0893450859642878
				-0.0434671363509261 0.0700038964219275
				-0.0639349166878195 0.0521094792338181
				-0.0903369454514193 0.0367350217700825
				-0.12029488715874 0.0238845764526063
				-0.151859834062604 0.0129371313350338
				-0.18374529714204 0.00311522776569154
				-0.215393572268615 -0.00610864881768446
				-0.24656549751962 -0.0148681228479401
				-0.276955553990188 -0.0230131241753917
				-0.306369122201306 -0.0302577199802664
				-0.334724969123771 -0.0362740409770887
				-0.361934679073017 -0.0407715237820755
				-0.387859444544832 -0.0434922102339981
				-0.412294050389093 -0.044133578319693
				-0.43506147130327 -0.0424671611672791
				-0.45605413800849 -0.0385293484071993
				-0.475205952900959 -0.0327268603293753
				-0.49248095896416 -0.0258520386474447
				-0.507894425866765 -0.0189832126491431
				-0.521520340066599 -0.0131973594564019
				-0.533453333191668 -0.00913329327282509
				-0.543777978578384 -0.00682221234627405
				-0.552585785556191 -0.00590125657587083
				-0.560000895238436 -0.005873638006999
				-0.566185505269273 -0.00628349714364924
				-0.571325901198997 -0.00680077296597495
				-0.575607267499787 -0.00723361664587621
				-0.579186596076568 -0.00749003319479613
				-0.582190217378933 -0.00755271149951212
				-0.584723731291835 -0.00745698771385335
				-0.58687327608404 -0.00724921135145495
				-0.588708156195061 -0.00696648483102053
				-0.590283037264918 -0.00663118814419384
				-0.5916401049744 -0.00625366036253287
				-0.592811610211017 -0.00583836441198213
				-0.59382265438261 -0.0053898033855861
				-0.594693582773436 -0.00491574162577694
				-0.59544168715246 -0.00442751480808386
				-0.596082380517752 -0.00393875144789675
				-0.596629793251622 -0.00346333630047838
				-0.597096940760451 -0.00301349031024145
				-0.597495686265819 -0.00259851195443966
				-0.597836778926263 -0.00222440164100612
				-0.598129565284268 -0.00189383217479265
				-0.598381733772177 -0.00160653864113347
				-0.598599556341572 -0.00136023605131583
				-0.598788135363723 -0.00115144087231535
				-0.598951609541323 -0.000976092594465154
				-0.599093255758835 -0.000829939654581322
				-0.599215724424214 -0.000708851190687085
				-0.599321606759362 -0.000609175353055781
				-0.599413257981411 -0.000527632817363987
				-0.59949224464354 -0.000461033862439773
				-0.599559887066061 -0.000406555817360102
				-0.599617632216598 -0.000361897191242478
				-0.599666831432342 -0.000325127050697392
				-0.599708703969405 -0.000294635098997757
				-0.599744310190278 -0.000269089138775225
				-0.599774508106204 -0.000247384194627556
				-0.599800043672959 -0.000228646092342614
				-0.599821633900174 -0.000212225491456896
				-0.599839845301583 -0.000197605722977147
				-0.599854614801466 -0.00018419245268013
				-0.599866776988915 -0.000171918262676578
				-0.599877989165114 -0.00016109376266142
				-0.599888870914182 -0.000151587757886129
				-0.5998994139855 -0.000143057904214983
				-0.599909404071807 -0.000135161280969064
				-0.59991863966385 -0.000127642503640714
				-0.599927011489034 -0.000120344809609439
				-0.599934503842674 -0.00011318682848935
				-0.599941165224144 -0.000106133247856461
				-0.599947076224508 -9.91727141320082e-05
				-0.599952326357908 -9.23057841871825e-05
				-0.599957001474016 -8.55404529758323e-05
				-0.59996117890066 -7.88914490204865e-05
				-0.599964926628687 -7.23802861707845e-05
				-0.599968303825049 -6.60345272288161e-05
				-0.599971361364636 -5.9885971676928e-05
				-0.599974142104986 -5.39681729941825e-05
			};
		\addplot [semithick, firebrick1861843, mark=*, mark size=1, mark options={solid}, only marks]
		table {%
				-0.59999070950224 -2.18003329833516e-06
			};
		\addplot [semithick, firebrick1861843, mark=*, mark size=1, mark options={solid}, only marks]
		table {%
				-0.0248508640072399 0.467325000633126
			};
		\addplot [semithick, firebrick1861843, mark=*, mark size=1, mark options={solid}, only marks]
		table {%
				0.396904642043533 0.395041479116036
			};
		\addplot [semithick, firebrick1861843, mark=*, mark size=1, mark options={solid}, only marks]
		table {%
				0.07 0.41
			};
		\draw (axis cs:0.05,0.43) node[
			scale=0.75,
			anchor=base west,
			text=black,
			rotate=0.0
		]{\bfseries $\vec{p}_{0}$};
		\draw (axis cs:-0.6,-0.01) node[
			scale=0.75,
			anchor=base west,
			text=black,
			rotate=0.0
		]{\bfseries $\vec{p}_\mathrm{f}$};
		\draw (axis cs:0.18,0.35) node[
			scale=0.75,
			anchor=base west,
			text=black,
			rotate=0.0
		]{\bfseries $\vec{p}_\mathrm{1, avoid}$};
		\draw (axis cs:-0.05,0.5) node[
			scale=0.75,
			anchor=base west,
			text=black,
			rotate=0.0
		]{\bfseries $\mathcal{X}_\mathrm{unsafe}(\mathcal{L}_{t_1})$};
	\end{axis}

\end{tikzpicture}

%% file: tikz/scen_handover.tex
\begin{tikzpicture}
	\definecolor{darkgray176}{RGB}{176,176,176}
	\definecolor{acinblue}{RGB}{0,102,153}
	\definecolor{acinyellow}{RGB}{252, 204, 71}
	\definecolor{acingreen}{RGB}{0,190,65}
	\definecolor{acinred}{RGB}{186,18,43}
	\definecolor{lightgray204}{RGB}{204,204,204}

	\begin{axis}[
			view={80}{10},
			xlabel={$x$ / \si{\meter}},
			ylabel={$y$ / \si{\meter}},
			zlabel={$z$ / \si{\meter}},
			zmin=0,
			axis equal,
			tick align=outside,
			tick pos=left,
			x grid style={darkgray176},
			y grid style={darkgray176},
			xtick style={color=black},
			ytick style={color=black},
			legend cell align={left},
			legend columns=1,
			legend style={
					fill opacity=0.8,
					draw opacity=1,
					text opacity=1,
					draw=lightgray204,
					at={(0.45,0.85)},
					anchor=west,
				},
			legend entries ={Ours (No correction), Ours (Faster speed)}
		]
		\pgfplotstableread{data/p_traj_handover.txt}\ptraj;
		\pgfplotstableread{data/r_traj_handover.txt}\rtraj;

		\pgfplotstableread{data/p_slow_handover.txt}\pslow;

		\addplot3 [patch,patch type=rectangle, patch refines=0, shader=flat, opacity=0.25, color=acingreen, forget plot] coordinates
			{
				(0.1,0.19,0.0) (0.5,0.19,0.0) (0.5,0.6,0.0) (0.1,0.6,0.0)
				(0.1,0.19,0.25) (0.5,0.19,0.25) (0.5,0.6,0.25) (0.1,0.6,0.25)
				(0.1,0.19,0.0) (0.1,0.19,0.25)  (0.1,0.6,0.25) (0.1,0.6,0.0)
				(0.5,0.19,0.0) (0.5,0.19,0.25)  (0.5,0.6,0.25) (0.5,0.6,0.0)
			};
		\addplot3 [patch,patch type=rectangle, patch refines=0, shader=flat, opacity=0.25, color=black, forget plot] coordinates
			{
				(-1.0, -0.4, 0.0) (-1.0, -0.3, 0.0) (-1.0, -0.3, 0.8) (-1.0, -0.4, 0.8)
				(-0.3, -0.4, 0.0) (-0.3, -0.3, 0.0) (-0.3, -0.3, 0.8) (-0.3, -0.4, 0.8)
				(-0.3, -0.4, 0.0) (-0.3, -0.4, 0.8) (-1.0, -0.4, 0.8) (-1.0, -0.4, 0.0)
				(-0.3, -0.3, 0.0) (-0.3, -0.3, 0.8) (-1.0, -0.3, 0.8) (-1.0, -0.3, 0.0)
			};
		\node[inner sep=0pt, opacity=0.5] (qf) at (axis cs:0,0.11,0.475)
		{\includegraphics[width=.32\textwidth]{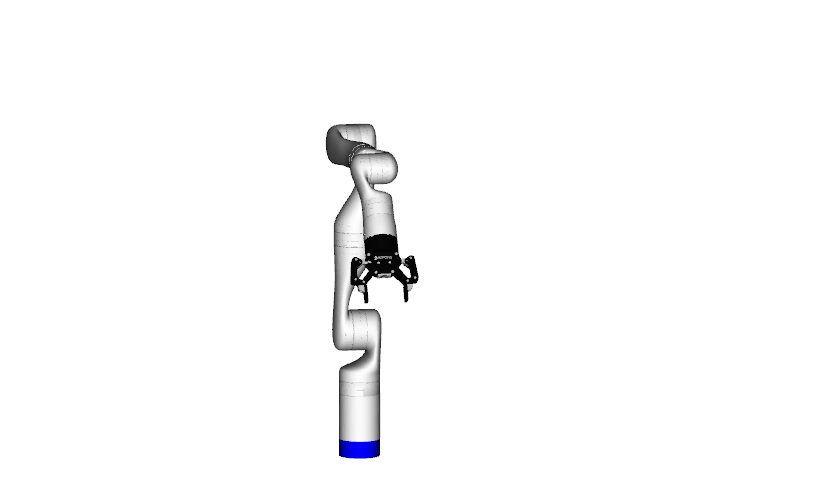}};
		\node[inner sep=0pt, opacity=0.5] (qf) at (axis cs:0,0.11,0.475)
		{\includegraphics[width=.32\textwidth]{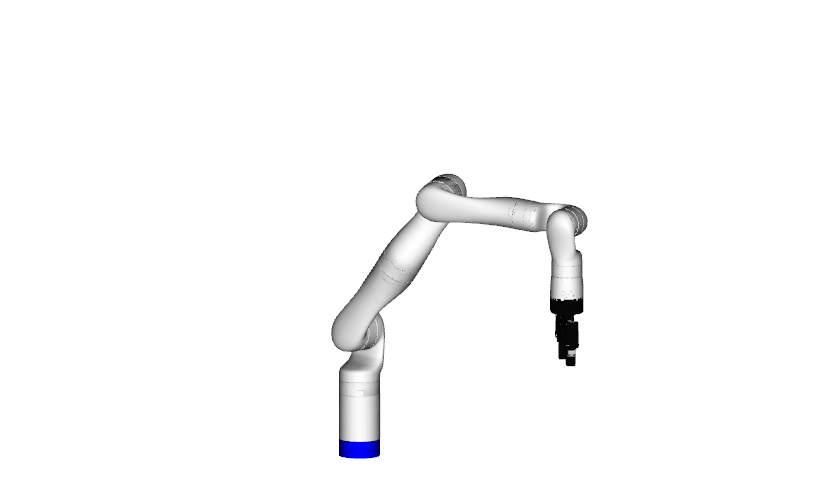}};
		\node[inner sep=0pt, opacity=0.5] (qf) at (axis cs:0,0.11,0.475)
		{\includegraphics[width=.32\textwidth]{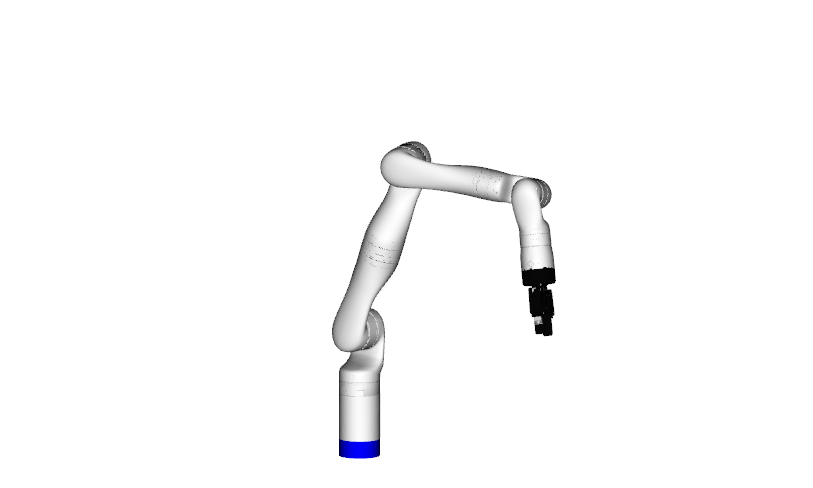}};
		\node[inner sep=0pt, opacity=0.5] (qf) at (axis cs:0,0.11,0.475)
		{\includegraphics[width=.32\textwidth]{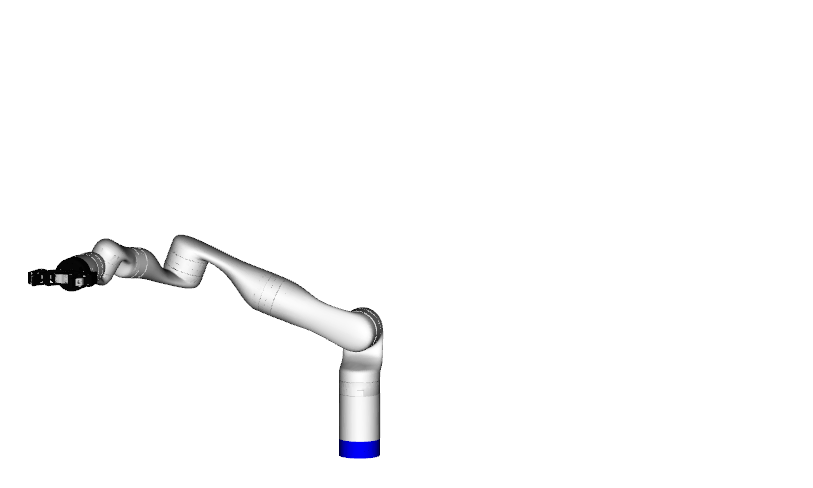}};

		\addplot3 [acinblue, thick]
		table [
				x expr=\thisrowno{0},
				y expr=\thisrowno{1},
				z expr=\thisrowno{2}
			] {\pslow};
		\addplot3 [acinred, thick]
		table [
				x expr=\thisrowno{0},
				y expr=\thisrowno{1},
				z expr=\thisrowno{2}
			] {\ptraj};

		\addplot3 [patch,patch type=rectangle, patch refines=0, shader=flat, opacity=0.25, color=black, forget plot] coordinates
			{
				(1.0, -0.4, 0.0) (1.0, -0.3, 0.0) (1.0, -0.3, 0.8) (1.0, -0.4, 0.8)
				(0.1, -0.4, 0.0) (0.1, -0.3, 0.0) (0.1, -0.3, 0.8) (0.1, -0.4, 0.8)
				(0.1, -0.4, 0.0) (0.1, -0.4, 0.8) (1.0, -0.4, 0.8) (1.0, -0.4, 0.0)
				(0.1, -0.3, 0.0) (0.1, -0.3, 0.8) (1.0, -0.3, 0.8) (1.0, -0.3, 0.0)
			};

		\addplot3[
			acinred,
			only marks,
			mark=*,
			mark size=1pt,
		] coordinates {
				(4.673250006555440539e-01, -2.485086265104662073e-02, 4.406266176809824353e-01)
				(3.943866119181922425e-01, 3.537795209367164961e-01, 3.402435563053695655e-01)
				(0.40002527, 0.40000, 0.27003069)
				(3.133537630700566345e-01, -7.500356833938049972e-01, 4.700101900858166815e-01)
			};
		\draw (axis cs:0.3,-0.9,0.4) node[
		scale=1.0,
		anchor=base west,
		text=black,
		rotate=0.0
		]{\bfseries $\vec{p}_{\mathrm{f}}$};
		\draw (axis cs:0.5,-0.1,0.37) node[
		scale=1.0,
		anchor=base west,
		text=black,
		rotate=0.0
		]{\bfseries $\vec{p}_{\mathrm{0}}$};
		\draw (axis cs:0.5,0.28,0.41) node[
		scale=1.0,
		anchor=base west,
		text=black,
		rotate=0.0
		]{\bfseries $\vec{p}_{\mathrm{1}}$};

		\node[inner sep=0pt, opacity=0.8] (screw) at (axis cs:0.4,0.4,0.27)
		{\includegraphics[width=.03\textwidth]{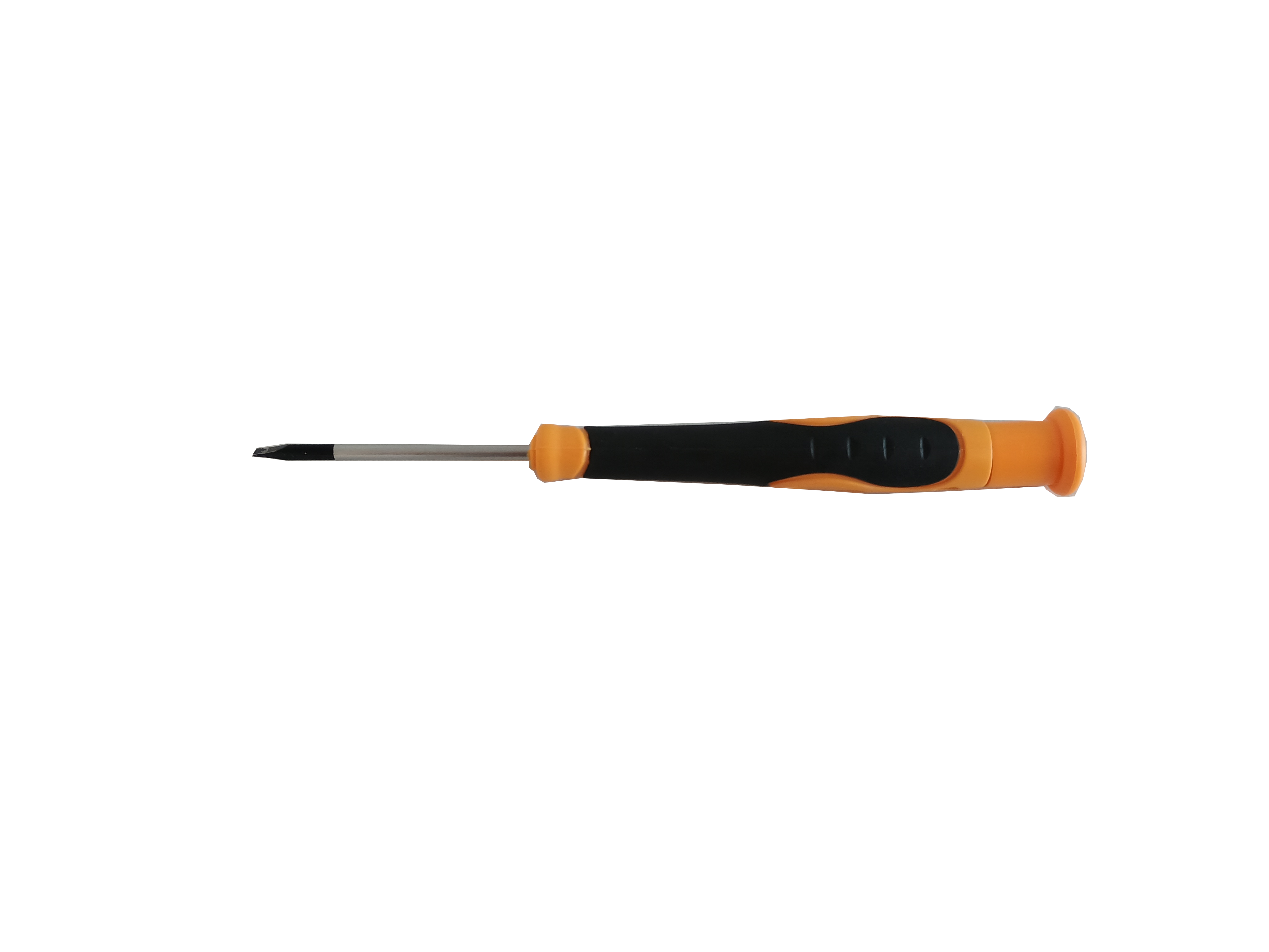}};
	\end{axis}
\end{tikzpicture}

%% file: tikz/handover_speed_fix.tex
\begin{tikzpicture}

	\definecolor{darkgrey176}{RGB}{176,176,176}
	\definecolor{firebrick1861843}{RGB}{186,18,43}
	\definecolor{lightgrey204}{RGB}{204,204,204}
	\definecolor{teal0101153}{RGB}{0,101,153}

	\begin{axis}[
			legend cell align={left},
			legend style={fill opacity=0.8, draw opacity=1, text opacity=1, draw=lightgrey204},
			tick align=outside,
			tick pos=left,
			x grid style={darkgrey176},
			xlabel={\(\displaystyle t\) / \si{\second}},
			xmajorgrids,
			xmin=0, xmax=34.20000051,
			xtick style={color=black},
			xtick={0,4.8,6.1,12.300000184,34.20000051},
			xticklabels={$0$,$4.8 \quad$,\(\displaystyle \quad 6.1\),$12.3$,$34.2$},
			y grid style={darkgrey176},
			ylabel={\(\displaystyle \lVert \vec{v} \rVert_2\) / \si{\meter\per\second}},
			ymin=0, ymax=0.708557251269611,
			ytick style={color=black},
			ytick={0,0.35,0.7},
			yticklabels={\(\displaystyle 0\),\(\displaystyle 0.35\),\(\displaystyle 0.7\)}
		]
		\addplot [semithick, firebrick1861843]
		table {%
				0 2.05627645448993e-06
				0.100000002 0.00852311141962193
				0.200000003 0.0510124725484056
				0.300000005 0.0984747017542483
				0.400000006 0.130898942132105
				0.500000008 0.147120568898081
				0.600000009 0.152617633106407
				0.700000011 0.152962372973457
				0.800000012 0.151696310323084
				0.900000014 0.150417301795099
				1.000000015 0.149523854713944
				1.100000017 0.148895691172524
				1.200000018 0.148307931286416
				1.30000002 0.14760544100315
				1.400000021 0.146730960752753
				1.500000023 0.145692557641739
				1.600000024 0.144522537738619
				1.700000026 0.143249234997106
				1.800000027 0.141884560845855
				1.900000029 0.140422790049176
				2.00000003 0.138845323092221
				2.100000032 0.137124247555337
				2.200000033 0.135227560645429
				2.300000035 0.133122596801629
				2.400000036 0.13077773891047
				2.500000038 0.12816254337228
				2.600000039 0.124249961115099
				2.700000041 0.116319814666046
				2.800000042 0.108583766136243
				2.900000044 0.104374210381159
				3.000000045 0.103530116404736
				3.100000047 0.103805907050039
				3.200000048 0.102616638399045
				3.30000005 0.0980729534806136
				3.400000051 0.0896141785068219
				3.500000053 0.0788673626152647
				3.600000054 0.0697034713499686
				3.700000056 0.0642354354288374
				3.800000057 0.0602389745063603
				3.900000059 0.0553617203547693
				4.00000006 0.0491547925219351
				4.100000062 0.0422350704075399
				4.200000063 0.0353625223171833
				4.300000065 0.0290689504453109
				4.400000066 0.0236203045797701
				4.500000067 0.0190861833621362
				4.600000069 0.0154183296788442
				4.70000007 0.0125233811117944
				4.800000072 0.0103019273442751
				4.900000073 0.00400731958849101
				5.000000075 0.0323729208734946
				5.100000076 0.0719764860047595
				5.200000078 0.100642885401478
				5.300000079 0.113332947456193
				5.400000081 0.112054788352068
				5.500000082 0.102220149297481
				5.600000084 0.0900922628804909
				5.700000085 0.0807983850281424
				5.800000087 0.0765813618100891
				5.900000088 0.0764117391758549
				6.00000009 0.0778846578529356
				6.100000091 0.079209971020429
				6.200000093 0.0796753531298931
				6.300000094 0.0792841447101774
				6.400000096 0.0783365046183952
				6.500000097 0.0771606338789675
				6.600000099 0.0760106081603557
				6.7000001 0.0750288567041726
				6.800000102 0.0742650214265321
				6.900000103 0.0737114514015575
				7.000000105 0.0733341083299902
				7.100000106 0.0730927075616326
				7.200000108 0.0729508897747837
				7.300000109 0.0728797363614877
				7.400000111 0.0728578013109377
				7.500000112 0.072869779519738
				7.600000114 0.0729049173832735
				7.700000115 0.0729556315535329
				7.800000117 0.0730165259864582
				7.900000118 0.0730835886361117
				8.00000012 0.0731536908521777
				8.100000121 0.0732242888537119
				8.200000123 0.0732932456987392
				8.300000124 0.0733587426099385
				8.400000126 0.0734192327184387
				8.500000127 0.0734734226305827
				8.600000129 0.0735202495155248
				8.70000013 0.0735588447387854
				8.800000132 0.0735885169044505
				8.900000133 0.0736087821531221
				9.000000135 0.0736193329428929
				9.100000136 0.0736194027892063
				9.200000138 0.0736073500809398
				9.300000139 0.0735861544108768
				9.400000141 0.0735574077631043
				9.500000142 0.073520649391574
				9.600000143 0.0734747327433726
				9.700000145 0.0734187675365954
				9.800000146 0.0733520275769932
				9.900000148 0.073274173278297
				10.000000149 0.0731852418876324
				10.100000151 0.0730854617836328
				10.200000152 0.0729751283674794
				10.300000154 0.0728546399714956
				10.400000155 0.0727248008801812
				10.500000157 0.0725859266195523
				10.600000158 0.0724383802559116
				10.70000016 0.072282132054791
				10.800000161 0.0721173792180651
				10.900000163 0.0719406826993458
				11.000000164 0.0716929286831543
				11.100000166 0.0711517757067398
				11.200000167 0.07012660141319
				11.300000169 0.0686450093198889
				11.40000017 0.0668772835186196
				11.500000172 0.0650424712165759
				11.600000173 0.0632420984199001
				11.700000175 0.0603758092224541
				11.800000176 0.0550100981147117
				11.900000178 0.0530540310003827
				12.000000179 0.0544633036369225
				12.100000181 0.0575121905179193
				12.200000182 0.060934747330661
				12.300000184 0.0639769781242123
				12.400000185 0.0663289036382425
				12.500000187 0.0679762394013158
				12.600000188 0.0690443033411649
				12.70000019 0.0696927300008551
				12.800000191 0.0700581163003835
				12.900000193 0.0702360479203447
				13.000000194 0.0702874223746271
				13.100000196 0.0702522608438424
				13.200000197 0.0701608496533764
				13.300000199 0.070039099224415
				13.4000002 0.0699092636872874
				13.500000202 0.0697883902930527
				13.600000203 0.0697481178062325
				13.700000205 0.0700017486644811
				13.800000206 0.0704827660028643
				13.900000208 0.0711476239303187
				14.000000209 0.0719767833026701
				14.100000211 0.0727694322040502
				14.200000212 0.0732682873132337
				14.300000214 0.0733076240246667
				14.400000215 0.0727062740548816
				14.500000217 0.0712742830961205
				14.600000218 0.0689592413968442
				14.700000219 0.0660222123018535
				14.800000221 0.0630864756888681
				14.900000222 0.0606495587585332
				15.000000224 0.058946893307611
				15.100000225 0.0580510579303517
				15.200000227 0.0579398096437312
				15.300000228 0.0585004796508064
				15.40000023 0.0594472109858318
				15.500000231 0.060409679453657
				15.600000233 0.0612578426775601
				15.700000234 0.0619353764905725
				15.800000236 0.0625142490628119
				15.900000237 0.0630951134142652
				16.000000239 0.0637544880077742
				16.10000024 0.0645542477325937
				16.200000242 0.0655716183649146
				16.300000243 0.0665868812678719
				16.400000245 0.06749976511205
				16.500000246 0.0682855167226716
				16.600000248 0.0689361207814151
				16.700000249 0.0694618622271657
				16.800000251 0.0698814199217589
				16.900000252 0.0702111702031247
				17.000000254 0.070457051455903
				17.100000255 0.0706174010632308
				17.200000257 0.0706942980350082
				17.300000258 0.0706800326018225
				17.40000026 0.0706002777088858
				17.500000261 0.0704950225550692
				17.600000263 0.0703717690219356
				17.700000264 0.0702497263241179
				17.800000266 0.0701348569249648
				17.900000267 0.0700470225728598
				18.000000269 0.0699747491697746
				18.10000027 0.0699110132819947
				18.200000272 0.069854072072195
				18.300000273 0.0697996831725828
				18.400000275 0.0697618635340383
				18.500000276 0.0697284317482819
				18.600000278 0.0696923536554466
				18.700000279 0.0694526625669838
				18.800000281 0.0682923511663854
				18.900000282 0.0664427144515956
				19.000000284 0.0643127761854886
				19.100000285 0.0621862709572932
				19.200000287 0.0602561832369878
				19.300000288 0.0587059918600083
				19.40000029 0.0578328636485388
				19.500000291 0.0581624452622573
				19.600000293 0.0602744394517173
				19.700000294 0.0638015456811792
				19.800000295 0.0672441729243258
				19.900000297 0.0696491611812601
				20.000000298 0.0709039499916448
				20.1000003 0.0712905081738668
				20.200000301 0.0711684278761849
				20.300000303 0.0708245837122722
				20.400000304 0.0704359364188927
				20.500000306 0.0700856511500556
				20.600000307 0.0697956895711991
				20.700000309 0.0695563265486682
				20.80000031 0.0693466836873327
				20.900000312 0.0691463734225416
				21.000000313 0.0689404091443453
				21.100000315 0.0687199233555856
				21.200000316 0.0684808405623343
				21.300000318 0.0682219842508549
				21.400000319 0.0679434327017986
				21.500000321 0.0676454256641313
				21.600000322 0.0673278104118613
				21.700000324 0.066989877050848
				21.800000325 0.0666304113536224
				21.900000327 0.0662478300633922
				22.000000328 0.0658403178445311
				22.10000033 0.0654059320076025
				22.200000331 0.0649426629765836
				22.300000333 0.0644484618758052
				22.400000334 0.0639212499323112
				22.500000336 0.063358922082483
				22.600000337 0.0627593535169981
				22.700000339 0.0621204140497722
				22.80000034 0.0614399927896293
				22.900000342 0.0607159555778374
				23.000000343 0.0599460947527227
				23.100000345 0.0591287740535093
				23.200000346 0.0582624933606471
				23.300000348 0.057345816713088
				23.400000349 0.0563774758469769
				23.500000351 0.0553564693979248
				23.600000352 0.054282191358237
				23.700000354 0.0531546404536181
				23.800000355 0.0519747414482781
				23.900000357 0.050744717523365
				24.000000358 0.0494682820259453
				24.10000036 0.0481504645652346
				24.200000361 0.0467971883777345
				24.300000363 0.0454145329789363
				24.400000364 0.0440081596004447
				24.500000366 0.0425826342536497
				24.600000367 0.0411429670501213
				24.700000368 0.0396947432710944
				24.80000037 0.0382439530722927
				24.900000371 0.0367966948107906
				25.000000373 0.0353587831704762
				25.100000374 0.033935422683649
				25.200000376 0.0325310394685372
				25.300000377 0.0311492794872875
				25.400000379 0.0297931326839619
				25.50000038 0.0284651202103711
				25.600000382 0.0271674844684511
				25.700000383 0.0259023279325315
				25.800000385 0.0246716799526831
				25.900000386 0.023477604548128
				26.000000388 0.0223220598766645
				26.100000389 0.0212067723371115
				26.200000391 0.0201331288303511
				26.300000392 0.0191020942132868
				26.400000394 0.0181141727893139
				26.500000395 0.0171694116584636
				26.600000397 0.0162674393146438
				26.700000398 0.0154075268309711
				26.8000004 0.0145886590498936
				26.900000401 0.0138096069888081
				27.000000403 0.0130689960908675
				27.100000404 0.0123653671572595
				27.200000406 0.011697226094974
				27.300000407 0.0110630812881957
				27.400000409 0.0104614680902126
				27.50000041 0.00989096130863468
				27.600000412 0.0093501789566556
				27.700000413 0.00883778091505014
				27.800000415 0.0083524656911173
				27.900000416 0.00789296753668665
				28.000000418 0.0074580550718372
				28.100000419 0.00704653162067919
				28.200000421 0.00665723685829664
				28.300000422 0.00628904911458281
				28.400000424 0.0059408994216804
				28.500000425 0.00561178781358113
				28.600000427 0.00530069177396414
				28.700000428 0.00500663315415651
				28.80000043 0.0047286963010585
				28.900000431 0.00446601896914517
				29.000000433 0.00421778562041648
				29.100000434 0.00398322345998782
				29.200000436 0.00376159871146206
				29.300000437 0.00355221364217669
				29.400000439 0.00335440499236694
				29.50000044 0.0031675429781365
				29.600000442 0.00299103022888428
				29.700000443 0.00282430048057402
				29.800000444 0.00266681713750946
				29.900000446 0.00251807182908301
				30.000000447 0.00237758311572664
				30.100000449 0.00224489531638298
				30.20000045 0.00211957738511852
				30.300000452 0.00200122176391554
				30.400000453 0.00188944323800821
				30.500000455 0.00178387853096671
				30.600000456 0.00168418788015443
				30.700000458 0.00159005178709749
				30.800000459 0.00150116312394563
				30.900000461 0.00141722861764111
				31.000000462 0.00133797084038672
				31.100000464 0.00126312799201806
				31.200000465 0.00119245243148196
				31.300000467 0.0011257119770154
				31.400000468 0.00106268893765081
				31.50000047 0.00100317830271641
				31.600000471 0.000946986451513606
				31.700000473 0.000893930282405721
				31.800000474 0.000843836579750959
				31.900000476 0.000796645622534634
				32.000000477 0.000752630377446128
				32.100000479 0.000711768982145682
				32.20000048 0.000673804743052562
				32.300000482 0.000638426267942815
				32.400000483 0.000605276887665367
				32.500000485 0.000574024021086865
				32.600000486 0.000544403579842207
				32.700000488 0.000516224886337406
				32.800000489 0.000489357148309051
				32.900000491 0.000463712164664565
				33.000000492 0.000439229907903062
				33.100000494 0.000415868060479461
				33.200000495 0.000393594601715145
				33.300000497 0.000372382588007431
				33.400000498 0.000352206622242666
				33.5000005 0.00033304035790344
				33.600000501 0.000314854666324218
				33.700000503 0.00029761723003045
				33.800000504 0.000281292462474448
				33.900000506 0.000265842457090688
				34.000000507 0.000251227901296434
				34.100000509 0.000237406194379619
				34.20000051 0.000224336851479652
			};
		\addlegendentry{Ours (No correction)}
		\addplot [semithick, teal0101153]
		table {%
				0 2.05139476082407e-06
				0.100000002 0.00852311348088697
				0.200000003 0.0510124721927731
				0.300000005 0.0984747003917308
				0.400000006 0.130898940617837
				0.500000008 0.1471205677113
				0.600000009 0.152617632372666
				0.700000011 0.152962372617251
				0.800000012 0.151696310217074
				0.900000014 0.150417301826164
				1.000000015 0.149523854802888
				1.100000017 0.148895691270407
				1.200000018 0.148307931363414
				1.30000002 0.147605441167007
				1.400000021 0.14673096135941
				1.500000023 0.145692558371584
				1.600000024 0.144522538318151
				1.700000026 0.143249235376449
				1.800000027 0.141884562315363
				1.900000029 0.140422796250346
				2.00000003 0.138845330677846
				2.100000032 0.137124253656791
				2.200000033 0.13522756448873
				2.300000035 0.133122598592271
				2.400000036 0.130777739165481
				2.500000038 0.128162542579724
				2.600000039 0.124249959524166
				2.700000041 0.1163198115297
				2.800000042 0.108583761106889
				2.900000044 0.104374207834713
				3.000000045 0.103530113597059
				3.100000047 0.103805902711215
				3.200000048 0.102616637828659
				3.30000005 0.0980729516469165
				3.400000051 0.089614169775922
				3.500000053 0.0788673491953938
				3.600000054 0.069703460732966
				3.700000056 0.0642354287178954
				3.800000057 0.0602389697894145
				3.900000059 0.0553617161371483
				4.00000006 0.0491547881339561
				4.100000062 0.0422350655167125
				4.200000063 0.0353625166369854
				4.300000065 0.0290689435276654
				4.400000066 0.0236202955376345
				4.500000067 0.0190861792182105
				4.600000069 0.0154183619758575
				4.70000007 0.0125234328006108
				4.800000072 0.0103019656154499
				4.900000073 0.00400729924973468
				5.000000075 0.0323726513024637
				5.100000076 0.0719759079163287
				5.200000078 0.100642035464513
				5.300000079 0.11333193830094
				5.400000081 0.11205376771349
				5.500000082 0.102219279477149
				5.600000084 0.0900917012974603
				5.700000085 0.0807982357782433
				5.800000087 0.076581604572191
				5.900000088 0.0764122363867733
				6.00000009 0.0778852499872312
				6.100000091 0.0792105698654385
				6.200000093 0.102751619186338
				6.300000094 0.240478941256854
				6.400000096 0.423149911611812
				6.500000097 0.574511918452316
				6.600000099 0.671118673830689
				6.7000001 0.708557251269611
				6.800000102 0.692850305415798
				6.900000103 0.642033700900971
				7.000000105 0.57519661741904
				7.100000106 0.50547261630453
				7.200000108 0.442004612847029
				7.300000109 0.393348045091184
				7.400000111 0.370567525398409
				7.500000112 0.381187608706367
				7.600000114 0.422452507401715
				7.700000115 0.475561144957381
				7.800000117 0.518814032239234
				7.900000118 0.537221219396921
				8.00000012 0.522700408900988
				8.100000121 0.481702835334935
				8.200000123 0.439543259184578
				8.300000124 0.402656404220022
				8.400000126 0.382258487061691
				8.500000127 0.375949435253721
				8.600000129 0.36459093357715
				8.70000013 0.340374642425843
				8.800000132 0.307534650019634
				8.900000133 0.273475078347226
				9.000000135 0.244370680451929
				9.100000136 0.22283731186289
				9.200000138 0.209753719613475
				9.300000139 0.201444753168827
				9.400000141 0.193200813339046
				9.500000142 0.182309500504018
				9.600000143 0.168341684477736
				9.700000145 0.15213989642587
				9.800000146 0.134759036384851
				9.900000148 0.116915646645685
				10.000000149 0.0990174932755331
				10.100000151 0.0814615088938848
				10.200000152 0.0648192505236017
				10.300000154 0.0497748212584531
				10.400000155 0.0369230241337984
				10.500000157 0.026588695310966
				10.600000158 0.0187541366668928
				10.70000016 0.0131018809411348
				10.800000161 0.00914621599242925
				10.900000163 0.00641403513297922
				11.000000164 0.00460582052745374
				11.100000166 0.00363231661080425
				11.200000167 0.00340009589689493
				11.300000169 0.00357304540241114
				11.40000017 0.00378478387402783
				11.500000172 0.00384178281914934
				11.600000173 0.00368813236470098
				11.700000175 0.00334089437237504
				11.800000176 0.00285160843352514
				11.900000178 0.00228409324705517
				12.000000179 0.00170139866489134
				12.100000181 0.00116007729119362
				12.200000182 0.00071476390984435
				12.300000184 0.000445713592138438
			};
		\addlegendentry{Ours (With speedup)}
	\end{axis}

\end{tikzpicture}